\documentclass[11pt,a4paper]{article}
\PassOptionsToPackage{table}{xcolor}
\usepackage{preprint}
\usepackage[numbers,sort&compress]{natbib}
\usepackage{booktabs,tabularx,multirow,pifont,listings,wrapfig}
\usepackage[capitalise]{cleveref}
\setEyebrow{Preprint}
\paperauthors{%
  {\centering\bfseries\textcolor{black}{EvoSkill3D Research Team}\par}%

  \vspace{0.35em}%

  {\centering\small
    \href{https://github.com/Fabulous1496/EvoSkill3D}{%
      \raisebox{-0.15\height}{%
        \includegraphics[height=1.05em]{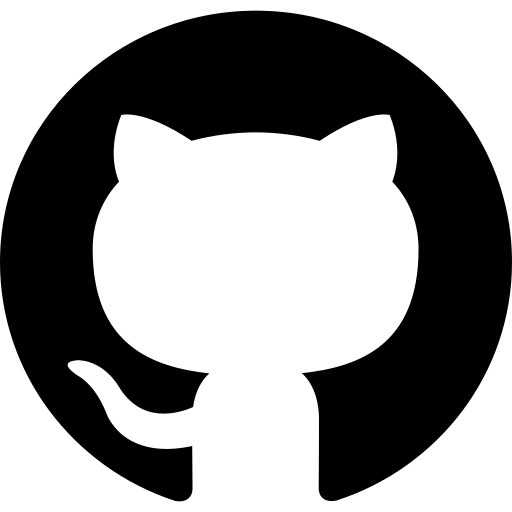}}%
      \hspace{0.3em}\textcolor{black}{GitHub}}%
    \hspace{1.5em}%
    \href{https://huggingface.co/datasets/Fabulous1496/EvoSkill3D-Bench}{%
      \raisebox{-0.15\height}{%
        \includegraphics[height=1.05em]{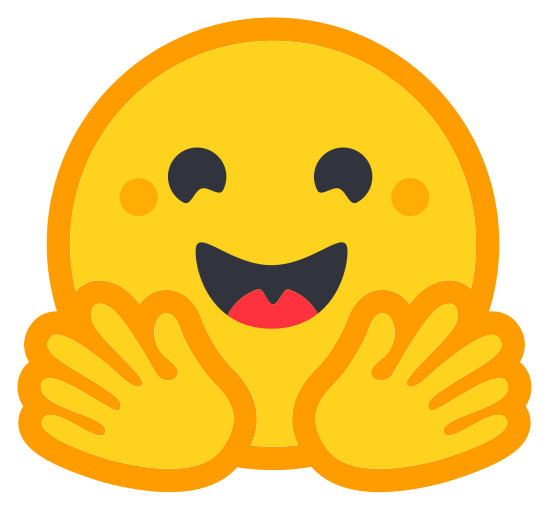}}%
      \hspace{0.3em}\textcolor{black}{Hugging Face}}%
    \par
  }%
}

\paperaffil{}

\papernotes{}
\newcommand{\institutionheader}[1]{%
  \includegraphics[height=#1]{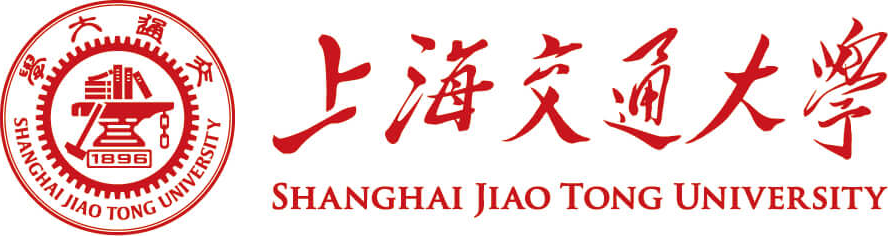}\hfill
  \includegraphics[height=#1]{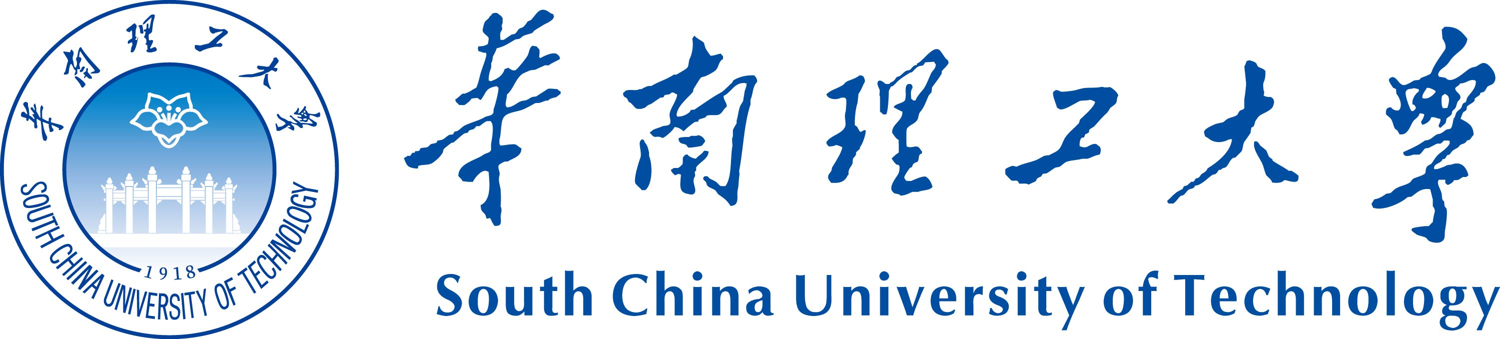}\hfill
  \includegraphics[height=#1]{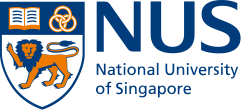}\hfill
  \includegraphics[height=#1]{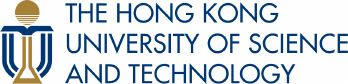}%
  \hspace{3mm}{\color{ZJUBlue!65}\rule{0.5pt}{#1}}\hspace{3mm}%
  \includegraphics[height=#1]{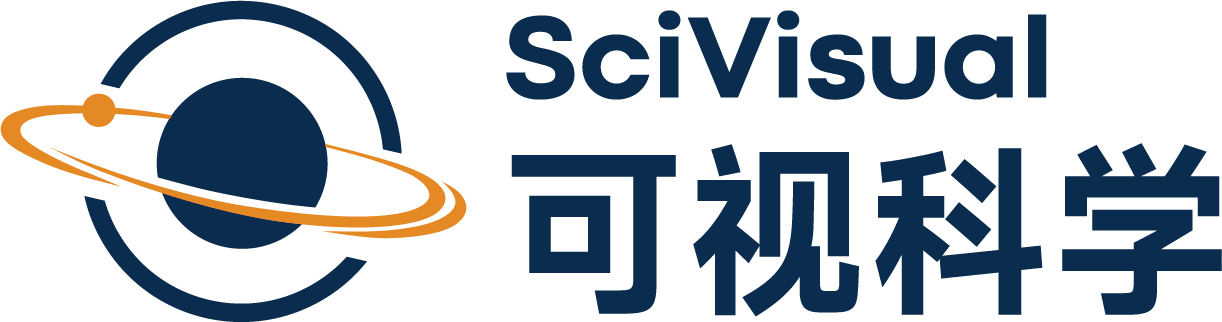}}

\setReportTag{EvoSkill3D}
\newcommand{\modellogo}[1]{\raisebox{-0.2\height}{\includegraphics[height=2.5ex]{figures/logos/#1.png}}}
\usepackage{etoc}
\newcommand{\cmark}{\textcolor{green!60!black}{\ding{51}}}
\newcommand{\pmark}{\textcolor{orange!85!black}{$\triangle$}}
\newcommand{\xmark}{\textcolor{red}{\ding{55}}}
\renewenvironment{abstract}{\begin{preprintAbstract}\noindent\textbf{Abstract.}\enspace}{\end{preprintAbstract}}
\papertitle{EvoSkill3D: Benchmarking Skill-Centric Recursive Self-Improvement in 3D Coding Agents}
\hypersetup{pdftitle={EvoSkill3D: Benchmarking Skill-Centric Recursive Self-Improvement in 3D Coding Agents}}
\begin{document}
\etocdepthtag.toc{main}
\makepreprinttitle
\begin{abstract}
Recent advances in LLM-based agents have enabled code-driven 3D modeling. Existing benchmarks evaluate diverse modeling tasks using geometric, structural, and visual metrics. However, they primarily focus on individual task outcomes, leaving skill acquisition from feedback, cross-task transfer, and multi-skill composition largely unexplored. We introduce \textbf{EvoSkill3D}, a unified benchmark and recursive self-improvement framework for studying skill evolution in 3D coding agents. EvoSkill3D-Bench comprises 1,000 reconstruction tasks organized around five foundational modeling skills, including challenging multi-skill scenarios. By integrating skill-oriented task organization with ground-truth-based evaluation, it provides fine-grained feedback to diagnose modeling deficiencies and guide skill refinement. Building on this design, we develop EvoSkill3D-RSI, an Actor--Evaluator--Evolver framework that enables agents to learn from experience and accumulate reusable skills across tasks. These skills are maintained in a hierarchical Library--Tree--Graph structure to support skill organization, reuse, and cross-skill interaction. To assess these capabilities, EvoSkill3D-Bench further establishes evaluation protocols for skill acquisition, cross-task transfer, and multi-skill composition. EvoSkill3D offers a unified testbed for evaluating and advancing self-improving 3D coding agents.
\end{abstract}

\begin{figure}[!htbp]
    \centering
    \includegraphics[width=\textwidth]{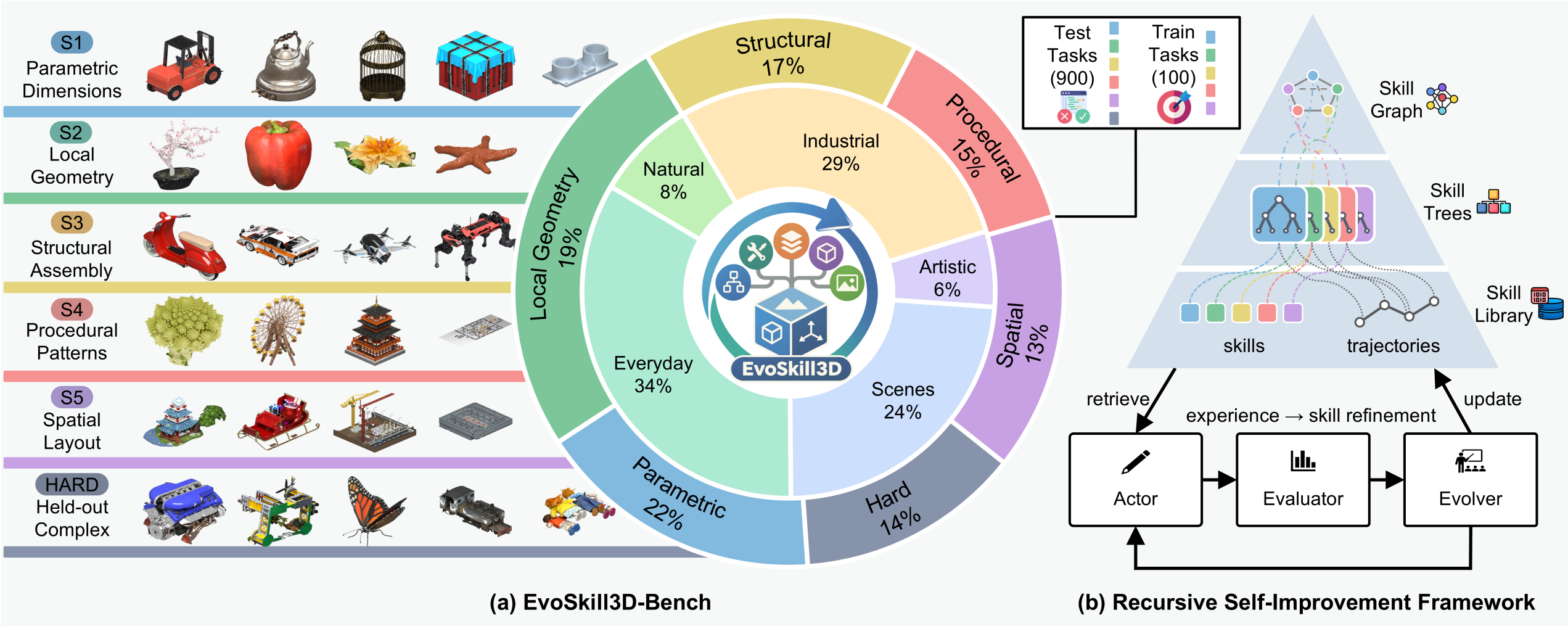}
    \caption{
        \textbf{Overview of EvoSkill3D.}
        We propose EvoSkill3D, which consists of a 3D reconstruction benchmark and a recursive self-improvement framework. 
        \textbf{(a)} EvoSkill3D-Bench.The benchmark covers six categories of 3D reconstruction tasks, with 900 tasks for evaluation and 100 reserved for agent self-improvement.
        \textbf{(b)} Recursive Self-Improvement Framework. The framework organizes reusable skills into a hierarchical structure of skill libraries, skill trees, and skill graphs. Through an iterative Actor–Evaluator–Evolver loop, agents can retrieve and execute skills, evaluate task outcomes, and refine skills based on accumulated experience.
    }
    \label{fig:overview}
\end{figure}

\section{Introduction}
\label{sec:intro}

Recent advances in LLMs have enabled agents to use code for software interaction and digital content creation~\cite{wang2024executable,liang2023code,ning2026code}. Unlike GUI-based agents, which rely on screenshots and sequences of mouse--keyboard actions, code-based agents can directly invoke application APIs, reducing the need for repeated visual interactions and making operations easier to modify and reuse~\cite{xie2024osworld,qin2025ui,abhyankar2026osworld,abaskohi2026revision,song2025coact,chen2026autorpa,huang2026work2skill}. This is particularly useful for 3D reconstruction, where agents can generate Blender Python scripts to construct geometry, assemble objects, and arrange scenes. These scripts can also be refined and reused across tasks, providing a basis for acquiring reusable modeling skills~\cite{lu2025ll3m,gao20263dcodebench,li2026lego}.

To evaluate these capabilities, recent benchmarks have introduced a range of code-driven 3D modeling tasks, from individual object reconstruction to complex scene creation, with metrics covering geometric accuracy, structural consistency, and visual quality~\cite{gu2025blendergym,gao20263dcodebench,yang2026p3d,fang2026blenderworld}. Some recent systems also support iterative refinement through execution or visual feedback~\cite{li2026lego,wang20263dcodeedit}. However, these evaluations largely focus on the outcome of individual tasks. Improving a reconstruction through repeated attempts does not necessarily mean that an agent has learned something useful, how agents can acquire reusable modeling skills from experience, transfer them to new tasks, and combine them to handle more complex reconstructions remains an open question.

3D reconstruction offers a natural setting for studying these capabilities. Across different objects and scenes, modeling tasks often involve a shared set of fundamental skills, including dimensional modeling, local geometry construction, structural assembly, procedural patterns, and spatial layout. While simpler tasks may rely on only a subset of these skills, complex reconstructions often require several of them to work together. However, most existing evaluation methods assess reconstruction quality at the whole-model level, providing only coarse-grained scores without precisely locating geometric errors or offering actionable feedback for improvement~\cite{gu2025blendergym,gao20263dcodebench,fang2026blenderworld}. This makes it difficult to identify specific modeling deficiencies and determine which skills need to be refined.

To address these challenges, we introduce \textbf{EvoSkill3D}, a benchmark and recursive self-improvement framework for code-driven 3D reconstruction. Its benchmark component, EvoSkill3D-Bench, consists of 1,000 object and scene reconstruction tasks organized into five foundational modeling skills, together with a challenging subset of multi-skill tasks. Each task provides calibrated multi-view images and camera parameters, while the corresponding 3D asset is kept private for evaluation. Agents are required to generate executable Blender Python programs to reconstruct the target geometry. To assess reconstruction quality beyond overall similarity, we develop a ground-truth-based evaluator with six complementary metrics covering surface accuracy, geometric details, absolute scale, and semantic similarity. It adaptively partitions the ground-truth geometry into blocks based on local geometric density and evaluates each block individually, providing spatially localized feedback that pinpoints
reconstruction errors within an asset. This evaluation design provides both a consistent measure of reconstruction quality and detailed feedback for subsequent skill refinement.

Building on EvoSkill3D-Bench, we introduce \textbf{EvoSkill3D-RSI}, an Actor--Evaluator--Evolver framework for developing reusable modeling skills across reconstruction
tasks~\cite{he2510evotest,tang2026trinity}.
The Actor generates Blender Python programs, the Evaluator provides structured geometric feedback, and the Evolver uses this feedback to refine existing skills and distill reusable procedures.
These are maintained in a persistent repository comprising a Library of shared tools and modeling experiences, five Skill Trees aligned with the benchmark's foundational skills, and a Skill Graph linking related procedures.
The agent thus improves through skill accumulation and reuse while its model parameters remain fixed.
We further define Skill Acquisition, Transfer, and Composition protocols to assess whether learned skills support subsequent tasks, generalize across skill categories, and combine for complex tasks.

In summary, our contributions are threefold:
\begin{itemize}
    \item We introduce EvoSkill3D-Bench, a skill-centric benchmark comprising 1,000 reconstruction tasks across five foundational modeling skills and challenging multi-skill scenarios, with dedicated protocols for evaluating skill acquisition, transfer, and composition.

    \item We develop a fine-grained ground-truth-based evaluator that combines six complementary reconstruction metrics with detail-aware geometric partitioning, providing fine-grained, spatially localized feedback to support capability assessment and skill refinement.

    \item We propose EvoSkill3D-RSI, an Actor--Evaluator--Evolver framework with a hierarchical Library--Tree--Graph skill organization, enabling persistent skill acquisition, refinement, and cross-task reuse through feedback-driven evolution.
\end{itemize}

\section{Related Work}
\label{sec:Related Works}

\begin{table*}[t]
\centering
\caption{
\textbf{Comparison of existing 3D modeling and reconstruction benchmarks.}
\cmark, \xmark, and \pmark indicate full, absent, and partial support.
\textbf{Code-based} denotes programmatic 3D modeling;
\textbf{3D Model GT} indicates the availability of reference 3D geometry;
\textbf{Fine-grained Geometry} refers to detailed geometric evaluation beyond global similarity;
\textbf{Skill Taxonomy} denotes an explicit categorization of modeling skills;
and \textbf{Skill-centric RSI Evaluation} indicates whether recursive self-improvement is explicitly evaluated.
}
\label{tab:benchmark_comparison}
\footnotesize
\renewcommand{\arraystretch}{1.2}

\begin{tabular*}{\textwidth}{@{\extracolsep{\fill}}lccccc@{}}
\toprule
\textbf{Benchmark}
& \shortstack{\textbf{Code-}\\\textbf{based}}
& \shortstack{\textbf{3D Model}\\\textbf{GT}}
& \shortstack{\textbf{Fine-grained}\\\textbf{Geometry}}
& \shortstack{\textbf{Skill}\\\textbf{Taxonomy}}
& \shortstack{\textbf{Skill-centric}\\\textbf{RSI Evaluation}} \\
\midrule

BlenderGym~\cite{gu2025blendergym}
& \cmark & \cmark & \pmark & \pmark & \xmark \\

3DCodeBench~\cite{gao20263dcodebench}
& \cmark & \cmark & \pmark & \pmark & \xmark \\

P3D-Bench~\cite{yang2026p3d}
& \cmark & \cmark & \cmark & \pmark & \xmark \\

BlenderWorld~\cite{fang2026blenderworld}
& \pmark & \xmark & \pmark & \pmark & \xmark \\

3DCodeEdit~\cite{wang20263dcodeedit}
& \cmark & \pmark & \pmark & \cmark & \xmark \\

AssemblyWorld~\cite{zhang2026assemblyworld}
& \xmark & \cmark & \pmark & \pmark & \xmark \\

BVB~\cite{tang2026bvb}
& \cmark & \xmark & \xmark & \xmark & \xmark \\

\midrule
\textbf{EvoSkill3D (Ours)}
& \textbf{\cmark}
& \textbf{\cmark}
& \textbf{\cmark}
& \textbf{\cmark}
& \textbf{\cmark} \\

\bottomrule
\end{tabular*}
\end{table*}

\paragraph{Code-based 3D Modeling Agents.}
Recent studies have explored the use of large language models (LLMs)
for 3D modeling through code generation.
3D-GPT~\cite{sun20253d} introduces a multi-agent framework
for instruction-driven procedural 3D modeling, consisting of
a task dispatch agent, a conceptualization agent, and a modeling
agent. These agents decompose modeling tasks into manageable
steps and generate Python code to interface with 3D software.
SceneCraft~\cite{hu2024scenecraft} presents an LLM agent that
converts text descriptions into Blender-executable Python scripts.
It constructs a scene graph to represent spatial relationships
among assets, translates these relationships into numerical
constraints for scene layout, and iteratively refines the generated
scenes based on visual feedback from VLMs. LL3M~\cite{lu2025ll3m} reformulates 3D shape generation as a code-writing task and employs specialized agents to plan, retrieve,
write, debug, and refine Blender scripts. More recently, LEGO-Anything~\cite{li2026lego} presents an Image-to-Code framework in which a coding agent iteratively writes and executes Blender code, inspects the resulting scenes and renderings, and revises the program
for 3D reconstruction.

\paragraph{3D Reconstruction Benchmarks.}
Existing benchmarks for code-based 3D modeling evaluate reconstruction
quality through geometric fidelity, perceptual similarity, and
structural consistency, as summarized in
Table~\ref{tab:benchmark_comparison}.
BlenderGym~\cite{gu2025blendergym} evaluates graphics editing
by comparing modified Blender scenes against target scenes,
while 3DCodeBench~\cite{gao20263dcodebench} assesses procedural
3D modeling using geometric and visual similarity metrics.
P3D-Bench~\cite{yang2026p3d} further evaluates fine-grained
geometric fidelity and part-level structural accuracy.
BlenderWorld~\cite{fang2026blenderworld} considers diverse input
formats and evaluates generated scenes through multi-view rendering
and task-specific criteria.
3DCodeEdit~\cite{wang20263dcodeedit} introduces a taxonomy of
program editing tasks, ranging from parameter modifications to
structural changes and scene composition.
AssemblyWorld~\cite{zhang2026assemblyworld} evaluates object
assembly through geometric correspondence with ground-truth
assemblies, while BVB~\cite{tang2026bvb} assesses programmatic
video reconstruction using perceptual and semantic comparisons.
Despite these advances, existing benchmarks do not systematically
evaluate how agents acquire reusable modeling skills, transfer them
across tasks, and compose them to solve complex problems. EvoSkill3D addresses this gap through skill-centric evaluation protocols for acquisition, transfer, and composition.

\paragraph{Recursive Self-Improvement.}
Self-improvement in LLM agents has been studied through iterative
refinement and learning from previous task experiences.
Self-Refine~\cite{madaan2023self} and
Reflexion~\cite{shinn2023reflexion} use feedback from previous
attempts to improve subsequent outputs and decisions.
Voyager~\cite{wang2023voyager} maintains a skill library to
support skill reuse, while LILO~\cite{grand2024lilo}
learns reusable code abstractions from task experience.
Building on these ideas, SkillWeaver~\cite{zheng2025skillweaver}
explores skill discovery and refinement through interaction with
the environment, and CoEvoSkills~\cite{zhang2026coevoskills}
investigates the co-evolution of skill generation and verification.
Graph-of-Skills~\cite{liu2026graph} organizes reusable skills
through dependency-aware graphs to support efficient retrieval,
while SkillsBench~\cite{li2026skillsbench} evaluates the
effectiveness of agent skills across diverse tasks.
Despite these efforts, the systematic evaluation of skill-centric
recursive self-improvement in code-based 3D reconstruction remains underexplored.

\section{EvoSkill3D}
\label{sec:evoskill3d}

We present \textbf{EvoSkill3D}, consisting of two complementary components. \textbf{EvoSkill3D-Bench} (Sec.~\ref{sec:evoskill3d-bench}) evaluates code-based 3D reconstruction across five foundational modeling skills using reference geometry, while \textbf{EvoSkill3D-RSI} (Sec.~\ref{sec:evoskill3d-rsi}) studies skill acquisition, transfer, and composition through recursive self-improvement without updating model parameters.

\subsection{EvoSkill3D-Bench}
\label{sec:evoskill3d-bench}

\paragraph{Task Definition.}
We formulate code-based 3D reconstruction as the generation of an executable modeling program conditioned on calibrated multiview observations. Given RGB images, their camera parameters, and a standardized instruction, an agent generates a complete Blender Python program to reconstruct the target in a shared world coordinate system. The program is executed to obtain a 3D model, which is then compared against reference geometry. Reference meshes, source metadata, and skill annotations are not provided to the agent.
For individual objects, we render four canonical orthographic views at azimuths of $45^\circ$, $135^\circ$, $225^\circ$, and $315^\circ$ and an elevation of approximately $30^\circ$, adding views when needed. For complex objects and scenes, overview views capture the overall structure and spatial arrangement, while close-up views emphasize geometric details. Human reviewers inspect the rendered observations for image quality and geometric coverage. Camera calibration specifies projection parameters and poses, and the common instruction defines the reconstruction objective and submission format.

\noindent\begin{minipage}{\textwidth}\textbf{Dataset Construction.}\end{minipage}\par\smallskip

\begin{wrapfigure}{l}{0.47\textwidth}
\vspace{-\intextsep}
\includegraphics[width=\linewidth]{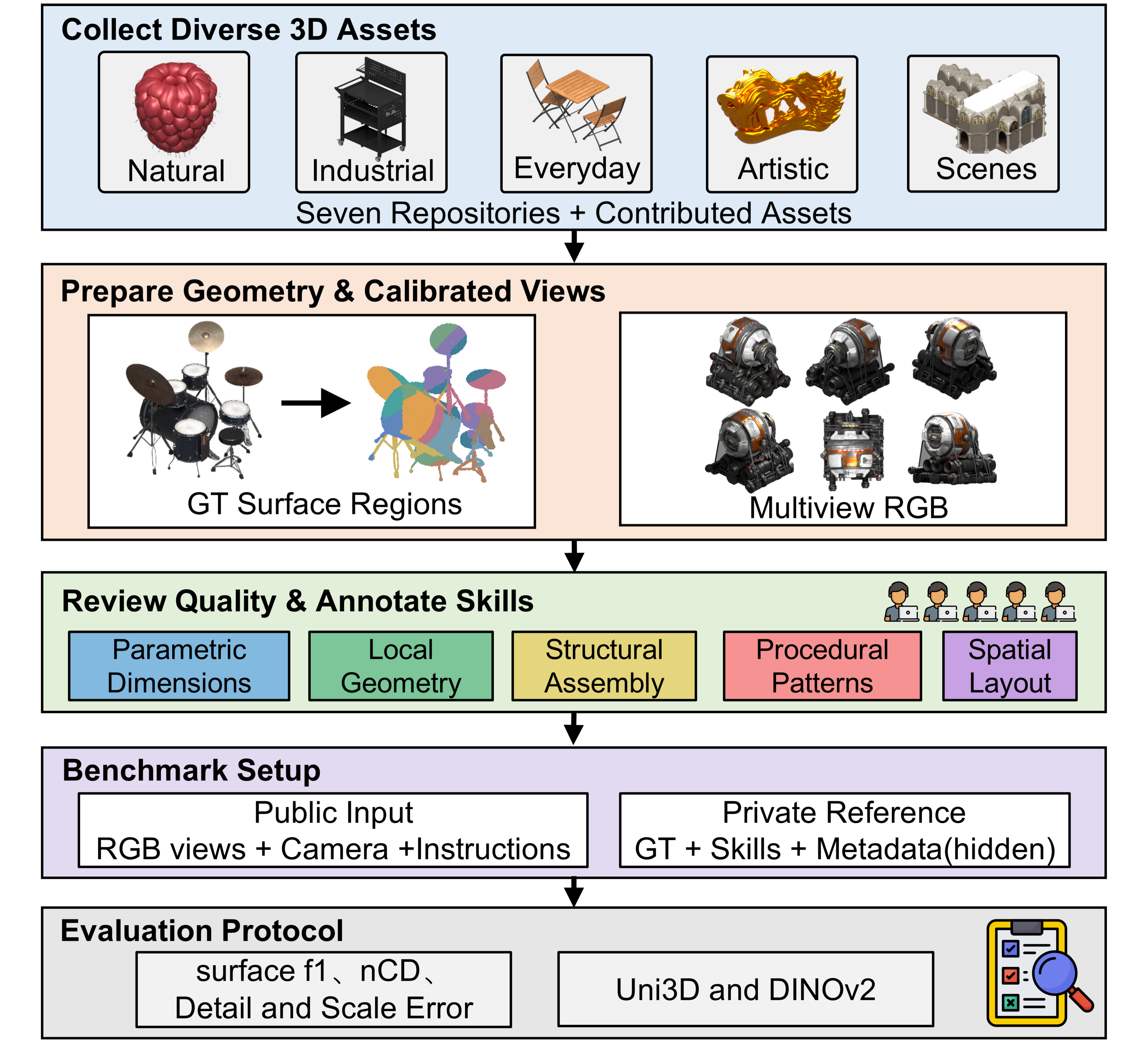}
\caption{\textbf{EvoSkill3D-Bench construction and evaluation.}
We collect 3D assets, create reference models and calibrated views, and have each asset reviewed and annotated by experts. Each task includes images, camera parameters, and instructions. The submitted Blender scripts are evaluated by an independent evaluator.}
\label{fig:data}
\vspace{-\intextsep}
\end{wrapfigure}

We curate 3D assets from existing repositories and contributed models, including Objaverse~\cite{deitke2023objaverse}, Poly Haven, the FreeCAD Parts Library, NVIDIA SimReady, Amazon Berkeley Objects~\cite{collins2022abo}, Google Scanned Objects~\cite{downs2022google}, and MuJoCo Menagerie~\cite{menagerie2022MuJoCo}.
As illustrated in Fig.~\ref{fig:data}, the construction pipeline covers asset collection, task generation, expert annotation, and geometric evaluation. The resulting benchmark contains 1,000 reconstruction tasks with 7,700 calibrated RGB views (2--20 per asset) spanning five content domains: \emph{Natural}, \emph{Industrial}, \emph{Everyday}, \emph{Artistic}, and \emph{Scenes}.

To assess asset quality and identify the primary modeling requirements, five experts with extensive experience in areas including mechanical design and artistic modeling review the candidates. We exclude any candidate receiving a negative quality vote. The remaining assets are organized by five foundational modeling skills: \emph{Parametric Dimensions}, concerning dimensions and proportions; \emph{Local Geometry}, covering curved surfaces, holes, bevels, and other details; \emph{Structural Assembly}, combining parts into coherent structures; \emph{Procedural Patterns}, constructing repeated, symmetric, or structured forms; and \emph{Spatial Layout}, describing object placement and spatial relations. Primary skill labels are assigned by plurality vote, with an independent reviewer resolving ties. These labels identify the dominant modeling requirement rather than all skills needed for reconstruction. We additionally construct a \emph{Hard} subset of complex assets involving multiple modeling operations.\footnote{
\href{https://objaverse.allenai.org/}{Objaverse};
\href{https://polyhaven.com/models}{Poly Haven};
\href{https://github.com/FreeCAD/FreeCAD-library}{FreeCAD Parts Library};
\href{https://huggingface.co/datasets/nvidia/simready-assets}{NVIDIA SimReady};
\href{https://registry.opendata.aws/amazon-berkeley-objects/}{Amazon Berkeley Objects};
\href{https://research.google/pubs/google-scanned-objects-a-high-quality-dataset-of-3d-scanned-household-items/}{Google Scanned Objects};
\href{https://github.com/google-deepmind/mujoco_menagerie}{MuJoCo Menagerie}.
}

For each asset, we extract the reference triangle mesh while preserving object transformations, modifiers, and instances. We then precompute area-weighted surface samples, sharp-edge descriptors, and multiscale curvature estimates with validity masks. To evaluate geometric details at different spatial scales, the reference surface is divided into hierarchical regions according to area and local geometric complexity. Automated validation further checks image integrity and geometric consistency of the references. Details of the region partitioning algorithm are provided in Supplementary Sec. A.2.

\paragraph{Evaluation Protocol.}
Our evaluator extracts the generated geometry and renders it under the recorded reference cameras and rendering settings, with separate resource budgets for extraction and rendering. We assess reconstruction quality along two complementary dimensions: \emph{geometric fidelity} and \emph{semantic similarity}.
Details of metrics are provided in the supplementary material (Sec. A.3).

\begin{itemize}
    \item \textbf{Geometric fidelity.}
We evaluate geometric accuracy using surface F1, normalized Chamfer-L1 distance (nCD), Detail, and Scale Error. Surface F1 measures bidirectional surface agreement within $1\%$ of the reference bounding-box diagonal, whereas nCD measures scale-normalized surface distance. Detail evaluates fine-grained geometry through sharp-edge agreement and multiscale curvature consistency, complementing the global surface metrics. Scale Error quantifies discrepancies in axis-aligned bounding-box dimensions.
    \item \textbf{Semantic similarity.}
We measure semantic consistency with pretrained Uni3D~\cite{zhou2024uni3d} and DINOv2~\cite{oquab2023dinov2}. Uni3D compares normalized point-cloud features of predicted and reference shapes, whereas DINOv2 measures feature similarity between reference images and corresponding reconstruction views.
\end{itemize}

In addition to these global scores, we report regional surface F1 and Detail over the hierarchical partitions, enabling more localized assessment and feedback for skill refinement. Results are averaged over valid, applicable reconstructions and reported alongside execution failure rates. Detailed metric definitions, region construction, and implementation settings are provided in the supplementary material.

\begin{figure*}
    \centering
    \includegraphics[width=\textwidth]{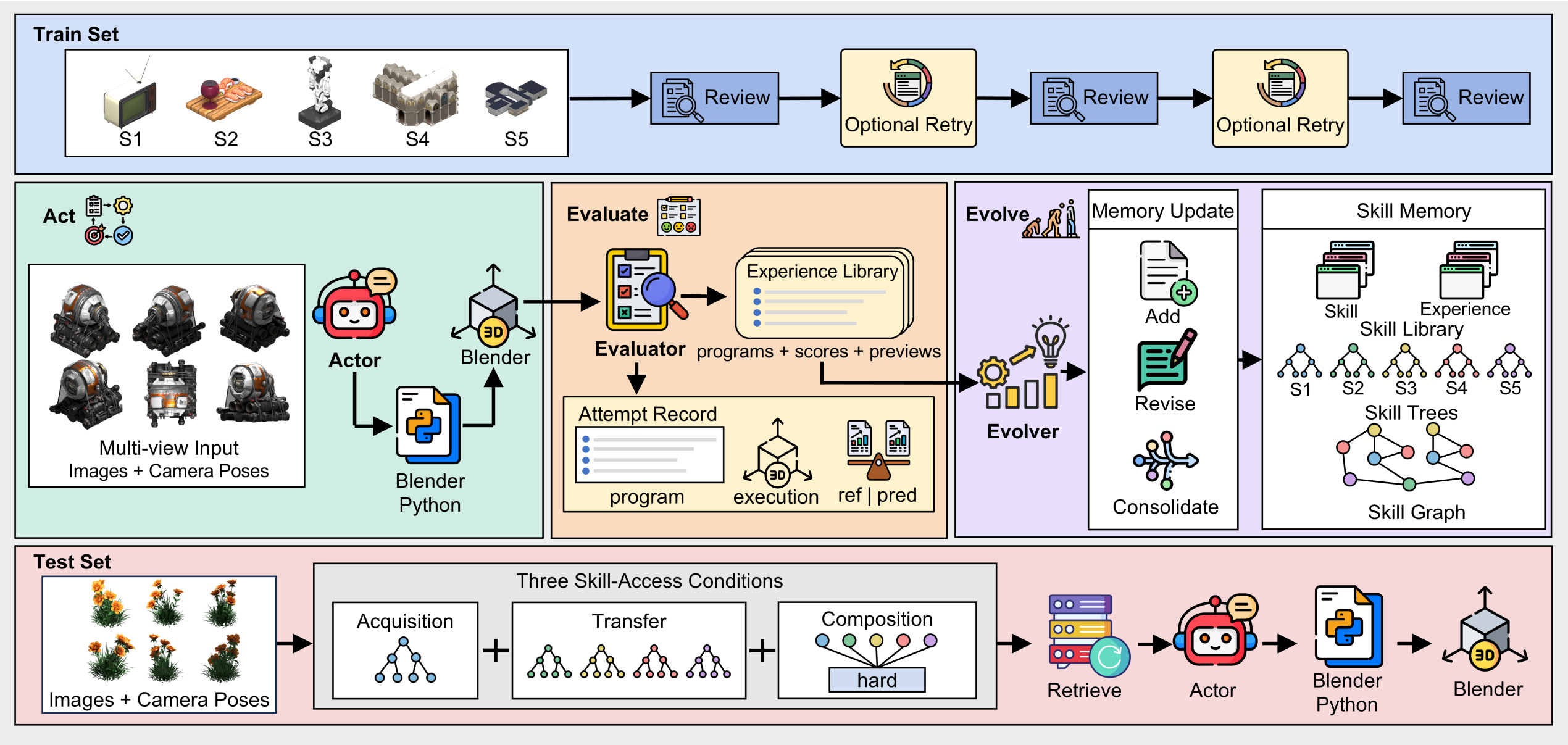}
    \caption{\textbf{Pipeline of EvoSkill3D-RSI framework.} During training, the Actor generates Blender programs from multiview images. The Evaluator executes these programs and provides scores, which the Evolver uses to update the skill memory. During testing, the memory is fixed and evaluated under three conditions in acquisition, transfer, and composition. For each task, the Actor retrieves the skills available in each setting and generates a Blender program.}
    \label{fig:fig2}
\end{figure*}
\subsection{EvoSkill3D-RSI}
\label{sec:evoskill3d-rsi}

\paragraph{Actor--Evaluator--Evolver Framework.}
As illustrated in Fig.~\ref{fig:fig2}, EvoSkill3D-RSI follows an iterative \emph{Actor}--\emph{Evaluator}--\emph{Evolver} workflow to learn reusable modeling procedures without modifying model weights. Training is organized into batches of five assets. For each task, the Evolver retrieves relevant procedures from the current skill memory based on public observations. The Actor uses these procedures to generate a Blender Python program, which the Evaluator executes to obtain execution feedback, six global metrics, regional surface and Detail scores, and rendered previews.

After the first reconstruction attempt on each of the five assets, the Evolver reviews the programs and their evaluation records, including reference and prediction previews. It can select up to two tasks for targeted refinement and permit at most two additional scored attempts for each selected task. Reviews follow each optional revision wave, giving at most three review points per batch and three scored attempts per task.
At each review, the Evolver either keeps the current memory or proposes new procedures, revisions, and consolidation of existing skills and associations. Proposed updates must be supported by training evidence and satisfy structural, Python syntax, dependency, and memory-budget checks before publication. The resulting memory version becomes available to subsequent tasks, allowing procedures learned from earlier reconstructions to inform later attempts. Training results use the final attempted reconstruction rather than the best intermediate score.

At test time, the learned memory is frozen and the Actor is evaluated under three skill-access conditions. The Actor can inspect the enabled skill catalog, Tree hierarchy, association Graph, and read-only skill files to select relevant procedures. Skill retrieval, tool interactions, and code generation share a budget of five normal model requests; at most three additional retries are allowed for API or formatting errors. Each task yields one Blender program for independent execution and scoring. Evaluation feedback remains hidden from the Actor, and neither reconstruction revision nor memory updates are permitted after submission.

\paragraph{Three-level Skill Memory.}
We organize the acquired modeling knowledge into three complementary layers. The \emph{Skill Library} maintains reusable procedures together with applicability conditions, counterexamples, executable code, and supporting training experiences. Five \emph{Skill Trees}, each rooted in one foundational category, organize procedures through parent--child relations. An association \emph{Graph} connects related skills through weighted, undirected edges supported by training evidence. Executable code dependencies are maintained separately as a directed acyclic graph to ensure consistent reuse.
Training begins with five category roots and shared foundational tools, but no learned skills. Acquired procedures may be added to any Tree during training. During evaluation, the available Trees are controlled by the skill-access protocol, while the foundational tools remain accessible in every condition.

\paragraph{Skill-centric Evaluation.}
We sample 20 assets from each of the five foundational categories for training, resulting in 100 training tasks. The remaining 900 assets, including the complete Hard subset, form the held-out test set. After training, we freeze the learned memory and evaluate skill acquisition, transfer, and composition under three access conditions. \emph{Acquisition} evaluates tasks from category $S_i$ with only its corresponding Tree enabled. \emph{Transfer} uses the same tasks but disables that Tree while enabling the other four, examining cross-category reuse. \emph{Composition} evaluates all 132 Hard assets with access to all five Trees to assess their joint use on complex reconstruction tasks. The frozen memory and foundational tools remain identical across the three conditions.

\section{Experiment}
\label{sec:Experiment}

In this section, we comprehensively evaluate
foundation models and coding agents on EvoSkill3D-Bench.
We first introduce the evaluated baselines
(Sec.~\ref{sec:evaluated_baselines}) and analyze their
reconstruction performance (Sec.~\ref{sec:main_results}).
We then investigate the effects of reasoning effort
(Sec.~\ref{sec:reasoning_ablation}) and the alignment
between automatic metrics and human judgments
(Sec.~\ref{sec:human_blind_ranking}).
Finally, we evaluate skill-centric recursive
self-improvement (Sec.~\ref{sec:rsi_evaluation})
and examine the cost--quality trade-off
(Sec.~\ref{sec:cost_quality}).

\begin{table*}[t]
\centering
\small
\setlength{\tabcolsep}{4.2pt}

% Morandi color palette
\definecolor{morandifirst}{RGB}{196,214,188}
\definecolor{morandisecond}{RGB}{224,216,198}
\definecolor{morandithird}{RGB}{214,198,192}
\definecolor{morandirank}{RGB}{132,169,149}

\newcommand{\first}[1]{\cellcolor{morandifirst}#1}
\newcommand{\second}[1]{\cellcolor{morandisecond}#1}
\newcommand{\third}[1]{\cellcolor{morandithird}#1}
\newcommand{\rankcell}[2]{%
  \cellcolor{morandirank!#1!white}\textbf{#2}}

\caption{\textbf{Main results on EvoSkill3D-Bench.}
For each metric, the
\protect\colorbox{morandifirst}{first},
\protect\colorbox{morandisecond}{second}, and
\protect\colorbox{morandithird}{third}
results are highlighted.
Overall ranks are determined by the mean ranks
across all nine metrics.}
\label{tab:main_results}

\resizebox{\textwidth}{!}{
\begin{tabular}{l cc cccccc cc}
\toprule
\multirow{2}{*}{\textbf{Model}}
& \multirow{2}{*}{\textbf{Rank}}
& \multirow{2}{*}{\textbf{Exec.} $\uparrow$}
& \multicolumn{6}{c}{\textbf{Geometry}}
& \multicolumn{2}{c}{\textbf{Semantic}} \\
\cmidrule(lr){4-9} \cmidrule(lr){10-11}
&  &  & \textbf{F1} $\uparrow$ & \textbf{Local-F1} $\uparrow$ & \textbf{Detail} $\uparrow$ & \textbf{Local-Detail} $\uparrow$ & \textbf{nCD} $\downarrow$ & \textbf{Scale(\%)} $\downarrow$ & \textbf{Uni3D} $\uparrow$ & \textbf{DINOv2} $\uparrow$ \\
\midrule

% Rank 1
\modellogo{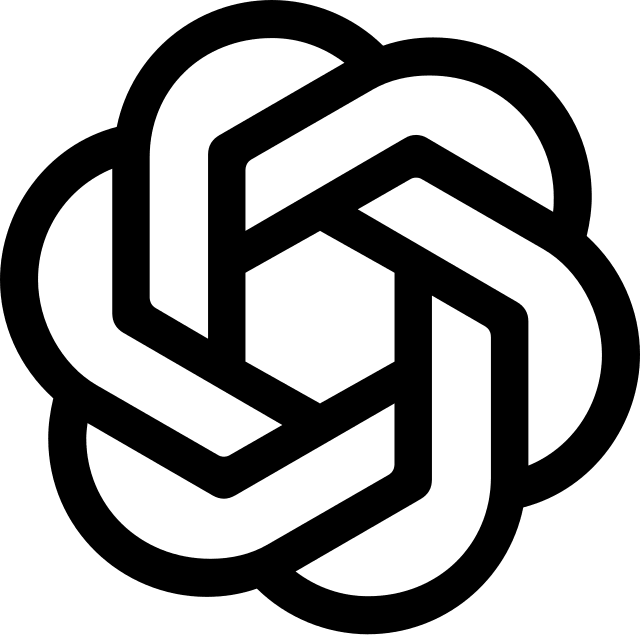}~GPT-6.1 Sol & \rankcell{90}{1} & \second{95.7} & \third{68.8} & \second{36.0} & \third{39.9} & \third{18.0} & \first{2.1} & \second{4.3} & \second{80.9} & \second{81.1} \\

% Rank 2
\modellogo{openai}~GPT-6 Astra & \rankcell{85}{2} & \first{98.3} & 68.4 & \third{35.6} & 39.4 & 17.8 & \third{2.3} & \third{4.5} & \first{81.1} & \third{80.9} \\

% Rank 3
\modellogo{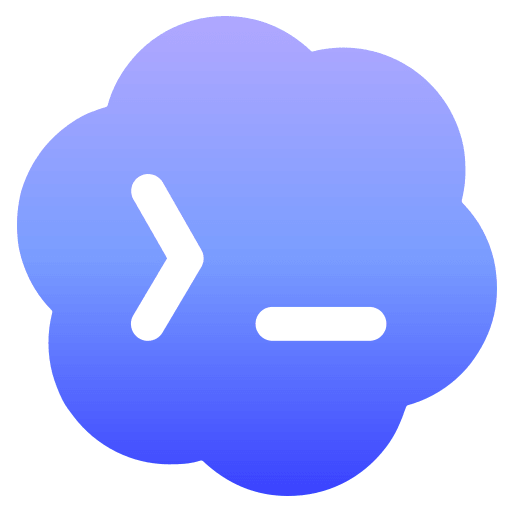}~Codex + GPT-6 Astra & \rankcell{80}{3} & 92.9 & \first{69.8} & \first{37.0} & \second{40.8} & \second{18.7} & 2.4 & 5.7 & \third{80.8} & \first{81.2} \\

% Rank 4
\modellogo{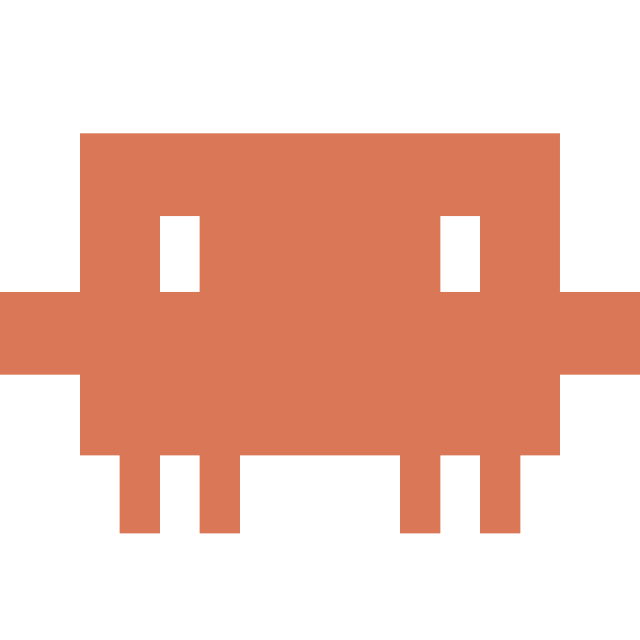}~Claude Code + Claude Opus 5.5 & \rankcell{75}{4} & 92.1 & \second{68.9} & 34.6 & \first{42.4} & \first{19.1} & \second{2.2} & \first{3.7} & 76.3 & 74.2 \\

% Rank 5
\modellogo{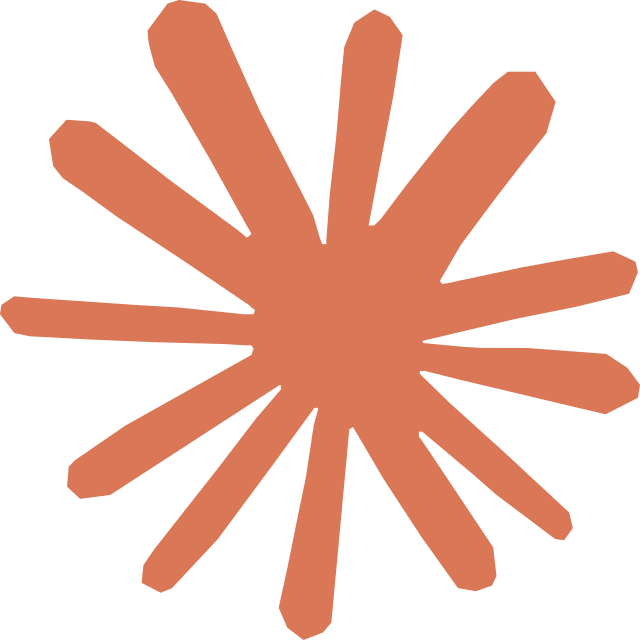}~Claude Opus 5.5 & \rankcell{70}{5} & 93.8 & 64.1 & 30.3 & 37.9 & 15.0 & 2.5 & 4.6 & 77.3 & 71.8 \\

% Rank 6
\modellogo{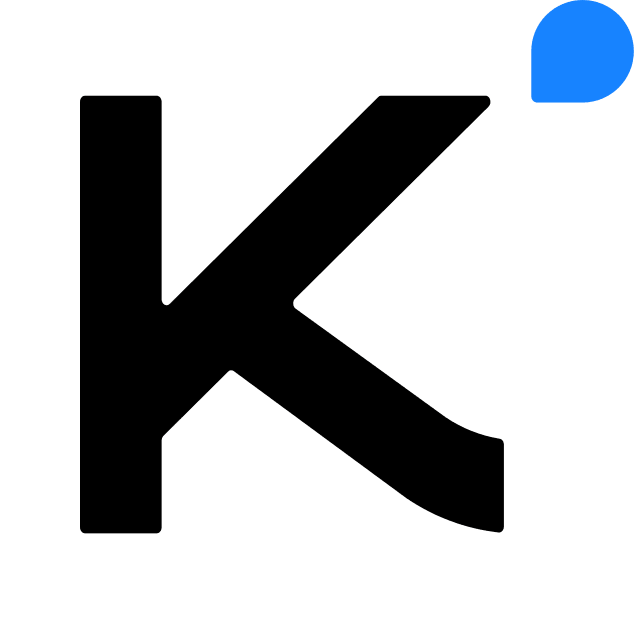}~Kimi Code + Kimi K3.1 Pro & \rankcell{65}{6} & 90.4 & 65.2 & 33.1 & 36.0 & 14.1 & 2.4 & 6.2 & 75.5 & 76.5 \\

% Rank 7
\modellogo{kimi}~Kimi K3.1 Pro & \rankcell{60}{7} & 78.6 & 64.1 & 32.0 & 35.3 & 14.7 & 2.7 & 5.3 & 77.2 & 77.5 \\

% Rank 8
\modellogo{claude}~Claude Fable 5.1 & \rankcell{55}{8} & \third{94.9} & 62.3 & 28.2 & 35.1 & 13.0 & 2.6 & 5.5 & 77.1 & 72.1 \\

% Rank 9
\modellogo{claude}~Claude Sonnet 5.5 & \rankcell{50}{9} & 94.0 & 59.0 & 26.8 & 32.8 & 12.2 & 3.0 & 6.0 & 75.1 & 69.5 \\

% Rank 10
\modellogo{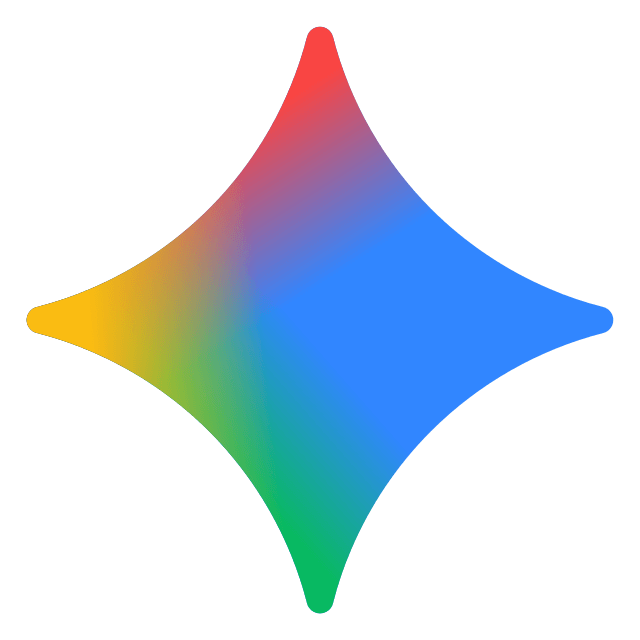}~Gemini-4 Argon & \rankcell{45}{10} & 72.9 & 54.2 & 25.0 & 28.7 & 10.9 & 3.7 & 10.0 & 79.9 & 77.4 \\

% Rank 11
\modellogo{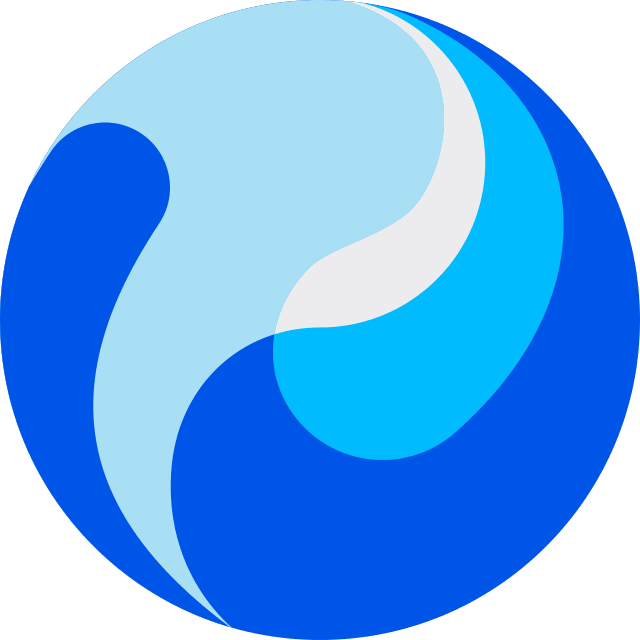}~Hy4 & \rankcell{40}{11} & 84.2 & 57.1 & 26.2 & 29.0 & 10.7 & 3.1 & 5.8 & 74.3 & 73.3 \\

% Rank 12
\modellogo{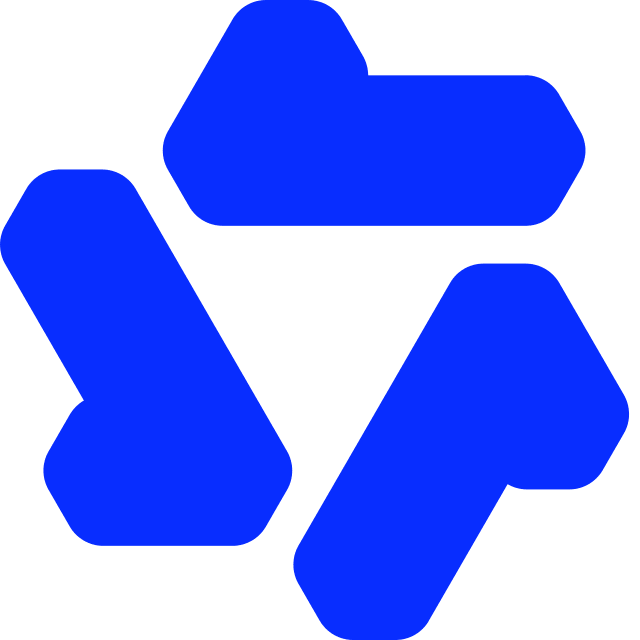}~Qwen3.8-Max & \rankcell{35}{12} & 78.9 & 49.8 & 21.2 & 24.0 & 7.8 & 4.3 & 10.9 & 71.9 & 66.1 \\

% Rank 13
\modellogo{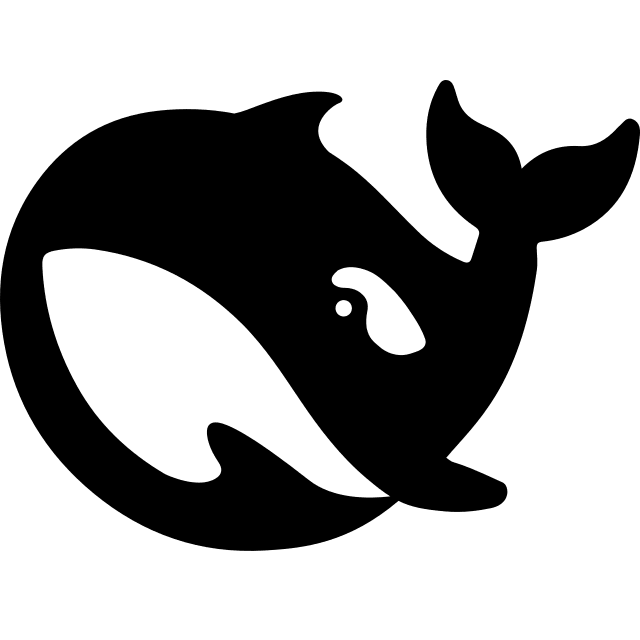}~DeepSeek Harness + DeepSeek V4.1 Flash & \rankcell{30}{13} & 88.6 & 44.6 & 19.0 & 20.2 & 6.5 & 5.2 & 14.8 & 69.5 & 67.8 \\

% Rank 14
\modellogo{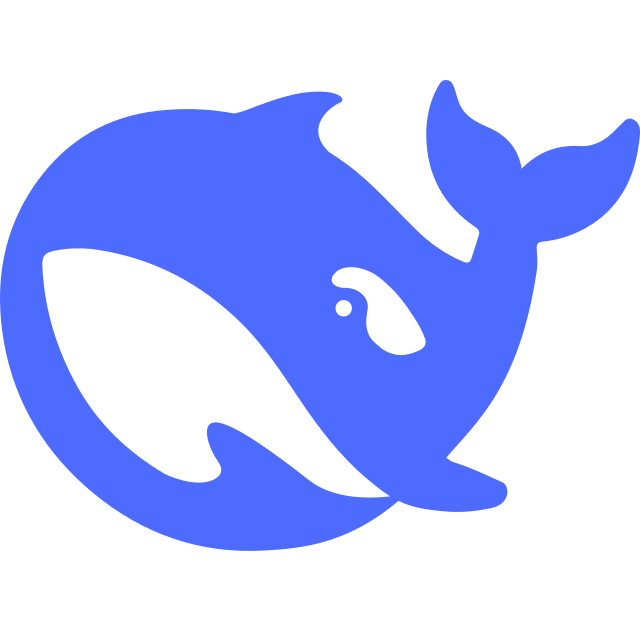}~DeepSeek V4.1 Flash & \rankcell{25}{14} & 84.5 & 42.7 & 17.3 & 18.3 & 5.4 & 4.6 & 12.9 & 69.8 & 68.5 \\

% Rank 15
\modellogo{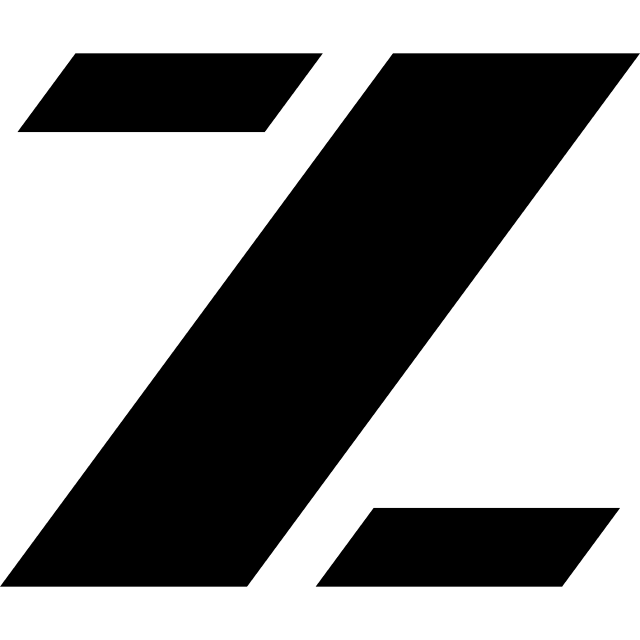}~GLM-5.3 Flash & \rankcell{20}{15} & 69.6 & 39.9 & 15.3 & 16.9 & 4.7 & 5.1 & 15.3 & 68.0 & 64.6 \\

% Rank 16
\modellogo{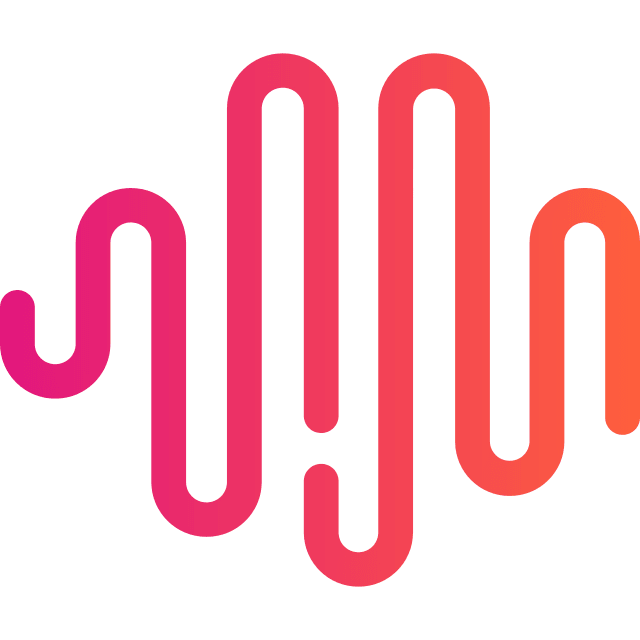}~MiniMax-M3.1-Flash & \rankcell{15}{16} & 66.5 & 37.7 & 15.2 & 16.1 & 4.8 & 131.1 & 565.7 & 66.8 & 64.7 \\

\bottomrule
\end{tabular}
}
\par\smallskip
\end{table*}

\begin{table*}[t]
\centering
\small
\setlength{\tabcolsep}{4.2pt}
\caption{\textbf{Ablation study on reasoning effort.}}
\label{tab:reasoning_ablation}
\resizebox{\textwidth}{!}{
\begin{tabular}{l c c cccccc cc}
\toprule
\multirow{2}{*}{\textbf{Model}}
& \multirow{2}{*}{\textbf{Reasoning Effort}}
& \multirow{2}{*}{\textbf{Exec.} $\uparrow$}
& \multicolumn{6}{c}{\textbf{Geometry}}
& \multicolumn{2}{c}{\textbf{Semantic}} \\
\cmidrule(lr){4-9} \cmidrule(lr){10-11}
&  &  & \textbf{F1} $\uparrow$ & \textbf{Local-F1} $\uparrow$ & \textbf{Detail} $\uparrow$ & \textbf{Local-Detail} $\uparrow$ & \shortstack{\textbf{nCD} $\downarrow$} & \textbf{Scale(\%)} $\downarrow$ & \shortstack{\textbf{Uni3D} $\uparrow$} & \shortstack{\textbf{DINOv2} $\uparrow$} \\
\midrule
\multirow{3}{*}{\modellogo{openai}~GPT-6.1 Sol} & High & 95.7 & 68.8 & 36.0 & 39.9 & 18.0 & 2.1 & 4.3 & 80.9 & 81.1 \\
  & Medium & 98.0 & 61.8 & 30.2 & 33.8 & 14.1 & 2.7 & 6.2 & 79.3 & 78.5 \\
  & Low & 99.2 & 56.9 & 27.2 & 29.8 & 12.0 & 3.2 & 8.1 & 77.4 & 74.9 \\
\midrule
\multirow{3}{*}{\modellogo{openai}~GPT-6 Astra} & High & 98.3 & 68.4 & 35.6 & 39.4 & 17.8 & 2.3 & 4.5 & 81.1 & 80.9 \\
  & Medium & 98.5 & 63.1 & 31.2 & 35.0 & 14.8 & 2.6 & 5.8 & 79.8 & 78.0 \\
  & Low & 98.9 & 61.6 & 30.2 & 33.6 & 14.2 & 2.7 & 6.3 & 79.1 & 76.7 \\
\midrule
\multirow{3}{*}{\modellogo{claude}~Claude Fable 5.1} & High & 94.9 & 62.3 & 28.2 & 35.1 & 13.0 & 2.6 & 5.5 & 77.1 & 72.1 \\
  & Medium & 97.7 & 54.5 & 23.1 & 28.5 & 9.6 & 3.3 & 8.1 & 73.9 & 67.0 \\
  & Low & 97.7 & 49.4 & 20.3 & 24.4 & 7.6 & 4.3 & 12.1 & 70.8 & 63.2 \\
\midrule
\multirow{3}{*}{\modellogo{deepseek}~DeepSeek V4.1 Flash} & High & 84.5 & 42.7 & 17.3 & 18.3 & 5.4 & 4.6 & 12.9 & 69.8 & 68.5 \\
  & Medium & 91.8 & 42.4 & 17.2 & 18.1 & 5.4 & 4.4 & 11.8 & 69.8 & 68.4 \\
  & Low & 92.8 & 40.6 & 16.3 & 17.0 & 4.9 & 4.9 & 14.5 & 68.9 & 67.2 \\
\midrule
\multirow{3}{*}{\modellogo{kimi}~Kimi K3.1 Pro} & High & 78.6 & 64.1 & 32.0 & 35.3 & 14.7 & 2.7 & 5.3 & 77.2 & 77.5 \\
  & Medium & 78.9 & 64.8 & 32.8 & 35.6 & 15.2 & 2.7 & 5.2 & 77.7 & 78.0 \\
  & Low & 84.8 & 51.4 & 23.0 & 25.4 & 9.2 & 4.0 & 11.0 & 73.6 & 74.2 \\
\bottomrule
\end{tabular}
}
\end{table*}

\subsection{Evaluated Baselines}
\label{sec:evaluated_baselines}

We evaluate 16 configurations on EvoSkill3D-Bench, including
12 foundation models and four coding agents. Specifically,
we consider GPT-6.1 Sol~\cite{openai2026gpt61sol}, GPT-6 Astra~\cite{openai2026gpt6astra}, Claude Fable 5.1~\cite{anthropic2026claudefable51}, Claude Opus 5.5~\cite{anthropic2026claudeopus55}, Claude Sonnet 5.5~\cite{anthropic2026claudesonnet55}, Gemini 4 Argon~\cite{deepmind2026gemini4argon}, Kimi K3.1 Pro~\cite{moonshot2026kimik31pro}, Hy4, Qwen3.8-Max~\cite{qwen38}, DeepSeek V4.1 Flash-1017~\cite{deepseekai2026deepseekv41flash}, GLM-5.3 Flash~\cite{glm5team2026glm5vibecodingagentic}, and MiniMax-M3.1-Flash~\cite{minimax2026m31flash}. We additionally evaluate four coding-agent configurations: Codex with GPT-6 Astra, Claude Code with Claude Opus 5.5, Kimi Code with Kimi K3.1 Pro, and DeepSeek Harness with DeepSeek V4.1 Flash-1017. All configurations are evaluated on the same benchmark tasks using identical evaluation metrics. Details of model integration are provided in Supplementary Sec.~B.

\subsection{Main Results}
\label{sec:main_results}
Table~\ref{tab:main_results} presents the reconstruction
performance of 16 model and agent configurations.
GPT-6.1 Sol achieves the highest overall rank, while
different systems exhibit distinct strengths across
evaluation dimensions. Codex with GPT-6 Astra achieves
the highest Surface F1 (69.8) and Local-F1 (37.0),
whereas Claude Code with Claude Opus 5.5 leads in
Detail (42.4) and Local-Detail (19.1).
This variation suggests that overall shape fidelity
and fine-grained geometric reconstruction do not
necessarily improve together. Moreover, high code
execution rates do not always translate into
geometrically accurate reconstructions.

A more prominent finding is the substantial gap between
global and local geometric metrics across all models.
For instance, GPT-6.1 Sol achieves 68.8 Surface F1
and 39.9 Detail, but only 36.0 Local-F1 and 18.0
Local-Detail. A similar pattern is observed for
Codex with GPT-6 Astra, whose scores drop from
69.8 to 37.0 in F1 and from 40.8 to 18.7 in Detail
under local evaluation.
Unlike global metrics, local evaluation applies
region-specific geometric measurements, exposing
discrepancies that may be obscured by aggregate scores.
These results suggest that current models can capture
the overall geometry of 3D objects reasonably well,
yet still struggle to preserve localized structures
and geometric details. Fine-grained reconstruction
therefore remains a key challenge even for
the strongest models. Detailed skill-wise results are provided in the supplementary material (Sec. E).

\subsection{Effect of Reasoning Effort}
\label{sec:reasoning_ablation}

Table~\ref{tab:reasoning_ablation} investigates the effect of
reasoning effort under a single-attempt setting without
execution feedback or retries. Increasing reasoning effort
generally improves both global and local geometric fidelity,
with larger relative gains in fine-grained metrics.
For GPT-6.1 Sol, moving from Low to High effort improves
Detail by 33.9\% and Local-Detail by 50.0\%.
A similar pattern is observed for Claude Fable 5.1,
with relative improvements of 43.9\% and 71.1\%,
respectively. These results suggest that additional
reasoning is particularly beneficial under more demanding
local geometric evaluation.
However, the benefits of increased reasoning vary
substantially across models. While GPT-6 Astra achieves
consistent quality gains with only a slight reduction
in execution success, DeepSeek V4.1 Flash exhibits
diminishing returns: moving from Medium to High effort
improves F1 by only 0.3 points while reducing execution
success by 7.3 percentage points. Moreover, Kimi K3.1 Pro
achieves better overall performance at Medium than High
effort. These findings reveal that reasoning capacity
and modeling reliability do not scale uniformly,
highlighting the importance of model-specific reasoning
budgets for code-based 3D reconstruction. Detailed results are provided in the supplementary material (Sec. E).

\begin{table*}[t]
\centering
\small
\begingroup
\setlength{\tabcolsep}{4.2pt}
\renewcommand{\arraystretch}{1.12}
\definecolor{rsistagegray}{RGB}{235,235,235}
\definecolor{rsired}{RGB}{225,65,65}
\definecolor{rsigreen}{RGB}{35,170,95}
\newcommand{\rsirise}[1]{\textsubscript{\textcolor{rsired}{\tiny #1}}}
\newcommand{\rsifall}[1]{\textsubscript{\textcolor{rsigreen}{\tiny #1}}}

\caption{\textbf{RSI results across Acquire, Transfer, and Composition.}
Subscripts indicate relative changes over the no-RSI Harness baseline,
with \textcolor{rsired}{red} and \textcolor{rsigreen}{green}
denoting improvements and regressions.}
\label{tab:rsi_results}

\resizebox{\textwidth}{!}{
\begin{tabular}{l cccccc cc}
\toprule
\multirow{2}{*}{\textbf{Model}}
& \multicolumn{6}{c}{\textbf{Geometry}}
& \multicolumn{2}{c}{\textbf{Semantic}} \\
\cmidrule(lr){2-7} \cmidrule(lr){8-9}
& \textbf{F1} $\uparrow$
& \textbf{Local-F1} $\uparrow$
& \textbf{Detail} $\uparrow$
& \textbf{Local-Detail} $\uparrow$
& \textbf{nCD} $\downarrow$
& \textbf{Scale(\%)} $\downarrow$
& \textbf{Uni3D} $\uparrow$
& \textbf{DINOv2} $\uparrow$ \\
\midrule

\rowcolor{rsistagegray}
\multicolumn{9}{c}{\textbf{Acquire}} \\
\modellogo{codex}~Codex + GPT-6 Astra
& 73.84\rsirise{+4.29}
& 41.48\rsirise{+7.51}
& 45.15\rsirise{+7.66}
& 22.04\rsirise{+12.44}
& 1.80\rsirise{+22.39}
& 4.00\rsirise{+30.99}
& 82.00\rsirise{+0.97}
& 82.85\rsirise{+1.05} \\
\modellogo{DSH}~DeepSeek Harness + DeepSeek V4.1 Flash
& 45.66\rsirise{+5.03}
& 20.35\rsirise{+7.52}
& 21.02\rsirise{+7.52}
& 7.40\rsirise{+17.65}
& 4.69\rsirise{+15.27}
& 13.52\rsirise{+19.01}
& 69.01\rsifall{-0.002}
& 66.98\rsifall{-0.39} \\
\midrule

\rowcolor{rsistagegray}
\multicolumn{9}{c}{\textbf{Transfer}} \\
\modellogo{codex}~Codex + GPT-6 Astra
& 73.82\rsirise{+4.73}
& 41.76\rsirise{+8.41}
& 45.00\rsirise{+8.69}
& 22.24\rsirise{+14.34}
& 1.83\rsirise{+19.60}
& 3.35\rsirise{+39.80}
& 81.87\rsirise{+0.96}
& 82.96\rsirise{+1.39} \\
\modellogo{DSH}~DeepSeek Harness + DeepSeek V4.1 Flash
& 46.60\rsirise{+3.86}
& 20.97\rsirise{+4.97}
& 21.52\rsirise{+4.21}
& 7.77\rsirise{+12.25}
& 4.80\rsirise{+10.62}
& 10.80\rsirise{+32.66}
& 69.14\rsifall{-1.36}
& 67.88\rsifall{-0.44} \\
\midrule

\rowcolor{rsistagegray}
\multicolumn{9}{c}{\textbf{Composition}} \\
\modellogo{codex}~Codex + GPT-6 Astra
& 64.43\rsirise{+10.36}
& 27.45\rsirise{+16.42}
& 33.84\rsirise{+15.31}
& 12.13\rsirise{+23.89}
& 2.44\rsirise{+24.73}
& 4.78\rsirise{+35.85}
& 78.84\rsirise{+1.84}
& 78.96\rsirise{+2.74} \\
\modellogo{DSH}~DeepSeek Harness + DeepSeek V4.1 Flash
& 41.65\rsifall{-1.12}
& 14.87\rsirise{+2.99}
& 17.80\rsifall{-0.76}
& 4.67\rsirise{+1.28}
& 5.38\rsirise{+6.45}
& 14.60\rsirise{+2.17}
& 67.15\rsirise{+0.33}
& 64.45\rsifall{-2.16} \\
\bottomrule
\end{tabular}
}
\endgroup
\end{table*}

\subsection{Human Blind Ranking}
\label{sec:human_blind_ranking}

We conduct a human blind ranking study to examine
the consistency between automatic evaluation metrics
and human judgments. Across 50 selected tasks,
reviewers rank five anonymized reconstructions against
the ground-truth 3D model using an interactive viewer,
considering overall shape, proportions, structural
completeness, and local geometric details.

\begin{wraptable}{l}{0.49\textwidth}
\vspace{-\intextsep}
\small
\setlength{\tabcolsep}{3.5pt}
% Morandi rank color (consistent with main table)
\definecolor{morandihumanrank}{RGB}{132,169,149}
\newcommand{\humanrankcell}[2]{%
  \cellcolor{morandihumanrank!#1!white}\textbf{#2}}
\caption{\textbf{Human blind ranking results and corresponding benchmark ranks.}
The results supports the alignment of our evaluation metrics
with human-perceived reconstruction quality.}
\label{tab:human_blind_ranking}
\resizebox{\linewidth}{!}{%
\begin{tabular}{@{}lcc@{}}
\toprule
\textbf{System} & \textbf{Human Rank} & \textbf{Benchmark Rank} \\
\midrule
\modellogo{codex}~Codex + GPT-6 Astra & 1.80 & \humanrankcell{80}{3} \\
\modellogo{claudecode}~Claude Code + Opus 5.5 & 2.65 & \humanrankcell{75}{4} \\
\modellogo{kimi}~Kimi K3.1 Pro & 2.81 & \humanrankcell{60}{7} \\
\modellogo{deepseek}~DeepSeek V4.1 Flash & 3.73 & \humanrankcell{25}{14} \\
\modellogo{minimax}~MiniMax M3.1 Flash & 4.00 & \humanrankcell{15}{16} \\
\bottomrule
\end{tabular}%
}
\vspace{-\intextsep}
\end{wraptable}
As shown in Table~\ref{tab:human_blind_ranking},
human preferences follow the same ordering as the
benchmark rankings across all five evaluated systems.
Notably, Codex with GPT-6 Astra achieves the highest
human preference, consistent with its strong geometric
reconstruction performance. Beyond system-level rankings,
automatic metrics also show meaningful correlations
with human judgments, with Spearman correlations of
0.651 for Surface F1, 0.683 for nCD, and 0.706 for DINOv2.
These findings suggest that our evaluation captures
important aspects of human-perceived reconstruction
quality, while providing more detailed geometric
measurements than overall human rankings.
Further details on the ranking protocol and
inter-reviewer agreement are provided in the
supplementary material. Details are provided in the supplementary material (Sec. A.5).

\subsection{Recursive Self-Improvement Evaluation}
\label{sec:rsi_evaluation}
Table~\ref{tab:rsi_results} evaluates skill-centric RSI
across Acquire, Transfer, and Composition against
the corresponding no-RSI Harness baselines.
Both systems achieve consistent geometric improvements
in Acquire and Transfer, suggesting that learned skills
can benefit unseen reconstruction tasks.
Notably, RSI yields substantial relative gains in
fine-grained geometry. For Codex with GPT-6 Astra,
Local-Detail improves by 12.44\%, 14.34\%, and
23.89\% across the three stages, respectively,
exceeding the corresponding relative gains in global
Detail. This highlights the potential of skill-based
refinement for improving localized geometric fidelity.
The two systems differ more substantially in
Composition. Codex achieves its largest gains in
this stage, improving F1 by 10.36\% and Detail
by 15.31\%. In contrast, DeepSeek Harness shows
slight regressions in both metrics, despite modest
improvements in local geometry and distance-based
metrics. These results demonstrate the effectiveness
of RSI while revealing model-dependent differences
in skill acquisition, transfer, and composition. Detailed results are provided in the Supplementary material (Sec. F).

\subsection{Cost--Quality Trade-off}
\label{sec:cost_quality}

\begin{wrapfigure}[13]{l}{0.47\textwidth}
\vspace{-\intextsep}
\includegraphics[width=\linewidth]{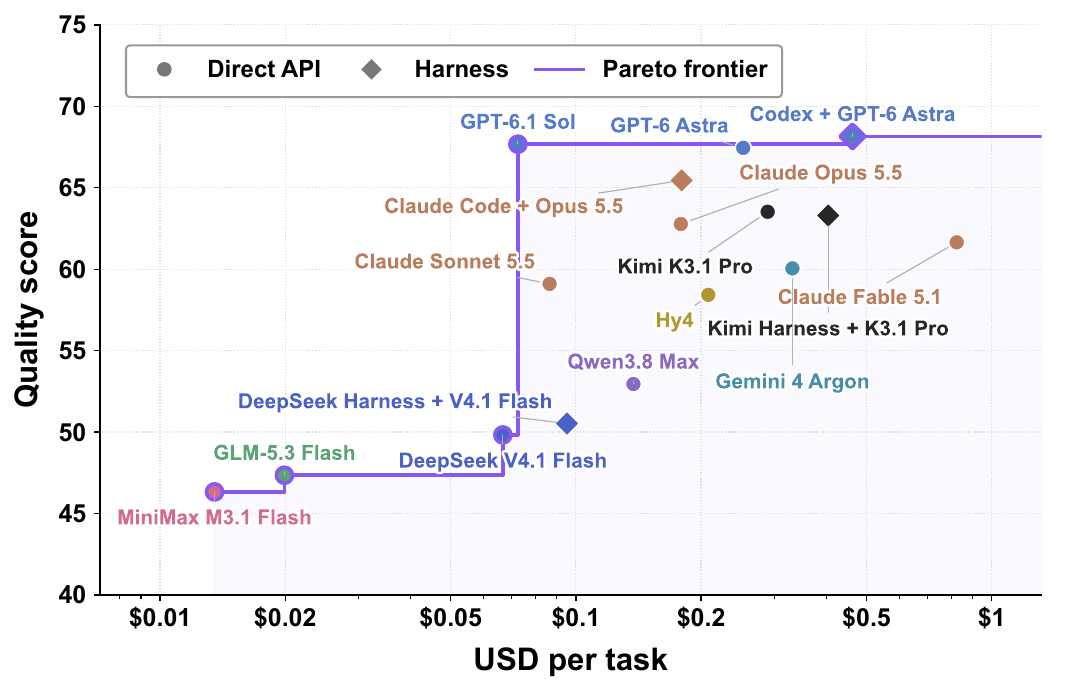}
\caption{\textbf{Cost--quality trade-off across foundation models and agents.} Pareto frontier indicates optimal configurations.}
\label{fig:cost_quality}
\vspace{-\intextsep}
\end{wrapfigure}
Figure~\ref{fig:cost_quality} examines the trade-off between
reconstruction quality and API token cost. We measure
overall quality by equally averaging Surface F1, Detail,
Uni3D, and DINOv2 similarities, while estimating costs
from recorded token usage and corresponding API pricing.
Coding Harnesses enable more extensive tool interactions
but do not necessarily improve cost efficiency.
For instance, Codex improves quality by only 0.7 points
over direct GPT-6 Astra while consuming approximately
4.9$\times$ more tokens. Most Harness configurations
fall below the global cost--quality Pareto frontier,
suggesting that additional agentic computation does
not always translate into quality gains. Details are provided in the Supplementary Sec.G.

\section{Conclusion}

We introduce \textbf{EvoSkill3D}, a skill-centric
benchmark and recursive self-improvement framework
for code-based 3D reconstruction agents.
EvoSkill3D-Bench comprises 1,000 reconstruction tasks
organized around five foundational modeling skills
and challenging multi-skill scenarios, with
fine-grained geometric evaluation grounded in
reference 3D assets. We further propose EvoSkill3D-RSI
to investigate skill acquisition, transfer, and
composition through structured feedback.
Extensive evaluations reveal a substantial gap
between global geometric fidelity and local detail,
while demonstrating the potential of recursive
self-improvement for enhancing reconstruction quality.
We hope EvoSkill3D will facilitate future research
on more capable and adaptive 3D coding agents.

\noindent\textbf{Limitations and Future Work.}
Our current benchmark focuses on Blender-based
reconstruction from calibrated images.
Future work will extend the evaluation to broader
3D creation tasks and explore long-term skill
evolution and cross-domain generalization.

\section{Authors}
\noindent\textbf{Core Authors:} Bolong Feng\textsuperscript{*,1}, Tianyu He\textsuperscript{*,2}, Pengfei Zhou\textsuperscript{*,3}, Yuanfan Yang\textsuperscript{5}, Dakai Wang\textsuperscript{5}, Wei Wang\textsuperscript{4}, Wangbo Zhao\textsuperscript{\(\dagger\),4}.\par\medskip

\noindent\textbf{Authors:} Zeyu Liang\textsuperscript{6}, Wenbo Yuan\textsuperscript{7}, Yihan Ma\textsuperscript{2}, Guanfei Duan\textsuperscript{2}, Kaixing Li\textsuperscript{8}, Ruizhi Qiu\textsuperscript{9}.\par\medskip

\noindent\textbf{Affiliations:} \textsuperscript{1} Shanghai Jiao Tong University \quad
\textsuperscript{2} South China University of Technology \quad
\textsuperscript{3} National University of Singapore \quad
\textsuperscript{4} The Hong Kong University of Science and Technology \quad
\textsuperscript{5} Anhui Scivisual Technology Co., Ltd. \quad
\textsuperscript{6} Jilin University \quad
\textsuperscript{7} China University of Mining Technology - Xuzhou \quad
\textsuperscript{8} Northeast Forest University \quad
\textsuperscript{9} Independent Researcher.\par\medskip

\noindent\textit{* Equal contribution; \(\dagger\) Corresponding author.}

{\small\bibliographystyle{ieeenat_fullname}\bibliography{reference}}
\clearpage
\appendix
\renewcommand{\topfraction}{0.95}
\renewcommand{\dbltopfraction}{0.95}
\renewcommand{\textfraction}{0.05}
\renewcommand{\floatpagefraction}{0.7}
\renewcommand{\dblfloatpagefraction}{0.7}
\setcounter{topnumber}{3}
\setcounter{dbltopnumber}{3}
\setcounter{totalnumber}{5}
\makeatletter
\setlength{\@dblfptop}{0pt}
\makeatother

\renewcommand{\thetable}{\arabic{table}}
\renewcommand{\thefigure}{\arabic{figure}}
\renewcommand{\theequation}{\arabic{equation}}
\newcommand{\NA}{\textnormal{N/A}}
\newcommand{\ind}{\mathbf{1}}

% RSI relative changes: red = improvement, green = degradation.
\definecolor{rsistagegray}{RGB}{235,235,235}
\definecolor{rsired}{RGB}{225,65,65}
\definecolor{rsigreen}{RGB}{35,170,95}
\newcommand{\rsirise}[1]{\textsubscript{\textcolor{rsired}{\tiny #1}}}
\newcommand{\rsifall}[1]{\textsubscript{\textcolor{rsigreen}{\tiny #1}}}

\lstset{basicstyle=\ttfamily\scriptsize,breaklines=true,columns=fullflexible,
  keepspaces=true,showstringspaces=false,frame=single,xleftmargin=2pt,xrightmargin=2pt}
\setcounter{figure}{0}
\setcounter{table}{0}
\setcounter{equation}{0}
\renewcommand{\thefigure}{\arabic{figure}}
\renewcommand{\thetable}{\arabic{table}}
\renewcommand{\theequation}{S\arabic{equation}}
\renewcommand{\theHfigure}{supp.\arabic{figure}}
\renewcommand{\theHtable}{supp.\arabic{table}}
\renewcommand{\theHequation}{supp.\arabic{equation}}
\pdfbookmark[0]{Supplementary Material}{supplement}
\section*{Supplementary Material}
\etocdepthtag.toc{supplement}
\etocsettagdepth{main}{none}
\etocsettagdepth{supplement}{subsection}
{\small\tableofcontents}
\clearpage
This supplement details benchmark construction and geometric feedback
(Sec.~\ref{supp:benchmark}), model integration
(Sec.~\ref{supp:integration}), the recursive skill-improvement process
(Sec.~\ref{supp:rsi}), and qualitative examples from saved experimental
artifacts (Sec.~\ref{supp:cases}), followed by results across modeling
skills and reasoning efforts (Sec.~\ref{supp:subset-results}) and RSI
results by modeling skill (Sec.~\ref{supp:rsi-subset-results}).
The configurations below specify the current settings of each execution
mode; model requests, reconstruction attempts, and evaluator recovery
attempts have separate budgets.

\section{Benchmark Construction and Evaluation}
\label{supp:benchmark}

\subsection{Asset curation and task packaging}

EvoSkill3D-Bench comprises 1,000 reconstruction tasks built around 1,000 assets and 7,700 calibrated RGB views.
The assets cover Natural, Industrial, Everyday, Artistic, and Scenes
content, spanning individual objects, articulated assemblies, repeated
structures, and multi-object environments.
Sources include Objaverse, Poly Haven, the FreeCAD Parts Library,
NVIDIA SimReady, Amazon Berkeley Objects, Google Scanned Objects,
MuJoCo Menagerie, and contributed assets.
We preserve per-asset attribution and licensing information.
The five content domains describe semantic coverage; the benchmark's
skill categories describe the modeling operations emphasized by an asset.

Five human experts with extensive 3D modeling experience across domains,
including mechanical design and artistic modeling, assessed candidate
asset quality and voted for each asset's most salient modeling skill.
The five foundational categories are S1: Parametric Dimensions,
S2: Local Geometry, S3: Structural Assembly,
S4: Procedural Patterns, and S5: Spatial Layout.
These primary labels do not imply that reconstruction requires only one
skill. A separate HardSet contains complex reconstruction tasks.
The generated task packages also underwent manual review and automated
structural checks.

\begin{table}[t]
\centering
\small
\begingroup
\setlength{\tabcolsep}{4.2pt}
\renewcommand{\arraystretch}{1.12}
\caption{Assets, RSI splits, and calibrated RGB views by modeling skill.}
\label{supp:tab:split}
\begin{tabularx}{\columnwidth}{Xrrrr}
\toprule
\textbf{Subset} & \textbf{Train} & \textbf{Test} & \textbf{Total} & \textbf{Views} \\
\midrule
S1: Parametric Dimensions & 20 & 149 & 169 & 1,147 \\
S2: Local Geometry & 20 & 236 & 256 & 1,794 \\
S3: Structural Assembly & 20 & 142 & 162 & 1,138 \\
S4: Procedural Patterns & 20 & 110 & 130 & 969 \\
S5: Spatial Layout & 20 & 131 & 151 & 1,552 \\
HardSet & 0 & 132 & 132 & 1,100 \\
\midrule
Total & 100 & 900 & 1,000 & 7,700 \\
\bottomrule
\end{tabularx}
\endgroup
\end{table}

\begin{figure}[t]
\centering
\includegraphics[width=\columnwidth]{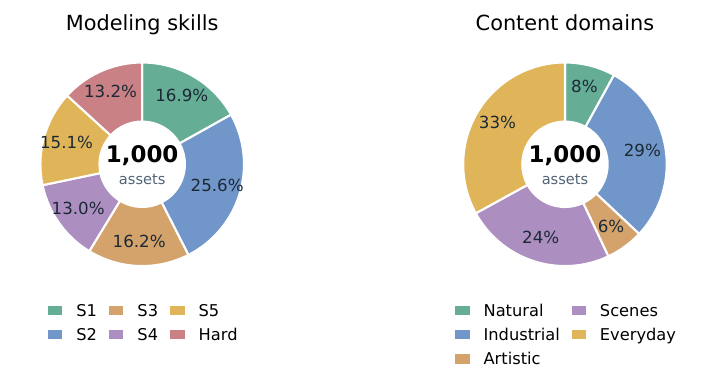}
\caption{Modeling-skill and content-domain composition of EvoSkill3D-Bench.}
\label{supp:fig:domains}
\end{figure}

Figure~\ref{supp:fig:domains} summarizes the six skill subsets and five content domains:
Natural (8\%), Industrial (29\%), Artistic (6\%), Scenes (24\%),
and Everyday (33\%).

Each task exposes RGB images, camera calibration, and a standardized
reconstruction instruction. The instruction is shared across assets;
asset-specific evidence comes from the images and cameras.
Camera metadata specifies resolution, projection, camera-to-world pose,
and orthographic mappings or perspective intrinsics as appropriate.
Coordinates are expressed in meters in the supplied Blender world frame.
The reference scale follows the source metadata.
Views include overall observations and close-ups where available.
The evaluator privately retains the world-space triangle mesh,
100,000 area-uniform surface samples with normals and integration
weights, a spatial partition, sharp-edge segments, multiscale curvature,
and frozen semantic features.
The public inputs exclude reference meshes, asset names, source metadata,
and human skill labels.
For RSI, we use the fixed partition in Table~\ref{supp:tab:split}.
Acquisition and Transfer each evaluate the 768 S1--S5 test tasks;
Composition evaluates the 132 HardSet tasks.

\subsection{Density-adaptive GT partitioning}
\label{supp:partition}

\paragraph{Surface-based geometric density.}
The partition adapts to local geometric detail using an area-uniform
surface probe, so a finely tessellated but flat patch does not receive
extra weight merely because it has more vertices.
Let $\mathcal{G}$ be the reference surface, $A$ its area, and
$D=\|\boldsymbol{b}_{\max}-\boldsymbol{b}_{\min}\|_2$ its bounding-box
diagonal. We sample $N=100{,}000$ points
$\{(\boldsymbol{x}_i,\boldsymbol{n}_i,w_i)\}_{i=1}^N$ with
$w_i=A/N$, where $\boldsymbol{n}_i$ is the source triangle normal.
We probe radii $\mathcal{H}=\{0.005D,0.01D,0.02D\}$.
For each point, we query its 32 nearest surface samples and retain those
within radius $h$. A radius is supported when at least 12 neighbors
remain. Its normal-variation estimate is
\begin{equation}
 v_i(h)=\frac{1}{|\mathcal{N}_i(h)|}
 \sum_{j\in\mathcal{N}_i(h)}
 \left(1-|\boldsymbol{n}_i^\top\boldsymbol{n}_j|\right).
\end{equation}
We set the detail strength $s_i$ to the maximum supported variation and
$h_i^*$ to the smallest supported radius.
Points within $0.005D$ of a reliable sharp edge receive $s_i=1$ and
$h_i^*=0.005D$. A point without a supported scale remains explicitly
unresolved.

Let $m_i$ indicate whether a scale is supported, including the sharp-edge
assignment above. The dimensionless geometric burden is
\begin{equation}
 q_i=\frac{w_i}{D^2}+\beta\,w_i m_i\frac{s_i}{(h_i^*)^2}.
 \label{supp:eq:burden}
\end{equation}
Here $\beta=10^{-4}$, and the second term is zero if $m_i=0$.
The first term accounts for surface area, while the second increases
the subdivision priority of detailed regions at small supported scales.
The corresponding local density is $q_i/w_i$.
The normal-variation probe determines partition density; the quadratic
curvature estimator described below determines a component of the
Detail metric.

\paragraph{Burden-prioritized subdivision.}
We initialize a binary spatial tree with all surface samples at the
root. At each step we select the open leaf with the largest burden
$Q(R)=\sum_{i\in R}q_i$, resolving equal priorities by node ID.
If $Q(R)\leq Q_{\max}$, the leaf stops.
Otherwise, each nondegenerate world axis proposes cut planes at the
35th, 50th, and 65th percentiles of cumulative burden.
Cuts are placed between distinct coordinates, using a numerical
coordinate tolerance $64\epsilon D$.

Each candidate must give both children at least 128 surface samples,
$10^{-5}$ of the total reference area, and 10\% of the parent's burden.
We discourage cutting through detailed neighborhoods with
\begin{equation}
 B(R,a,c)=
 \frac{\sum_{i\in R}w_i s_i\,
 \ind[i\text{ supported}]\,
 \ind[|x_{i,a}-c|<h_i^*]}
 {\sum_{i\in R}w_i s_i}.
 \label{supp:eq:boundary}
\end{equation}
The cost is zero when the denominator is zero.
Among candidates with $B\leq B_{\max}$, we choose the one with the lowest
$(B,a,c)$ in lexicographic order.
Points with $x_{i,a}<c$ enter the left child; the remaining points enter
the right child.

\begin{table}[t]
\centering
\small
\begingroup
\setlength{\tabcolsep}{4.2pt}
\renewcommand{\arraystretch}{1.12}
\caption{Parameters for density-adaptive GT partitioning.}
\label{supp:tab:partition}
\begin{tabularx}{\columnwidth}{Xr}
\toprule
\textbf{Construction parameter} & \textbf{Value} \\
\midrule
Surface probes / random seed & 100,000 / 20261029 \\
Normal-neighbor query / minimum support & 32 / 12 \\
Probe radii divided by $D$ & 0.005, 0.01, 0.02 \\
Sharp-angle threshold / analysis weld grid & $30^\circ$ / $10^{-8}D$ \\
Detail coefficient $\beta$ & $10^{-4}$ \\
Leaf burden threshold $Q_{\max}$ & 0.25 \\
Boundary-cost threshold $B_{\max}$ & 0.20 \\
Burden quantiles for cuts & 0.35, 0.50, 0.65 \\
Minimum samples per child & 128 \\
Minimum area fraction per child & $10^{-5}$ \\
Minimum child / parent burden & 0.10 \\
Maximum leaf count $K_{\mathrm{cap}}$ & 256 \\
\bottomrule
\end{tabularx}
\endgroup
\end{table}

\begin{figure*}[tp]
\centering
\includegraphics[width=\textwidth,height=0.42\textheight,keepaspectratio]{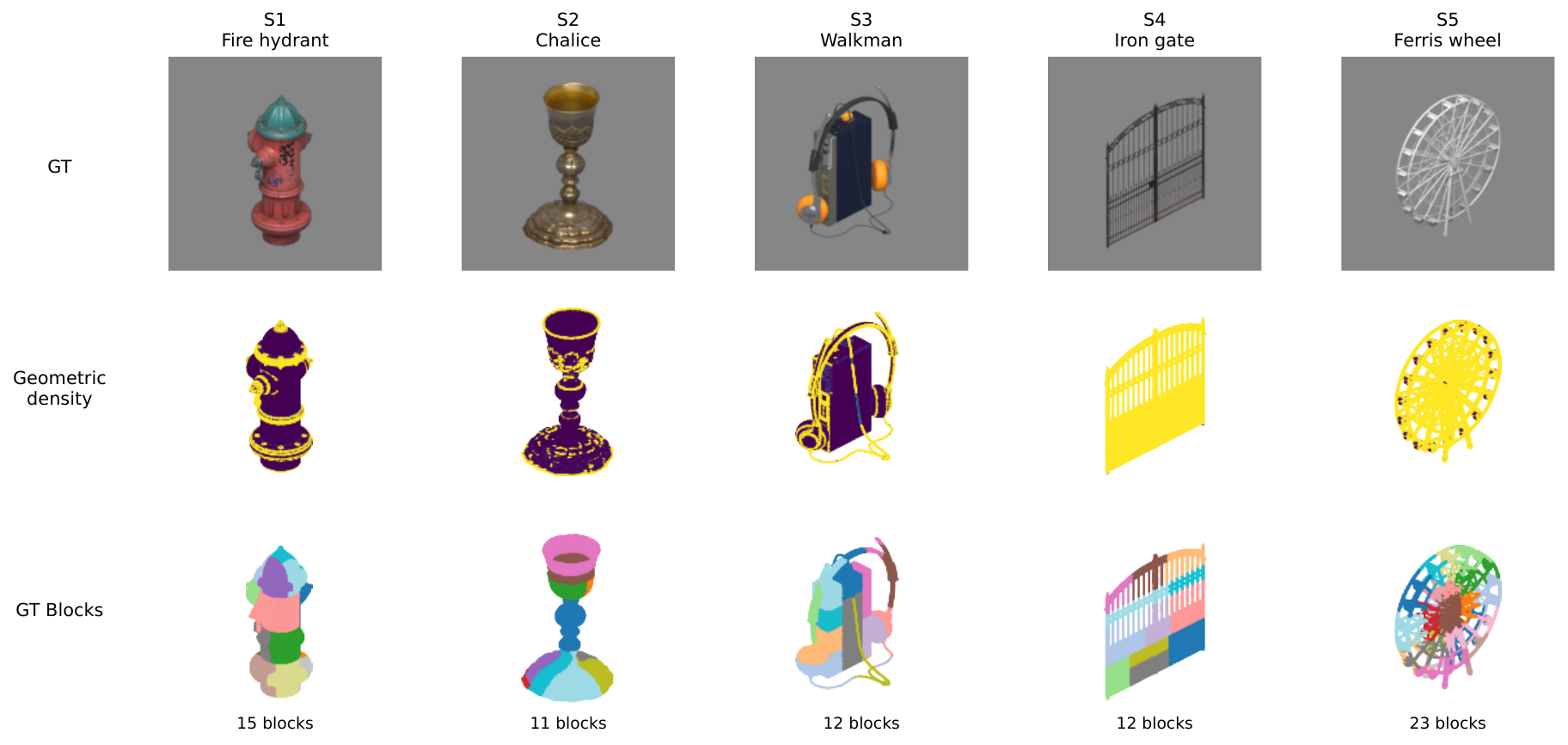}
\caption{GT appearance, geometric density, and adaptive blocks for five assets.}
\label{supp:fig:partition}
\end{figure*}

A leaf also stops when the region cap is reached or no admissible cut
exists. We distinguish burden-satisfied, resource-limited, and
unresolved leaves; missing detail support is not treated as evidence
that a region contains no detail.
Table~\ref{supp:tab:partition} lists the concrete parameters.
The five illustrated assets produce the block counts shown in
Fig.~\ref{supp:fig:partition}.

\paragraph{Exact triangle fragments.}
The spatial cuts define geometric regions rather than semantic parts.
An uncut triangle stores one region ID.
A triangle crossing a plane is clipped recursively, and the resulting
polygons are triangulated into fragments whose vertices are stored in
barycentric coordinates relative to the original triangle.
This preserves the reference surface and supports exact regional
bounds and area calculations without simplifying the mesh.
The builder checks that fragment and uncut-triangle areas sum to the
original surface area, using relative tolerance $10^{-9}$ and absolute
tolerance $10^{-12}A$.
The evaluator reuses this frozen partition for every prediction of the
same task.

\subsection{Evaluation Metrics}
\label{supp:metrics}

\paragraph{Shared alignment and distance queries.}
Let $\mathcal{P}$ be the predicted world-space mesh.
We translate it by the difference between the reference and prediction
bounding-box centers. We do not fit rotation, scale, or independent
object transforms.
Let $\{\boldsymbol{y}_j,u_j\}_{j=1}^{N}$ be 100,000 area-uniform
prediction samples, with $u_j=A_{\mathcal{P}}/N$.
Distances are Euclidean distances to the complete opposite triangle
surface:
\begin{equation}
 d_i^{G}=d(\boldsymbol{x}_i,\mathcal{P}),\qquad
 d_j^{P}=d(\boldsymbol{y}_j,\mathcal{G}).
\end{equation}
An exact triangle-distance implementation accelerated by a bounding
volume hierarchy supplies these queries.
The integration uses sampled surface points, while their distances
are measured against triangles.

\paragraph{Surface F1 ($\uparrow$).}
At $\tau=0.01D$, area-weighted precision and recall are
\begin{align}
 P_\tau&=\frac{\sum_j u_j\ind[d_j^{P}\leq\tau]}{\sum_j u_j},&
 R_\tau&=\frac{\sum_i w_i\ind[d_i^{G}\leq\tau]}{\sum_i w_i},\\
 F_\tau&=\frac{2P_\tau R_\tau}{P_\tau+R_\tau}.
\end{align}
The table entry is $\operatorname{Surface\mbox{-}F1}=100F_\tau$.
The F1 score is zero when both precision and recall are zero.
We uses the 1\% threshold.

\paragraph{Normalized Chamfer-L1 ($\downarrow$).}
We report
\begin{equation}
 \operatorname{nCD}=\frac{100}{D}\left(
 \frac{\sum_i w_i d_i^{G}}{\sum_i w_i}+
 \frac{\sum_j u_j d_j^{P}}{\sum_j u_j}\right).
 \label{supp:eq:chamfer}
\end{equation}

\paragraph{Scale error ($\downarrow$).}
Let $g_a$ and $p_a$ be the reference and prediction bounding-box
extents along world axis $a$. With
$\mathcal{A}=\{a:g_a>10^{-12}D\}$, we compute
\begin{equation}
 e_{\mathrm{scale}}=\frac{100}{|\mathcal{A}|}
 \sum_{a\in\mathcal{A}}\frac{|p_a-g_a|}{g_a}.
\end{equation}

\paragraph{Sharp-edge F1.}
On an analysis copy of each mesh, vertices are welded on a
$10^{-8}D$ grid and duplicate faces are removed for adjacency analysis.
An edge is reliable when exactly two consistently wound manifold faces
meet there. It is sharp when their normal angle exceeds $30^\circ$.

We sample 32,768 points along each nonempty sharp-edge set with
probability proportional to segment length.
Let $\mathcal{L}_G$ and $\mathcal{L}_P$ be the GT and predicted
sharp-edge segment sets, and let $\boldsymbol{e}_i^G$ and
$\boldsymbol{e}_j^P$ denote their length-uniform samples.
At $\tau_E=0.01D$, the edge precision, recall, and F1
\begin{align}
 P_E&=\frac{1}{M_P}\sum_{j=1}^{M_P}
 \ind[d(\boldsymbol{e}_j^P,\mathcal{L}_G)\leq\tau_E],\\
 R_E&=\frac{1}{M_G}\sum_{i=1}^{M_G}
 \ind[d(\boldsymbol{e}_i^G,\mathcal{L}_P)\leq\tau_E],\\
 E&=\frac{2P_ER_E}{P_E+R_E}.
 \label{supp:eq:edge}
\end{align}
Here $M_G=M_P=32{,}768$ for nonempty edge sets, and $E=0$
when $P_E+R_E=0$.
If the reference has sharp features but the prediction has
none, the edge F1 is zero.
If the reference has no reliable sharp features, this component is
inapplicable.

\paragraph{Multiscale curvature fitting.}
At the stored reference anchors we evaluate mean curvature $H$ and
Gaussian curvature $K$ at radii $h\in\mathcal{H}$.
The support is an ephemeral, deterministic sample of one million
surface points. Within each radius, neighborhoods remain on the
connected smooth sheet reached through consistently wound shared edges
below the sharp-angle threshold.
Connectivity is evaluated locally inside the anchor's ball, avoiding
shortcuts to nearby surface layers.
Anchors within $0.005D$ of a sharp edge are excluded from curvature
fitting.

We form tangent coordinates $(\xi,\eta,\zeta)$ after division by $h$
and fit
\begin{equation}
 \zeta=\tfrac{a}{2}\xi^2+b\xi\eta+
 \tfrac{c}{2}\eta^2+d\xi+e\eta+f.
\end{equation}
The fitted curvatures are
\begin{align}
 H&=\frac{(1+e^2)a-2deb+(1+d^2)c}
 {2(1+d^2+e^2)^{3/2}h},\\
 K&=\frac{ac-b^2}{(1+d^2+e^2)^2h^2}.
\end{align}
A valid fit requires at least 12 neighbors, rank six, design-matrix
condition number at most $10^4$, normalized RMS residual at most 0.05,
and mean neighbor-normal agreement at least 0.9.
At most 256 neighbors are retained in stable distance order with
evenly spaced selections when the neighborhood is larger.
Validity masks are stored with the GT values.

For a prediction, each GT anchor first queries its nearest predicted
surface point. A correspondence is accepted only within $0.01D$ and
with absolute normal dot product at least 0.9.
Prediction-side sharp neighborhoods and invalid fits remain
unmatched. When a candidate fit lacks the minimum support, one bounded
refill draws at most 256 additional points from connected smooth-sheet
triangles clipped to the radius bounding cube, followed by sphere
rejection. Refill is skipped above 4,096 candidate faces.

\paragraph{Coverage-sensitive curvature similarity.}
For a paired-valid anchor $i$ at radius $h$, define
\begin{align}
 e_{H,i,h}&=h\left||H^P_{i,h}|-|H^G_{i,h}|\right|,\\
 e_{K,i,h}&=h^2|K^P_{i,h}-K^G_{i,h}|,\\
 q_{i,h}&=\frac{1}{1+e_{H,i,h}+\sqrt{e_{K,i,h}}}.
\end{align}
Let $\mathcal{E}_h$ be the GT-eligible anchors and
$\mathcal{V}_h\subseteq\mathcal{E}_h$ the paired-valid anchors.
We compute
\begin{equation}
 C_h=\frac{\sum_{i\in\mathcal{V}_h}w_iq_{i,h}}
 {\sum_{i\in\mathcal{E}_h}w_i},\qquad
 C=\frac{1}{|\mathcal{H}_{\mathrm{app}}|}
 \sum_{h\in\mathcal{H}_{\mathrm{app}}}C_h,
 \label{supp:eq:curvature}
\end{equation}
where $\mathcal{H}_{\mathrm{app}}$ includes scales with eligible GT
anchors.
We also retain the coverage
$\sum_{\mathcal{V}_h}w_i/\sum_{\mathcal{E}_h}w_i$ and conditional
paired-point errors to distinguish agreement from correspondence loss.

\paragraph{Detail ($\uparrow$).}
When both components apply, the metric is
\begin{equation}
 \operatorname{Detail}=100\sqrt{EC},
\end{equation}
where $E$ is the edge F1 in Eq.~\ref{supp:eq:edge} and $C$ is the
coverage-sensitive curvature score in Eq.~\ref{supp:eq:curvature}.
The components $E$, $C$, and $C_h$ lie in $[0,1]$; the combined
table entry is on the $[0,100]$ scale.

\paragraph{Uni3D similarity ($\uparrow$).}
We use the frozen Uni3D-G checkpoint.
From each shape's deterministic 100,000-point surface sample,
8,192 points are selected without replacement using seed 42.
The point cloud is mean-centered and divided by its maximum radius;
each RGB channel is set to 0.4.
We L2-normalize the frozen feature vector and compute its cosine
similarity to the cached reference feature:
\begin{equation}
 S_{\mathrm{Uni3D}}=100\,
 \frac{\boldsymbol{f}_G^\top\boldsymbol{f}_P}
 {\|\boldsymbol{f}_G\|_2\|\boldsymbol{f}_P\|_2}.
\end{equation}
This encoder normalization makes the feature comparison complementary
to explicit metric-scale geometry measurements.

\paragraph{DINOv2 similarity ($\uparrow$).}
We use \texttt{facebook/\allowbreak dinov2-large}, with its fixed image
processor and model pooler output.
For each of the $V$ recorded cameras, the evaluator renders the
prediction using the corresponding reference lighting and render
profile, preserving the predicted materials.
We normalize features and average same-view cosine similarities:
\begin{equation}
 S_{\mathrm{DINO}}=\frac{100}{V}\sum_{v=1}^V
 \frac{\boldsymbol{g}_{G,v}^\top\boldsymbol{g}_{P,v}}
 {\|\boldsymbol{g}_{G,v}\|_2\|\boldsymbol{g}_{P,v}\|_2}.
\end{equation}
Both semantic scores multiply cosine similarity by 100.

\paragraph{Prediction assignment and regional tolerances.}
Let $R_r$ be a frozen GT leaf region, and $D_r$ the diagonal of the
region's exact triangle-fragment bounding box.
The GT samples retain their stored region IDs.
Each prediction sample is assigned to the region containing its
closest GT surface point. Sharp-edge prediction samples are routed
through the same nearest-GT-point rule.
Distances still query the complete opposite surface or edge set.
Regional surface and sharp-edge F1 use
\begin{equation}
 \tau_r=\min(0.01D_r,0.01D).
 \label{supp:eq:localtau}
\end{equation}
Curvature radii remain fractions of the global $D$.
Regional curvature restricts the anchor pool and GT denominator to
anchors belonging to $R_r$, and averages over scales applicable within
that region.

\paragraph{Local-F1 ($\uparrow$).}
Let $F_r$ be the region's surface F1 using Eq.~\ref{supp:eq:localtau} and
$\alpha_r=\sum_{i\in R_r}w_i/\sum_iw_i$ its sampled GT area fraction.
We report
\begin{equation}
 \operatorname{Local\mbox{-}F1}=100\sum_r\alpha_r F_r.
\end{equation}

\paragraph{Local-Detail ($\uparrow$).}
For GT and prediction edge-sample index sets $\mathcal{I}_r^G$
and $\mathcal{I}_r^P$ assigned to region $r$, compute
\begin{align}
 P_{E,r}&=\frac{\sum_{j\in\mathcal{I}_r^P}
 \ind[d(\boldsymbol{e}_j^P,\mathcal{L}_G)\leq\tau_r]}
 {|\mathcal{I}_r^P|},\\
 R_{E,r}&=\frac{\sum_{i\in\mathcal{I}_r^G}
 \ind[d(\boldsymbol{e}_i^G,\mathcal{L}_P)\leq\tau_r]}
 {|\mathcal{I}_r^G|},\\
 E_r&=\frac{2P_{E,r}R_{E,r}}{P_{E,r}+R_{E,r}}.
\end{align}
For regional eligible anchors $\mathcal{E}_{r,h}$ and paired-valid
anchors $\mathcal{V}_{r,h}$, curvature agreement is
\begin{align}
 C_{r,h}&=\frac{\sum_{i\in\mathcal{V}_{r,h}}w_iq_{i,h}}
 {\sum_{i\in\mathcal{E}_{r,h}}w_i},\\
 C_r&=\frac{1}{|\mathcal{H}_{r,\mathrm{app}}|}
 \sum_{h\in\mathcal{H}_{r,\mathrm{app}}}C_{r,h},
\end{align}
where $q_{i,h}$ uses the same curvature errors as the global score,
and $\mathcal{H}_{r,\mathrm{app}}$ contains GT-applicable regional scales.
The regional combined score is $D_r^{\mathrm{detail}}=\sqrt{E_rC_r}$
when both components apply.
With $\mathcal{R}_{\mathrm{valid}}$ the finite applicable regions,
\begin{equation}
 \operatorname{Local\mbox{-}Detail}=100\,
 \frac{\sum_{r\in\mathcal{R}_{\mathrm{valid}}}
 \alpha_r D_r^{\mathrm{detail}}}
 {\sum_{r\in\mathcal{R}_{\mathrm{valid}}}\alpha_r}.
 \label{supp:eq:localdetail}
\end{equation}
We accompany the mean with valid-region area coverage, which is
distinct from curvature-correspondence coverage.
An asset without any valid regional detail is $\NA$.

\paragraph{Spatial feedback representation.}
The evaluator returns the six global metrics, Local-F1 as an auxiliary
scalar, and per-region records with normalized spatial coordinates.
Local-Detail is derived for reporting from these regional records.
Each record contains a stable task-local region ID, GT area fraction,
surface F1, edge F1, curvature score, combined detail, and a normalized
regional bounding box.
For global GT box center $\boldsymbol{c}_G$, the location fields are
\begin{equation}
 \widetilde{\boldsymbol{b}}_{r,\min/\max}=
 \frac{\boldsymbol{b}_{r,\min/\max}-\boldsymbol{c}_G}{D}.
\end{equation}
The normalized center is the midpoint of these bounds in Blender XYZ.
This identifies where an error occurs without returning mesh vertices,
absolute reference dimensions, or semantic part labels.
The Evolver can combine these records with reference/prediction preview
sheets, execution logs, and source code to target revisions.
The Actor receives the selected skills, previous code when applicable,
and the Evolver's revision directive.

\begin{figure*}[tp]
\centering
\includegraphics[width=\textwidth,height=0.72\textheight,keepaspectratio]{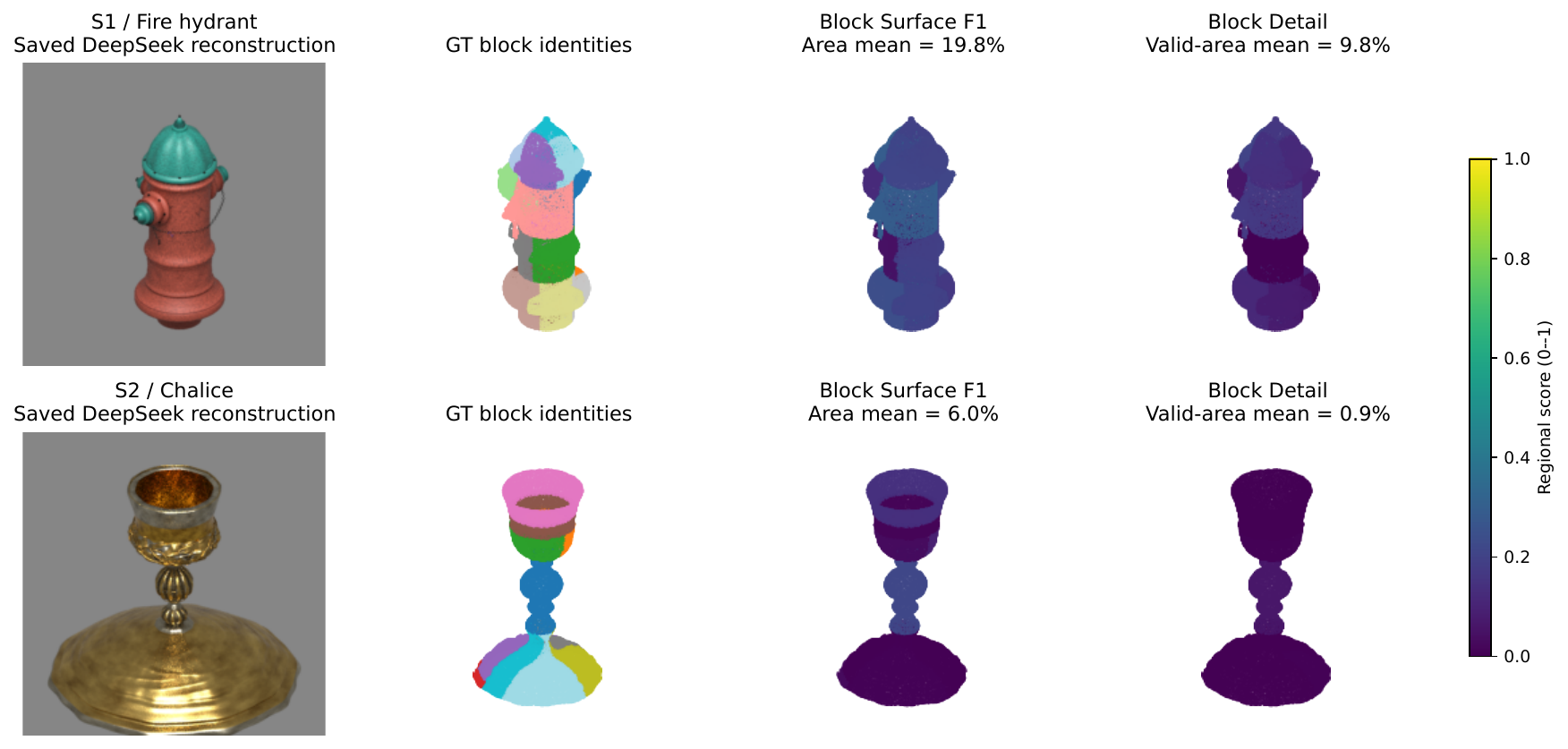}
\caption{DeepSeek reconstructions and spatially localized surface F1 and Detail feedback.}
\label{supp:fig:feedback}
\end{figure*}

\subsection{Evaluator design and execution budgets}

The evaluator separates untrusted reconstruction execution, artifact
validation, geometry extraction, reference rendering, and numerical
scoring. The generated program builds one standalone Blender scene
at frame 1, in meters, using built-in or procedural materials.
The runner saves \texttt{prediction.blend}; submitted scripts do not
render, set evaluation cameras, access private references, or invoke
the evaluator.

Execution uses headless Blender/bpy 5.0.1, CPython 3.11.5, and NumPy
1.26.4. The isolated worker has no network, credentials, or GT mount.
Separate Linux user, mount, PID, network, IPC, and UTS namespaces
restrict each job. Blender's automatic script execution is disabled
when loading the saved artifact.
The trusted extraction worker evaluates visible mesh-like objects,
instances, modifiers, and their world transforms, and exports the
finite triangle geometry in meters.
Reflected transforms have their triangle winding corrected.

\begin{table}[t]
\centering
\small
\begingroup
\setlength{\tabcolsep}{4.2pt}
\renewcommand{\arraystretch}{1.12}
\caption{Execution and evaluation resource limits.}
\label{supp:tab:runtime}
\begin{tabularx}{\columnwidth}{Xr}
\toprule
\textbf{Stage / constraint} & \textbf{Limit} \\
\midrule
Program execution, CPU and wall time & 1,800 s each \\
Program execution memory & 32 GiB \\
Standalone artifact validation & 120 s \\
Geometry extraction / scoring stage & 1,800 s \\
Fixed-camera rendering & 14,400 s \\
Overall evaluation & 21,600 s \\
Evaluated triangle count & 20 million \\
Single execution output file & 128 MiB \\
Execution workspace / temporary storage & 256 / 512 MiB \\
Workspace / temporary inodes & 1,024 / 4,096 \\
Open files / processes / threads & 256 / 64 / 512 \\
\bottomrule
\end{tabularx}
\endgroup
\end{table}

The renderer replaces submitted cameras and lights with the recorded
reference settings. It reconstructs orthographic or perspective
cameras, checks their resulting pixel projection against the recorded
mapping, and preserves the prediction's material nodes.
Reference profiles fix Cycles samples, lighting, background, view
transform, and exposure. Feature comparisons require every expected
view to be present in its original order.
Frozen Uni3D and DINOv2 weights are loaded from the local cache, using
PyTorch 2.7.1 and torchvision 0.22.1.

The public scoring status is \texttt{computed} or \texttt{fail}.
Low surface agreement or incomplete curvature correspondence is a
measured quality outcome. Failure of a requested encoder is an
evaluation failure. Inapplicable values remain null.
An internal diagnostics file retains component scores, fit validity,
coverage, thresholds, execution timing, and errors.
Summary means use finite applicable values together with valid counts,
attempted counts, and failure rates.

\subsection{Human Evaluation Protocol}
\label{supp:human}

We conduct a reference-based ranking study on 50 tasks, each with five
candidate reconstructions. Six reviewers are each assigned all 50 tasks,
giving 300 review assignments.

The web interface displays the GT and five anonymous candidates labeled
A--E. Candidate identities, model names, automatic scores, and other
reviewers' responses are hidden.
All models use a common scale and neutral materials.
Reviewers can rotate, zoom, pan, synchronize views, and inspect
wireframes in the interactive 3D viewer.
Reviewers rank the five candidates from 1 (closest to GT) to 5
(least similar), considering overall similarity, proportions,
structural completeness, and local detail.
Each reviewer has one current ranking per task.

\paragraph{Inter-reviewer agreement.}
Let $m_t$ be the number of available reviewers for task $t$, $n=5$
the number of candidates, and $r_{tjk}$ the rank assigned by reviewer
$j$ to candidate $k$. Define $R_{tk}=\sum_jr_{tjk}$ and
$\overline{R}_t=m_t(n+1)/2$. With strict rankings, Kendall's concordance is
\begin{equation}
 W_t=\frac{12\sum_{k=1}^{n}(R_{tk}-\overline{R}_t)^2}
 {m_t^2(n^3-n)}.
\end{equation}
For reviewers $a,b$, Spearman correlation is
\begin{equation}
 \rho_{t,ab}=1-
 \frac{6\sum_{k=1}^{n}(r_{tak}-r_{tbk})^2}{n(n^2-1)}.
\end{equation}
We average reviewer-pair correlations within each task and then
macro-average across tasks; $W_t$ is likewise macro-averaged.
The mean Kendall's $W$ is 0.560 (95\% CI: 0.515--0.604), and the
mean pairwise Spearman correlation is 0.472 (95\% CI: 0.418--0.524).

\paragraph{Comparison with automatic metrics.}
The mean submitted rank of each candidate defines the human consensus
within its task. We compare this consensus with each metric's ranking
of the same five candidates, using ascending nCD and scale error and
descending similarity scores; metric ties receive average ranks.
Spearman correlation is calculated per task and macro-averaged.
Surface F1, Detail, nCD, and DINOv2 achieve correlations of 0.651,
0.594, 0.683, and 0.706.

\section{Model Integration}
\label{supp:integration}

\paragraph{Shared task and submission interface.}
All evaluation modes use the task instruction, camera calibration,
and ordered RGB images. The controller removes evaluator-only metadata
before constructing model-visible inputs. Images are delivered through
the provider's native multimodal interface or the official CLI's image
tools. Each model ultimately submits a complete Blender Python program,
which is executed and scored independently under the resource limits
in Table~\ref{supp:tab:runtime}.

\paragraph{Direct API integration.}
Each task starts an independent conversation and requests one complete
candidate. Direct API runs use no agent CLI, external tools, skill
memory, Evolver, or execution-feedback loop. Once a valid response is
saved, only the submitted program is executed; code errors and low
scores do not trigger regeneration.

We use the native multimodal APIs for each provider: OpenAI Responses
with \texttt{instructions} and multimodal \texttt{input};
Anthropic Messages with \texttt{system} and multimodal
\texttt{messages}; and Gemini \texttt{generateContent} with
\texttt{systemInstruction}, \texttt{contents}, and inline image data.
Other models use verified multimodal Chat Completions-compatible
adapters. The adapters preserve image order, full camera calibration,
model identity, termination reason, and raw usage. Output limits are
set through the corresponding native fields:
\texttt{max\_output\_tokens}, \texttt{max\_tokens}, or
\texttt{maxOutputTokens}. Temperature, top-$p$, and sampling seeds
remain at provider defaults, with no additional overrides or
multiple-candidate selection. The single-shot adapters use streaming
responses, a 3,600 s outer generation limit, and a 900 s read timeout.

\paragraph{Official harness integration.}
Harness baselines use official coding agents with their native tool
loops. Each task receives an independent workspace containing public
inputs, a read-only Blender 5.0 API manual, and installed
image/geometry Python tools. The workspace excludes the Blender
runtime, GT assets, evaluator, cross-task memory, and external
Evolver. Agents may inspect images and documentation, perform
numerical calculations, and check Python syntax before submitting
their final programs.

The integrations use Codex 0.164.0, Claude Code 2.1.197,
DeepSeek dsh 0.2.2-rc.1, and Kimi Code 2.6.1, with upstream
interfaces based on Responses, Messages, a Messages-compatible relay,
and a Chat Completions relay, respectively. Each task permits at most
five normal model requests, shared across native tool interactions
and final-code generation, including across sessions and restarts.
Errors have a separate allowance of three retries, corresponding to
at most four failed attempts including the initial failure. The final
normal request disables tools and requires complete source code;
the controller rejects any sixth normal request. The native session
wall-time limit is 1,800 s, the upstream timeout is 900 s, and
evaluator infrastructure recovery allows at most three persistent
attempts using the original program.

A host-side relay enforces model identity and request-budget
accounting. Agents operate within a namespace using a placeholder
credential and loopback relay, while real credentials remain outside
the agent and evaluation workers. Saved events include requests,
stream completion, final source code, normal and error counters,
execution artifacts, and score receipts.

\paragraph{RSI role-specific request budgets.}
RSI separates Actor, retrieval, and review roles, distinguishing
a reconstruction attempt from a model-tool interaction. The training
configuration uses a code-only Actor with a 128,000-token output cap
and a 3,600 s session limit. Training retrieval is an Evolver call
that selects existing nodes, whereas training review is an Evolver
call that may update memory. Each RSI role permits at most five
normal model requests per task or role invocation, including tool
interactions and the final decision.

\section{Recursive Skill Improvement}
\label{supp:rsi}

\subsection{Actor--Evaluator--Evolver loop}

Let $M_t$ denote the current memory bundle and $o_i$ the public
observations of training task $i$.
The training Evolver first selects skill IDs from $M_t$ and provides
task-specific guidance. The Actor receives the resulting complete
skill context, the images and calibrated cameras, and any previous
program and revision directive.
It generates program $p_i^{(k)}$ in one response.
The independent Evaluator executes the program and computes global
and regional feedback $f_i^{(k)}$.
The controller stores the submitted code, execution outcome, selected
skill IDs, memory version, scores, and evidence reference.
The Evaluator is the deterministic Blender/numerical pipeline.

Training proceeds in 20 batches of five tasks.
Within a batch, all tasks in one wave use the same frozen memory
version. The controller waits for the wave to finish before review
and memory publication.
The Evolver then examines the batch's actual attempt histories,
six-metric scores, regional feedback, reference/prediction preview
sheets, and execution logs.
It may select at most two eligible tasks for revision in each round, accompanied
by concrete feedback, and may update memory.
There are at most three waves, so every task receives at most three
scored reconstruction attempts including its initial attempt.
The terminal review cannot request a further retry.

This produces two coupled recursions:
\begin{align}
 p_i^{(k+1)}&=\operatorname{Actor}
 \bigl(o_i,M_{t+1},p_i^{(k)},g_i^{(k)}\bigr),\\
 M_{t+1}&=\operatorname{Publish}
 \bigl(M_t,\operatorname{Evolver}(\mathcal{E}_t)\bigr),
\end{align}
where $g_i^{(k)}$ is the Evolver's revision directive and
$\mathcal{E}_t$ contains admissible training evidence.
The first recursion revises a reconstruction; the second retains
procedures that may help later tasks.
Model parameters remain fixed.

Low scores do not automatically reject a skill update.
Publication checks establish structural validity and evidence
provenance.
All attempts are preserved, and training reporting uses the last
attempt.
An execution failure can motivate an evidenced robustness lesson.

\subsection{Library, Tree, Graph, and code dependencies}

\paragraph{Skill Library.}
Persistent storage contains reusable skill nodes, shared executable
tools, and immutable experiences/trajectories.
A skill record contains its ID, home category, parent ID, title,
method, applicability conditions, counterexamples, executable code,
explicit dependencies, and training evidence references.
Experiences retain code and measured outcomes separately from the
skill's generalized description.
Shared tools initially provide scene initialization, primitive
creation, and basic materials.
The first training pass starts from these tools and five category
roots, without learned task-specific skills.

\paragraph{Five Skill Trees.}
Each learned skill belongs to one home category and has one parent
within that category. Parent pointers must terminate at the fixed
S1--S5 root without a cycle.
The Tree expresses organization and specialization.
Its home category is not a restriction on learning: evidence from any
training task may support a skill in any Tree or a shared tool.

\paragraph{Skill Graph.}
A Graph edge is an undirected association between two skill/tool
nodes. It contains endpoints, a weight in $[0,1]$, provenance
references, and hashes of the endpoint node versions.
Associations may be supported by training experiences or explicitly
marked as \texttt{content\_prior}.
The latter denotes a content-based relation, not a measured
co-utilization gain.
The Graph can connect related procedures within or across Trees.

\paragraph{Executable dependencies.}
Code dependencies are a separate directed acyclic graph over skill
and tool modules. They specify which modules must be present before
execution. The controller checks that dependency IDs exist and that
the directed graph is acyclic.
An association edge does not create an executable prerequisite.
Likewise, a Tree parent does not imply that the child's code calls
the parent's code.

\subsection{Evidence-backed updates}

At review, the Evolver returns either \texttt{wait} or an update
proposal, a reason, a compressed batch summary, and a retry list.
A proposal can upsert/remove skills, update/remove shared tools, and
replace the association edge list.

The controller applies the proposal to a copy of its declared parent
bundle and validates the candidate before publication:
(1) exact schema and unique IDs;
(2) fixed category roots and within-category acyclic parent pointers;
(3) parseable Python and complete, acyclic code dependencies;
(4) nonempty admissible training evidence for upserted skills;
(5) valid Graph endpoints, finite weights, and provenance;
(6) Graph endpoint hashes matching the candidate nodes; and
(7) the active memory-size budget.
If a node changes, associations referring to its previous version
must be replaced and rebound explicitly.

An accepted candidate is stored as an immutable content-addressed
bundle. Under a publication lock, the controller checks that its
expected parent is still current, writes the publication evidence and
acceptance decision, and atomically advances the current pointer.
This rejects stale concurrent updates and preserves complete lineage.
A valid \texttt{wait} leaves the memory unchanged.

\begin{figure*}[tp]
\centering
\includegraphics[width=\textwidth,height=0.72\textheight,keepaspectratio]{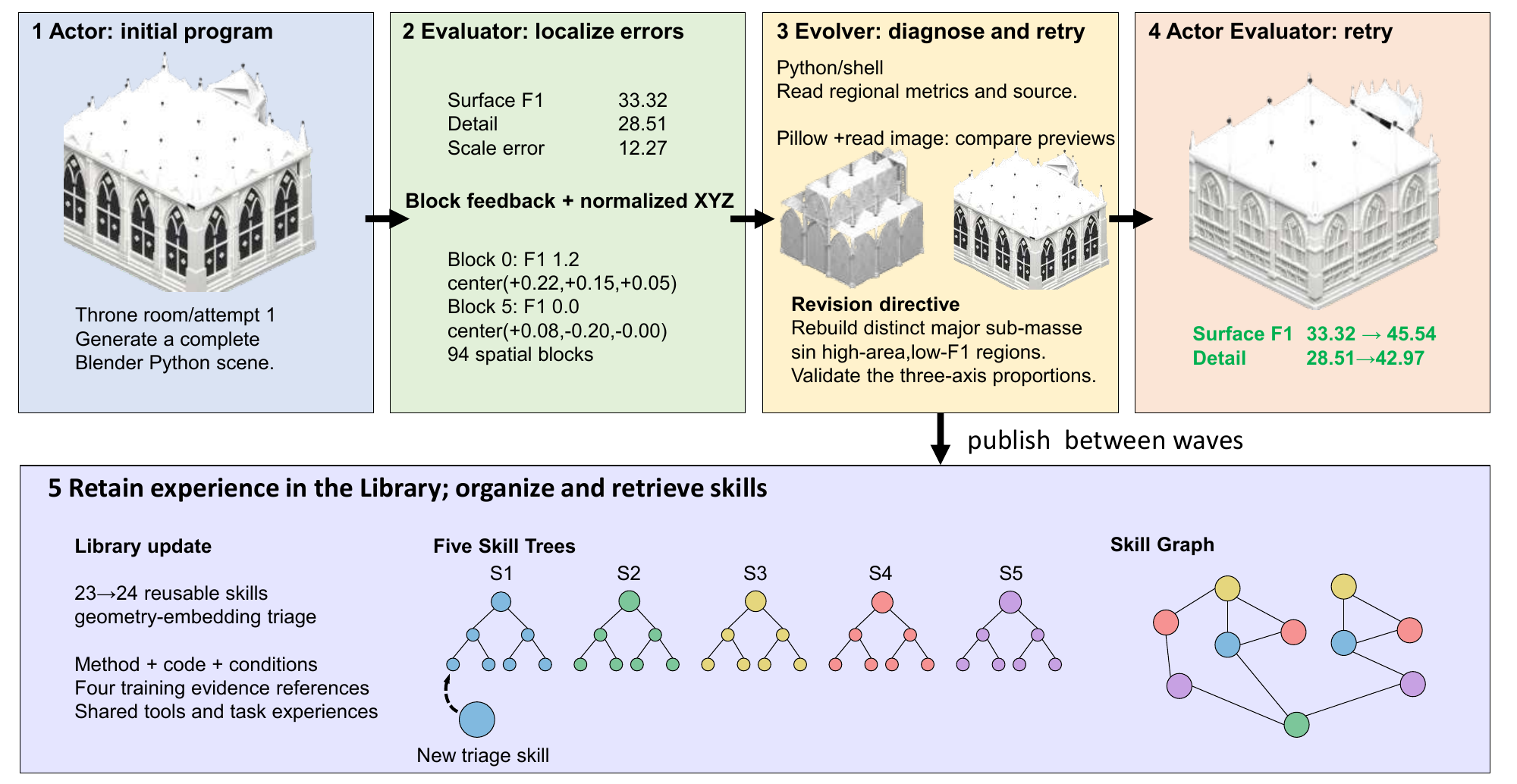}
\caption{A complete RSI learning cycle: modeling, localized evaluation,
tool-assisted review, targeted revision, and persistent skill organization.}
\label{supp:fig:memory}
\end{figure*}

Figure~\ref{supp:fig:memory} follows a real DeepSeek training batch.
The initial throne-room reconstruction receives surface F1 33.32
and Detail 28.51. The Evolver reads regional scores and coordinates,
examines source code, and uses Python and image tools to compare
reference and prediction previews.
Its directive targets distinct primary masses and three-axis
proportions; the revised attempt reaches F1 45.54 and Detail 42.97.
At the same batch review, the Evolver publishes a geometry--embedding
triage skill under S1, increasing the Library from 23 to 24 skills.
The five Trees organize the procedures, and the Graph retains six
associations for subsequent retrieval.

\subsection{Retrieval and skill use}

\paragraph{Training retrieval.}
The retrieval-role Evolver receives public observations, a compact
catalog, and read-only node/Graph files.
It selects exact existing IDs and returns task guidance.
The controller includes selected nodes, their Tree ancestors, the
foundation tools, and the complete transitive code-dependency closure.
It orders executable modules topologically and includes associations
whose endpoints are both present.
The Actor receives this bounded complete context and can adapt or
import its procedures into the reconstruction.
Recorded selected IDs establish retrieval, while actual use must be
assessed from generated code or tool traces.

\paragraph{Frozen test retrieval.}
After training, the enabled memory is frozen and exposed through a
catalog, Tree parent pointers, \texttt{graph.json}, complete node
files, and executable helper modules.
The test Actor chooses what to inspect and use inside its own native
tool loop. Retrieval and final code share the same limit of five
normal model requests per task.
There is no separate test Evolver or automatic Graph-neighbor
expansion step in this inference path.

\paragraph{Skill-access evaluation.}
For Acquisition on category $S_i$, only Tree $T_i$ is enabled.
For Transfer on the same tasks, $T_i$ is disabled and the other four
Trees are enabled.
For Composition on Hard tasks, all five Trees are enabled.
Shared tools remain available subject to dependency validity.
Masking removes disabled Tree nodes, descendants with blocked
parents, nodes with blocked dependencies, and incident associations
until the visible set is stable.
The underlying stored bundle remains unchanged.

\section{Real Cases and Visualizations}
\label{supp:cases}

\subsection{Material-preserving model comparisons}

\begin{figure*}[p]
\centering
\includegraphics[width=\textwidth,height=0.86\textheight,keepaspectratio]{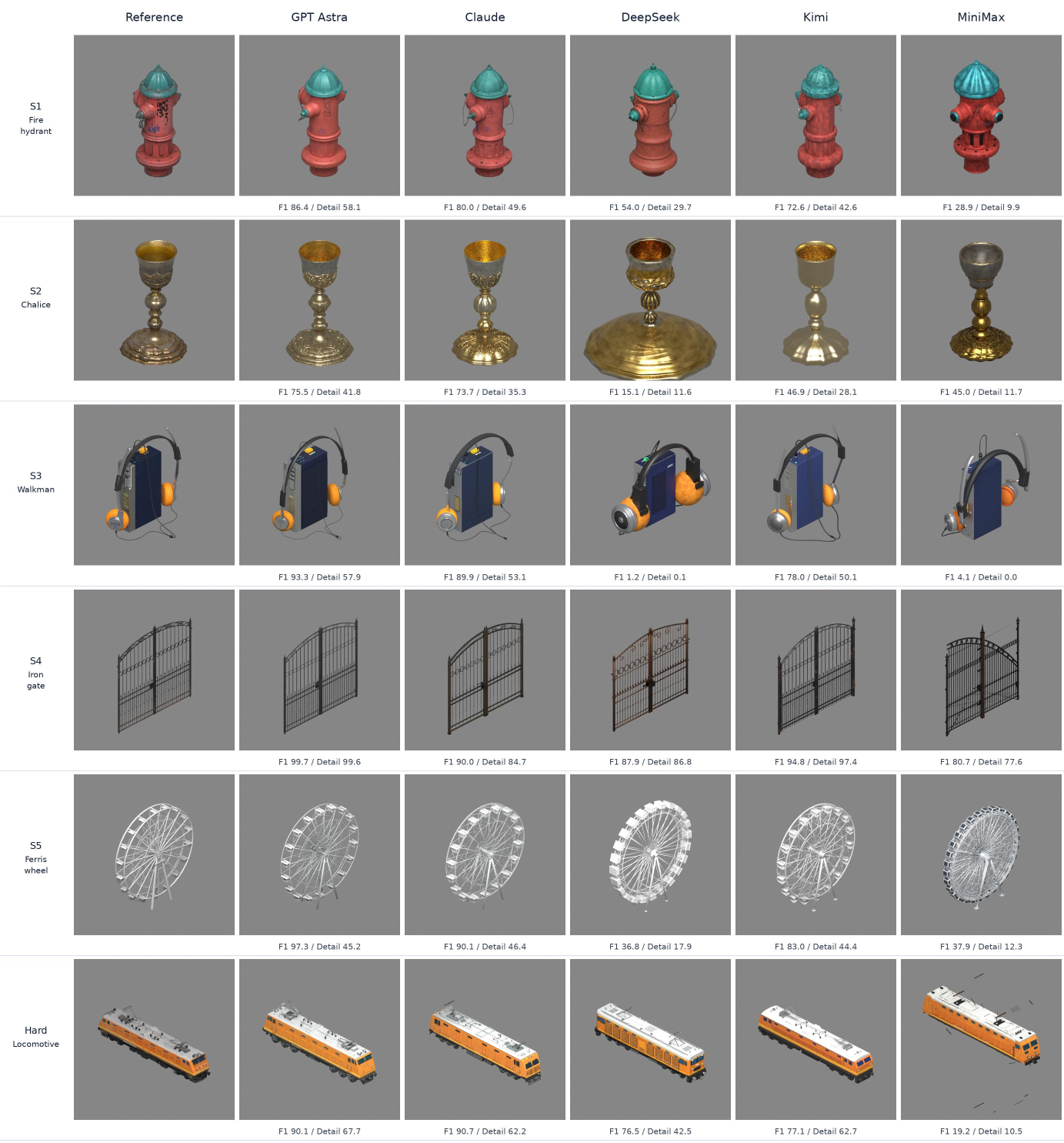}
\caption{Material-preserving model comparisons across S1--S5 and HardSet,
with surface F1 and Detail scores.}
\label{supp:fig:materials}
\end{figure*}

Figure~\ref{supp:fig:materials} compares six real assets, one from each
S1--S5 category and Hard.
The examples include dimension control, ornamental local geometry,
component assembly, repetition, spatial arrangement, and a complex
locomotive.

Every candidate uses the first recorded reference camera at the
original 512$\times$512 resolution, with the recorded Cycles profile,
lighting, background, and color transform.
We remove only the global bounding-box-center translation, retaining
metric scale and orientation.
Reference panels use the corresponding original RGB image.

All 30 candidate renders use the original saved scenes.
Figure annotations report each model's saved surface F1 and Detail.

\subsection{Interpreting block-based evaluator feedback}
Figures~\ref{supp:fig:partition} and \ref{supp:fig:feedback} connect
stored GT subdivision to actual feedback on the fire-hydrant and
chalice tasks.
For each asset, every method uses the same GT blocks, and a low score
can be localized through the normalized regional bounds.

These cases also show why the global and local measurements should
remain separate. Regional tolerances shrink with block span, while
curvature similarity includes correspondence coverage.
A low Local-Detail value can reflect missing or displaced detail,
sharp-edge mismatch, or failure of the curvature gates.
Conditional paired-anchor similarity and valid-region area coverage
help interpret those outcomes.

\subsection{Observed RSI reconstruction trajectories}

\begin{figure*}[p]
\centering
\includegraphics[width=\textwidth,height=0.86\textheight,keepaspectratio]{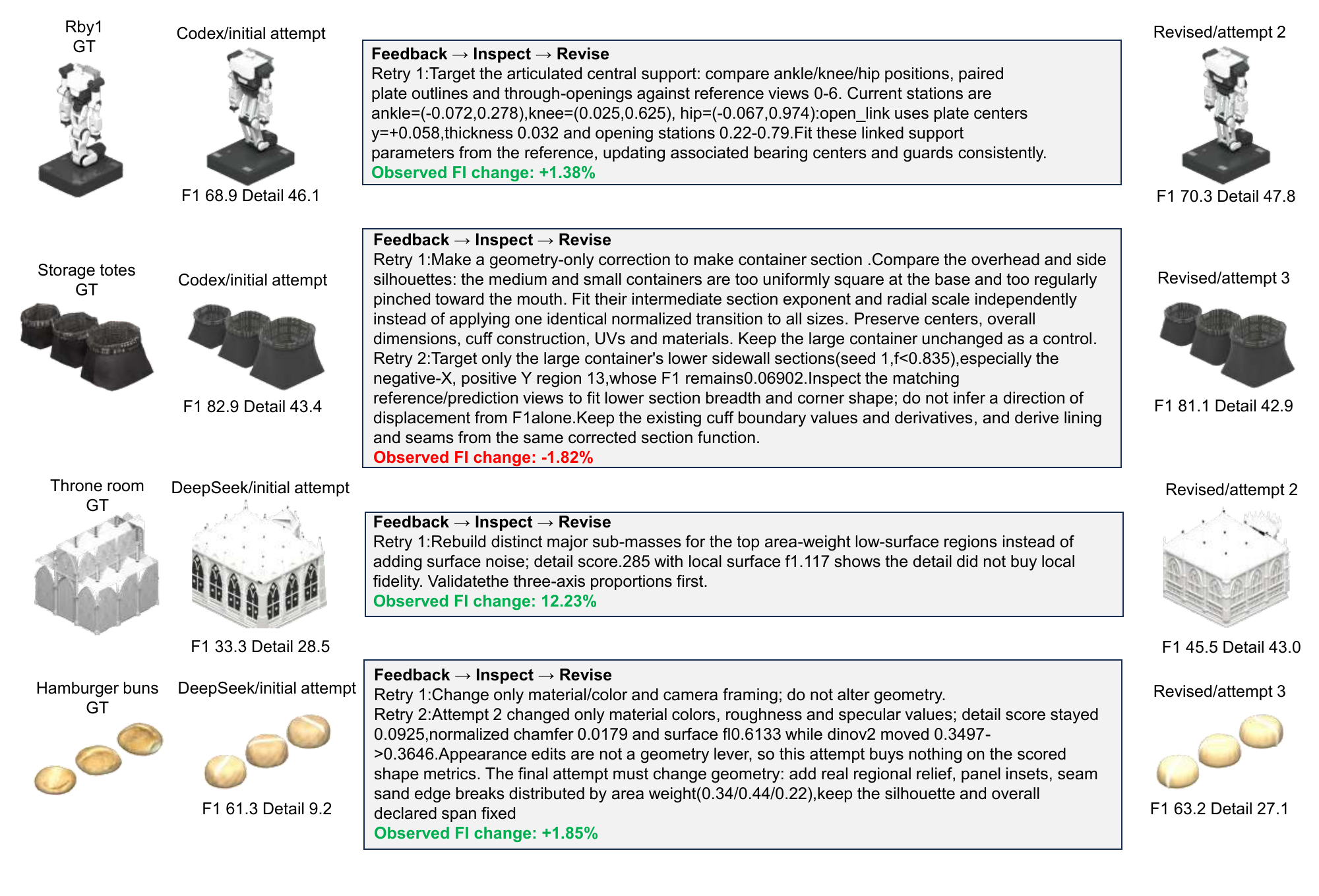}
\caption{RSI revision trajectories with Evolver directives and measured outcomes.}
\label{supp:fig:rsi-cases}
\end{figure*}

Figure~\ref{supp:fig:rsi-cases} shows four saved trajectories from continued training.
They connect the Evolver's revision directives with the first and
latest attempts, using their original evaluator renders and materials.
The examples include both improvements and a regression.

For Codex on Rby1, surface F1 changes from 68.94\% to 70.33\%, with
Detail increasing from 46.14\% to 47.80\%.
On storage totes, the intermediate surface F1 is 83.02\%, but the
last attempt reaches 81.12\%; the lower terminal value is retained.
For DeepSeek on the throne room, surface F1 changes from 33.32\% to
45.54\%, and Detail from 28.51\% to 42.97\%.
On hamburger buns, the first two attempts have identical saved
scores, followed by a final attempt with F1 63.19\% and Detail
27.13\%, compared with 61.33\% and 9.25\% initially.
The directives isolate distinct edits: linked support geometry,
container cross-sections, primary building masses, and regional relief.

\section{Skill-wise Results across Reasoning Efforts}
\label{supp:subset-results}

We evaluate 26 configurations, including the 16 main-table methods
and ten additional Medium or Low settings, all on the same
1,000-task benchmark.

Tables~\ref{supp:tab:results-s1}--\ref{supp:tab:results-hardset}
report results across S1--S5 and HardSet.
Execution success is computed over all tasks in each subset,
whereas quality metrics are averaged over valid applicable
reconstructions. To recover an overall mean, we weight each
subset mean by its valid sample count for the corresponding metric:
\begin{equation}
\bar{x}_m =
\frac{\sum_c n_{c,m}\bar{x}_{c,m}}
{\sum_c n_{c,m}},
\end{equation}
where $n_{c,m}$ denotes the number of valid values for metric $m$
in subset $c$.

\paragraph{Skill-dependent performance.}
Reconstruction difficulty varies across skills. For GPT-6.1 Sol
(High), S4 achieves 78.9 F1 and 51.0 Detail, compared with
61.2/32.3 for S2 and 57.8/28.0 for HardSet. Their corresponding
Local-Detail scores are 25.4, 14.0, and 9.3.
Codex with GPT-6 Astra shows a similar pattern, with F1 scores
of 79.0, 63.2, and 58.6, respectively.
The substantial gaps between global and local metrics further
highlight the difficulty of accurately reconstructing local details.

\begin{table*}[t]
\centering
\scriptsize
\begingroup
\setlength{\tabcolsep}{1.0pt}
\renewcommand{\arraystretch}{1.12}
\caption{Results on S1 by model and reasoning effort.}
\label{supp:tab:results-s1}
\begin{tabularx}{\textwidth}{@{}>{\raggedright\arraybackslash}p{5.4cm} c c c c c c c c c >{\centering\arraybackslash}X@{}}
\toprule
\multirow{2}{*}{\textbf{Model}} & \multirow{2}{*}{\textbf{Effort}} & \multirow{2}{*}{\textbf{Exec.} $\uparrow$} & \multicolumn{6}{c}{\textbf{Geometry}} & \multicolumn{2}{c}{\textbf{Semantic}} \\
\cmidrule(lr){4-9} \cmidrule(lr){10-11}
& & & \textbf{F1} $\uparrow$ & \textbf{Local-F1} $\uparrow$ & \textbf{Detail} $\uparrow$ & \textbf{Local-Detail} $\uparrow$ & \textbf{nCD} $\downarrow$ & \textbf{Scale(\%)} $\downarrow$ & \textbf{Uni3D} $\uparrow$ & \textbf{DINOv2} $\uparrow$ \\
\midrule
\mbox{\modellogo{openai}~GPT-6.1 Sol} & High & 97.6 & 73.1 & 46.3 & 45.1 & 23.8 & 1.9 & 4.0 & 81.9 & 84.2 \\
\mbox{\modellogo{openai}~GPT-6 Astra} & High & 96.4 & 74.8 & 47.0 & 46.2 & 24.2 & 1.8 & 3.1 & 82.1 & 84.2 \\
\mbox{\modellogo{codex}~Codex + GPT-6 Astra} & High & 90.5 & 75.7 & 48.7 & 47.4 & 25.3 & 1.8 & 2.8 & 82.3 & 85.2 \\
\mbox{\modellogo{claudecode}~Claude Code + Claude Opus 5.5} & High & 90.0 & 76.1 & 42.5 & 49.1 & 22.7 & 0.9 & 3.3 & 82.9 & 81.7 \\
\mbox{\modellogo{claude}~Claude Opus 5.5} & High & 91.7 & 67.0 & 37.7 & 41.9 & 18.6 & 2.3 & 3.9 & 78.5 & 75.9 \\
\mbox{\modellogo{kimi}~Kimi Code + Kimi K3.1 Pro} & High & 93.2 & 72.7 & 43.7 & 44.1 & 20.8 & 1.7 & 2.8 & 77.1 & 81.2 \\
\mbox{\modellogo{kimi}~Kimi K3.1 Pro} & High & 80.5 & 69.4 & 42.0 & 39.9 & 19.4 & 2.2 & 3.8 & 77.6 & 80.0 \\
\mbox{\modellogo{claude}~Claude Fable 5.1} & High & 94.7 & 63.3 & 32.3 & 36.6 & 14.1 & 2.7 & 7.3 & 76.8 & 74.7 \\
\mbox{\modellogo{claude}~Claude Sonnet 5.5} & High & 91.1 & 64.2 & 34.6 & 38.9 & 17.1 & 2.6 & 4.8 & 76.7 & 73.2 \\
\mbox{\modellogo{gemini}~Gemini-4 Argon} & Medium & 77.5 & 59.6 & 32.8 & 33.7 & 14.6 & 2.8 & 7.4 & 82.0 & 81.8 \\
\mbox{\modellogo{hunyuan}~Hy4} & High & 84.0 & 66.8 & 37.8 & 35.7 & 15.7 & 2.4 & 4.2 & 77.2 & 78.0 \\
\mbox{\modellogo{qwen}~Qwen3.8-Max} & High & 81.1 & 58.7 & 29.9 & 29.6 & 11.4 & 3.4 & 8.3 & 75.0 & 71.7 \\
\mbox{\modellogo{DSH}~DeepSeek Harness + DeepSeek V4.1 Flash} & High & 92.0 & 50.5 & 27.1 & 24.9 & 9.8 & 4.8 & 14.8 & 72.8 & 73.4 \\
\mbox{\modellogo{deepseek}~DeepSeek V4.1 Flash} & High & 85.1 & 49.7 & 25.1 & 21.9 & 8.0 & 4.2 & 10.8 & 71.3 & 72.3 \\
\mbox{\modellogo{zai}~GLM-5.3 Flash} & High & 65.7 & 44.9 & 22.2 & 19.2 & 6.7 & 4.4 & 13.0 & 69.5 & 68.4 \\
\mbox{\modellogo{minimax}~MiniMax-M3.1-Flash} & High & 68.6 & 45.1 & 23.2 & 19.8 & 6.7 & 243.4 & 864.6 & 68.5 & 71.1 \\
\mbox{\modellogo{openai}~GPT-6.1 Sol} & Medium & 95.9 & 67.8 & 40.7 & 39.0 & 18.5 & 2.3 & 4.9 & 80.4 & 81.9 \\
\mbox{\modellogo{openai}~GPT-6.1 Sol} & Low & 99.4 & 62.0 & 36.2 & 35.0 & 17.0 & 2.8 & 7.9 & 78.9 & 79.0 \\
\mbox{\modellogo{openai}~GPT-6 Astra} & Medium & 98.2 & 68.3 & 40.9 & 40.5 & 19.8 & 2.2 & 4.4 & 80.5 & 80.9 \\
\mbox{\modellogo{openai}~GPT-6 Astra} & Low & 98.8 & 67.5 & 39.3 & 39.0 & 18.8 & 2.3 & 4.9 & 80.7 & 80.2 \\
\mbox{\modellogo{claude}~Claude Fable 5.1} & Medium & 97.0 & 58.4 & 28.8 & 32.3 & 11.6 & 2.8 & 5.9 & 75.2 & 71.4 \\
\mbox{\modellogo{claude}~Claude Fable 5.1} & Low & 96.4 & 54.2 & 26.8 & 27.9 & 10.2 & 3.3 & 8.9 & 73.4 & 68.9 \\
\mbox{\modellogo{deepseek}~DeepSeek V4.1 Flash} & Medium & 89.3 & 47.5 & 23.5 & 20.6 & 7.1 & 3.8 & 9.1 & 71.2 & 72.7 \\
\mbox{\modellogo{deepseek}~DeepSeek V4.1 Flash} & Low & 91.7 & 47.1 & 21.7 & 21.3 & 7.9 & 4.1 & 11.2 & 70.9 & 71.8 \\
\mbox{\modellogo{kimi}~Kimi K3.1 Pro} & Medium & 82.8 & 71.7 & 45.3 & 42.5 & 21.8 & 2.0 & 3.0 & 78.3 & 82.4 \\
\mbox{\modellogo{kimi}~Kimi K3.1 Pro} & Low & 85.2 & 57.6 & 31.5 & 31.4 & 14.0 & 3.2 & 7.7 & 75.9 & 78.0 \\
\bottomrule
\end{tabularx}
\endgroup
\end{table*}
\begin{table*}[t]
\centering
\scriptsize
\begingroup
\setlength{\tabcolsep}{1.0pt}
\renewcommand{\arraystretch}{1.12}
\caption{Results on S2 by model and reasoning effort.}
\label{supp:tab:results-s2}
\begin{tabularx}{\textwidth}{@{}>{\raggedright\arraybackslash}p{5.4cm} c c c c c c c c c >{\centering\arraybackslash}X@{}}
\toprule
\multirow{2}{*}{\textbf{Model}} & \multirow{2}{*}{\textbf{Effort}} & \multirow{2}{*}{\textbf{Exec.} $\uparrow$} & \multicolumn{6}{c}{\textbf{Geometry}} & \multicolumn{2}{c}{\textbf{Semantic}} \\
\cmidrule(lr){4-9} \cmidrule(lr){10-11}
& & & \textbf{F1} $\uparrow$ & \textbf{Local-F1} $\uparrow$ & \textbf{Detail} $\uparrow$ & \textbf{Local-Detail} $\uparrow$ & \textbf{nCD} $\downarrow$ & \textbf{Scale(\%)} $\downarrow$ & \textbf{Uni3D} $\uparrow$ & \textbf{DINOv2} $\uparrow$ \\
\midrule
\mbox{\modellogo{openai}~GPT-6.1 Sol} & High & 93.8 & 61.2 & 31.7 & 32.3 & 14.0 & 2.5 & 4.5 & 78.3 & 78.0 \\
\mbox{\modellogo{openai}~GPT-6 Astra} & High & 97.7 & 60.2 & 31.0 & 31.1 & 14.1 & 2.7 & 4.9 & 77.7 & 77.4 \\
\mbox{\modellogo{codex}~Codex + GPT-6 Astra} & High & 93.0 & 63.2 & 33.8 & 34.1 & 15.5 & 2.7 & 5.2 & 78.6 & 78.9 \\
\mbox{\modellogo{claudecode}~Claude Code + Claude Opus 5.5} & High & 93.2 & 62.5 & 30.5 & 35.6 & 16.0 & 2.8 & 3.8 & 73.6 & 70.0 \\
\mbox{\modellogo{claude}~Claude Opus 5.5} & High & 94.9 & 56.2 & 26.3 & 30.1 & 11.7 & 3.0 & 4.5 & 72.5 & 65.6 \\
\mbox{\modellogo{kimi}~Kimi Code + Kimi K3.1 Pro} & High & 90.0 & 57.7 & 28.9 & 29.0 & 10.7 & 3.3 & 7.9 & 71.9 & 71.9 \\
\mbox{\modellogo{kimi}~Kimi K3.1 Pro} & High & 77.3 & 56.3 & 28.2 & 29.7 & 12.5 & 3.6 & 6.4 & 74.5 & 74.2 \\
\mbox{\modellogo{claude}~Claude Fable 5.1} & High & 96.1 & 53.7 & 24.1 & 27.0 & 9.8 & 3.2 & 5.2 & 73.0 & 66.7 \\
\mbox{\modellogo{claude}~Claude Sonnet 5.5} & High & 94.9 & 51.3 & 23.1 & 25.7 & 9.5 & 3.4 & 5.8 & 70.8 & 63.8 \\
\mbox{\modellogo{gemini}~Gemini-4 Argon} & Medium & 69.9 & 48.0 & 22.8 & 23.4 & 9.3 & 4.1 & 9.6 & 77.2 & 73.2 \\
\mbox{\modellogo{hunyuan}~Hy4} & High & 82.4 & 49.3 & 22.6 & 22.8 & 8.2 & 3.9 & 6.1 & 70.6 & 68.3 \\
\mbox{\modellogo{qwen}~Qwen3.8-Max} & High & 76.6 & 42.1 & 18.6 & 17.8 & 6.2 & 5.4 & 14.3 & 66.3 & 58.8 \\
\mbox{\modellogo{DSH}~DeepSeek Harness + DeepSeek V4.1 Flash} & High & 88.2 & 38.2 & 16.2 & 15.0 & 4.2 & 6.0 & 13.0 & 65.5 & 61.4 \\
\mbox{\modellogo{deepseek}~DeepSeek V4.1 Flash} & High & 84.6 & 38.0 & 15.1 & 14.1 & 3.7 & 5.3 & 13.2 & 66.5 & 64.1 \\
\mbox{\modellogo{zai}~GLM-5.3 Flash} & High & 68.0 & 33.4 & 12.5 & 12.1 & 3.0 & 6.0 & 17.3 & 63.9 & 59.3 \\
\mbox{\modellogo{minimax}~MiniMax-M3.1-Flash} & High & 66.4 & 34.7 & 14.2 & 14.2 & 4.5 & 26.8 & 375.4 & 63.7 & 60.8 \\
\mbox{\modellogo{openai}~GPT-6.1 Sol} & Medium & 98.8 & 53.6 & 26.0 & 26.7 & 11.3 & 3.1 & 6.5 & 76.0 & 74.7 \\
\mbox{\modellogo{openai}~GPT-6.1 Sol} & Low & 99.6 & 48.5 & 23.1 & 22.2 & 8.8 & 3.9 & 8.4 & 73.9 & 70.7 \\
\mbox{\modellogo{openai}~GPT-6 Astra} & Medium & 99.2 & 54.6 & 26.2 & 27.6 & 11.4 & 3.2 & 6.9 & 76.9 & 73.9 \\
\mbox{\modellogo{openai}~GPT-6 Astra} & Low & 99.2 & 53.9 & 25.8 & 26.1 & 10.7 & 3.2 & 6.7 & 75.8 & 73.1 \\
\mbox{\modellogo{claude}~Claude Fable 5.1} & Medium & 97.3 & 47.6 & 20.0 & 22.3 & 7.6 & 3.8 & 7.6 & 69.7 & 61.0 \\
\mbox{\modellogo{claude}~Claude Fable 5.1} & Low & 98.4 & 43.0 & 17.4 & 18.8 & 5.6 & 4.9 & 12.9 & 66.1 & 56.6 \\
\mbox{\modellogo{deepseek}~DeepSeek V4.1 Flash} & Medium & 93.0 & 35.8 & 14.3 & 13.0 & 3.6 & 5.1 & 13.6 & 66.1 & 63.4 \\
\mbox{\modellogo{deepseek}~DeepSeek V4.1 Flash} & Low & 94.5 & 35.6 & 14.8 & 12.8 & 3.6 & 5.4 & 14.1 & 65.4 & 62.2 \\
\mbox{\modellogo{kimi}~Kimi K3.1 Pro} & Medium & 78.5 & 57.5 & 28.3 & 29.8 & 12.4 & 3.0 & 4.7 & 75.3 & 75.1 \\
\mbox{\modellogo{kimi}~Kimi K3.1 Pro} & Low & 85.5 & 44.2 & 19.5 & 19.1 & 6.7 & 4.9 & 13.3 & 70.1 & 69.6 \\
\bottomrule
\end{tabularx}
\endgroup
\end{table*}
\begin{table*}[t]
\centering
\scriptsize
\begingroup
\setlength{\tabcolsep}{1.0pt}
\renewcommand{\arraystretch}{1.12}
\caption{Results on S3 by model and reasoning effort.}
\label{supp:tab:results-s3}
\begin{tabularx}{\textwidth}{@{}>{\raggedright\arraybackslash}p{5.4cm} c c c c c c c c c >{\centering\arraybackslash}X@{}}
\toprule
\multirow{2}{*}{\textbf{Model}} & \multirow{2}{*}{\textbf{Effort}} & \multirow{2}{*}{\textbf{Exec.} $\uparrow$} & \multicolumn{6}{c}{\textbf{Geometry}} & \multicolumn{2}{c}{\textbf{Semantic}} \\
\cmidrule(lr){4-9} \cmidrule(lr){10-11}
& & & \textbf{F1} $\uparrow$ & \textbf{Local-F1} $\uparrow$ & \textbf{Detail} $\uparrow$ & \textbf{Local-Detail} $\uparrow$ & \textbf{nCD} $\downarrow$ & \textbf{Scale(\%)} $\downarrow$ & \textbf{Uni3D} $\uparrow$ & \textbf{DINOv2} $\uparrow$ \\
\midrule
\mbox{\modellogo{openai}~GPT-6.1 Sol} & High & 95.1 & 72.5 & 36.2 & 42.2 & 18.2 & 2.0 & 4.8 & 84.0 & 84.8 \\
\mbox{\modellogo{openai}~GPT-6 Astra} & High & 100.0 & 71.9 & 35.5 & 41.5 & 17.2 & 2.2 & 4.9 & 84.6 & 85.2 \\
\mbox{\modellogo{codex}~Codex + GPT-6 Astra} & High & 95.7 & 73.1 & 36.5 & 42.1 & 17.5 & 2.0 & 4.7 & 84.1 & 84.7 \\
\mbox{\modellogo{claudecode}~Claude Code + Claude Opus 5.5} & High & 96.1 & 69.0 & 33.7 & 41.9 & 17.9 & 2.6 & 2.4 & 76.6 & 75.5 \\
\mbox{\modellogo{claude}~Claude Opus 5.5} & High & 96.3 & 65.7 & 30.3 & 37.5 & 13.3 & 2.8 & 6.2 & 80.2 & 76.3 \\
\mbox{\modellogo{kimi}~Kimi Code + Kimi K3.1 Pro} & High & 91.8 & 64.7 & 30.9 & 34.7 & 12.3 & 2.4 & 6.7 & 78.8 & 80.6 \\
\mbox{\modellogo{kimi}~Kimi K3.1 Pro} & High & 85.2 & 66.8 & 31.6 & 35.6 & 13.9 & 2.3 & 4.4 & 80.2 & 81.6 \\
\mbox{\modellogo{claude}~Claude Fable 5.1} & High & 97.5 & 64.8 & 28.9 & 35.5 & 12.4 & 2.4 & 4.4 & 80.5 & 76.5 \\
\mbox{\modellogo{claude}~Claude Sonnet 5.5} & High & 93.8 & 61.1 & 26.2 & 32.8 & 10.8 & 3.1 & 6.5 & 78.8 & 75.0 \\
\mbox{\modellogo{gemini}~Gemini-4 Argon} & Medium & 75.9 & 56.0 & 24.5 & 29.3 & 10.2 & 3.9 & 11.9 & 81.2 & 80.4 \\
\mbox{\modellogo{hunyuan}~Hy4} & High & 85.8 & 58.8 & 25.1 & 29.7 & 10.5 & 2.9 & 5.7 & 76.9 & 76.7 \\
\mbox{\modellogo{qwen}~Qwen3.8-Max} & High & 84.0 & 49.0 & 19.1 & 23.2 & 6.8 & 4.2 & 10.1 & 75.1 & 71.2 \\
\mbox{\modellogo{DSH}~DeepSeek Harness + DeepSeek V4.1 Flash} & High & 89.0 & 45.0 & 17.4 & 18.9 & 5.1 & 5.1 & 15.6 & 72.1 & 71.6 \\
\mbox{\modellogo{deepseek}~DeepSeek V4.1 Flash} & High & 86.6 & 40.9 & 15.5 & 17.2 & 4.3 & 4.8 & 13.7 & 72.7 & 72.2 \\
\mbox{\modellogo{zai}~GLM-5.3 Flash} & High & 71.0 & 41.2 & 14.5 & 16.4 & 3.9 & 5.1 & 15.2 & 71.0 & 69.8 \\
\mbox{\modellogo{minimax}~MiniMax-M3.1-Flash} & High & 66.7 & 36.9 & 13.6 & 14.3 & 4.1 & 477.2 & 1893.9 & 69.0 & 66.4 \\
\mbox{\modellogo{openai}~GPT-6.1 Sol} & Medium & 98.8 & 63.8 & 29.4 & 34.3 & 12.9 & 2.6 & 7.0 & 82.3 & 82.4 \\
\mbox{\modellogo{openai}~GPT-6.1 Sol} & Low & 98.8 & 59.9 & 26.6 & 30.1 & 10.8 & 2.8 & 7.9 & 81.0 & 80.7 \\
\mbox{\modellogo{openai}~GPT-6 Astra} & Medium & 96.9 & 66.1 & 30.9 & 35.9 & 14.1 & 2.4 & 5.9 & 83.0 & 83.1 \\
\mbox{\modellogo{openai}~GPT-6 Astra} & Low & 99.4 & 64.4 & 30.2 & 34.1 & 12.9 & 2.5 & 6.5 & 82.8 & 82.1 \\
\mbox{\modellogo{claude}~Claude Fable 5.1} & Medium & 99.4 & 56.0 & 23.5 & 28.7 & 9.0 & 3.4 & 10.8 & 77.0 & 72.4 \\
\mbox{\modellogo{claude}~Claude Fable 5.1} & Low & 98.1 & 49.9 & 19.9 & 23.6 & 6.7 & 4.7 & 12.4 & 74.3 & 69.7 \\
\mbox{\modellogo{deepseek}~DeepSeek V4.1 Flash} & Medium & 95.7 & 42.8 & 16.5 & 17.8 & 4.8 & 4.4 & 11.8 & 73.1 & 72.2 \\
\mbox{\modellogo{deepseek}~DeepSeek V4.1 Flash} & Low & 93.2 & 39.2 & 14.7 & 15.7 & 4.0 & 4.8 & 13.6 & 72.3 & 70.8 \\
\mbox{\modellogo{kimi}~Kimi K3.1 Pro} & Medium & 86.4 & 65.8 & 31.4 & 35.7 & 14.5 & 2.5 & 4.9 & 80.6 & 81.6 \\
\mbox{\modellogo{kimi}~Kimi K3.1 Pro} & Low & 90.7 & 53.7 & 23.1 & 26.2 & 8.6 & 3.6 & 10.7 & 77.4 & 78.0 \\
\bottomrule
\end{tabularx}
\endgroup
\end{table*}
\begin{table*}[t]
\centering
\scriptsize
\begingroup
\setlength{\tabcolsep}{1.0pt}
\renewcommand{\arraystretch}{1.12}
\caption{Results on S4 by model and reasoning effort.}
\label{supp:tab:results-s4}
\begin{tabularx}{\textwidth}{@{}>{\raggedright\arraybackslash}p{5.4cm} c c c c c c c c c >{\centering\arraybackslash}X@{}}
\toprule
\multirow{2}{*}{\textbf{Model}} & \multirow{2}{*}{\textbf{Effort}} & \multirow{2}{*}{\textbf{Exec.} $\uparrow$} & \multicolumn{6}{c}{\textbf{Geometry}} & \multicolumn{2}{c}{\textbf{Semantic}} \\
\cmidrule(lr){4-9} \cmidrule(lr){10-11}
& & & \textbf{F1} $\uparrow$ & \textbf{Local-F1} $\uparrow$ & \textbf{Detail} $\uparrow$ & \textbf{Local-Detail} $\uparrow$ & \textbf{nCD} $\downarrow$ & \textbf{Scale(\%)} $\downarrow$ & \textbf{Uni3D} $\uparrow$ & \textbf{DINOv2} $\uparrow$ \\
\midrule
\mbox{\modellogo{openai}~GPT-6.1 Sol} & High & 96.9 & 78.9 & 42.8 & 51.0 & 25.4 & 1.5 & 3.4 & 83.4 & 83.8 \\
\mbox{\modellogo{openai}~GPT-6 Astra} & High & 99.2 & 78.1 & 42.9 & 49.8 & 25.4 & 1.6 & 3.4 & 83.8 & 83.9 \\
\mbox{\modellogo{codex}~Codex + GPT-6 Astra} & High & 92.3 & 79.0 & 43.1 & 51.4 & 26.3 & 1.9 & 4.4 & 83.5 & 84.2 \\
\mbox{\modellogo{claudecode}~Claude Code + Claude Opus 5.5} & High & 90.2 & 74.2 & 37.6 & 47.0 & 22.6 & 1.9 & 6.4 & 78.8 & 78.3 \\
\mbox{\modellogo{claude}~Claude Opus 5.5} & High & 96.9 & 70.7 & 33.7 & 44.3 & 19.5 & 1.9 & 4.3 & 81.7 & 77.2 \\
\mbox{\modellogo{kimi}~Kimi Code + Kimi K3.1 Pro} & High & 93.6 & 77.0 & 40.8 & 45.6 & 19.8 & 1.3 & 4.4 & 79.1 & 80.2 \\
\mbox{\modellogo{kimi}~Kimi K3.1 Pro} & High & 85.4 & 73.2 & 37.0 & 42.8 & 18.9 & 1.8 & 4.6 & 81.0 & 79.9 \\
\mbox{\modellogo{claude}~Claude Fable 5.1} & High & 98.5 & 70.9 & 31.9 & 43.6 & 17.8 & 1.9 & 5.8 & 80.7 & 77.2 \\
\mbox{\modellogo{claude}~Claude Sonnet 5.5} & High & 96.9 & 65.1 & 30.3 & 38.3 & 15.8 & 2.5 & 6.6 & 77.0 & 73.7 \\
\mbox{\modellogo{gemini}~Gemini-4 Argon} & Medium & 76.2 & 62.4 & 28.8 & 35.2 & 13.7 & 2.9 & 10.0 & 82.7 & 81.7 \\
\mbox{\modellogo{hunyuan}~Hy4} & High & 88.5 & 64.5 & 29.5 & 35.7 & 13.8 & 2.4 & 4.9 & 76.3 & 77.2 \\
\mbox{\modellogo{qwen}~Qwen3.8-Max} & High & 83.1 & 61.9 & 26.5 & 34.1 & 11.7 & 2.7 & 8.6 & 77.0 & 73.7 \\
\mbox{\modellogo{DSH}~DeepSeek Harness + DeepSeek V4.1 Flash} & High & 88.8 & 56.2 & 23.6 & 29.7 & 11.1 & 3.8 & 17.1 & 71.6 & 71.9 \\
\mbox{\modellogo{deepseek}~DeepSeek V4.1 Flash} & High & 88.6 & 49.1 & 19.0 & 23.5 & 7.9 & 3.6 & 12.8 & 72.2 & 72.1 \\
\mbox{\modellogo{zai}~GLM-5.3 Flash} & High & 78.5 & 48.5 & 18.3 & 24.2 & 7.6 & 3.7 & 12.4 & 71.2 & 68.9 \\
\mbox{\modellogo{minimax}~MiniMax-M3.1-Flash} & High & 67.7 & 43.7 & 16.4 & 20.9 & 6.3 & 6.2 & 21.5 & 69.3 & 67.6 \\
\mbox{\modellogo{openai}~GPT-6.1 Sol} & Medium & 98.5 & 74.6 & 38.8 & 47.3 & 22.2 & 1.7 & 4.3 & 82.7 & 83.2 \\
\mbox{\modellogo{openai}~GPT-6.1 Sol} & Low & 98.5 & 68.8 & 35.1 & 41.6 & 19.7 & 2.2 & 5.8 & 81.0 & 80.2 \\
\mbox{\modellogo{openai}~GPT-6 Astra} & Medium & 100.0 & 74.2 & 39.3 & 45.1 & 22.1 & 1.7 & 3.9 & 82.7 & 82.6 \\
\mbox{\modellogo{openai}~GPT-6 Astra} & Low & 97.7 & 72.9 & 38.6 & 45.7 & 22.8 & 1.9 & 4.6 & 82.9 & 81.8 \\
\mbox{\modellogo{claude}~Claude Fable 5.1} & Medium & 100.0 & 63.4 & 26.9 & 36.1 & 13.7 & 2.5 & 7.5 & 77.5 & 72.6 \\
\mbox{\modellogo{claude}~Claude Fable 5.1} & Low & 96.2 & 60.3 & 24.8 & 34.2 & 12.1 & 3.6 & 14.3 & 74.6 & 69.8 \\
\mbox{\modellogo{deepseek}~DeepSeek V4.1 Flash} & Medium & 94.6 & 49.3 & 20.8 & 23.9 & 8.4 & 3.6 & 11.9 & 72.5 & 71.7 \\
\mbox{\modellogo{deepseek}~DeepSeek V4.1 Flash} & Low & 94.6 & 47.6 & 18.5 & 22.0 & 6.6 & 4.3 & 19.3 & 70.9 & 70.7 \\
\mbox{\modellogo{kimi}~Kimi K3.1 Pro} & Medium & 79.2 & 75.1 & 38.8 & 42.5 & 19.6 & 2.0 & 5.7 & 81.4 & 80.4 \\
\mbox{\modellogo{kimi}~Kimi K3.1 Pro} & Low & 84.6 & 61.6 & 27.6 & 33.6 & 12.8 & 2.9 & 8.4 & 77.0 & 78.3 \\
\bottomrule
\end{tabularx}
\endgroup
\end{table*}
\begin{table*}[t]
\centering
\scriptsize
\begingroup
\setlength{\tabcolsep}{1.0pt}
\renewcommand{\arraystretch}{1.12}
\caption{Results on S5 by model and reasoning effort.}
\label{supp:tab:results-s5}
\begin{tabularx}{\textwidth}{@{}>{\raggedright\arraybackslash}p{5.4cm} c c c c c c c c c >{\centering\arraybackslash}X@{}}
\toprule
\multirow{2}{*}{\textbf{Model}} & \multirow{2}{*}{\textbf{Effort}} & \multirow{2}{*}{\textbf{Exec.} $\uparrow$} & \multicolumn{6}{c}{\textbf{Geometry}} & \multicolumn{2}{c}{\textbf{Semantic}} \\
\cmidrule(lr){4-9} \cmidrule(lr){10-11}
& & & \textbf{F1} $\uparrow$ & \textbf{Local-F1} $\uparrow$ & \textbf{Detail} $\uparrow$ & \textbf{Local-Detail} $\uparrow$ & \textbf{nCD} $\downarrow$ & \textbf{Scale(\%)} $\downarrow$ & \textbf{Uni3D} $\uparrow$ & \textbf{DINOv2} $\uparrow$ \\
\midrule
\mbox{\modellogo{openai}~GPT-6.1 Sol} & High & 96.7 & 73.0 & 35.6 & 44.5 & 19.1 & 1.7 & 3.3 & 81.4 & 79.8 \\
\mbox{\modellogo{openai}~GPT-6 Astra} & High & 97.4 & 71.5 & 34.3 & 43.5 & 17.7 & 2.0 & 3.9 & 81.7 & 79.0 \\
\mbox{\modellogo{codex}~Codex + GPT-6 Astra} & High & 92.7 & 72.7 & 36.7 & 44.1 & 19.5 & 2.5 & 10.3 & 79.7 & 78.1 \\
\mbox{\modellogo{claudecode}~Claude Code + Claude Opus 5.5} & High & 90.4 & 75.9 & 39.1 & 51.9 & 23.9 & 1.9 & 3.7 & 80.5 & 78.0 \\
\mbox{\modellogo{claude}~Claude Opus 5.5} & High & 91.4 & 71.9 & 32.7 & 47.5 & 18.5 & 1.9 & 3.3 & 79.7 & 72.3 \\
\mbox{\modellogo{kimi}~Kimi Code + Kimi K3.1 Pro} & High & 87.4 & 66.2 & 31.0 & 38.7 & 13.8 & 2.5 & 6.6 & 75.4 & 74.6 \\
\mbox{\modellogo{kimi}~Kimi K3.1 Pro} & High & 72.2 & 67.3 & 30.7 & 39.3 & 15.2 & 2.5 & 4.9 & 77.3 & 75.4 \\
\mbox{\modellogo{claude}~Claude Fable 5.1} & High & 90.7 & 69.8 & 31.4 & 44.9 & 16.9 & 2.0 & 3.8 & 80.1 & 73.4 \\
\mbox{\modellogo{claude}~Claude Sonnet 5.5} & High & 95.4 & 62.5 & 27.6 & 38.0 & 13.4 & 2.8 & 6.1 & 78.1 & 69.4 \\
\mbox{\modellogo{gemini}~Gemini-4 Argon} & Medium & 70.2 & 52.2 & 22.1 & 30.7 & 11.4 & 4.2 & 10.1 & 80.1 & 74.9 \\
\mbox{\modellogo{hunyuan}~Hy4} & High & 84.1 & 57.1 & 23.9 & 30.3 & 10.0 & 2.8 & 5.6 & 74.7 & 72.6 \\
\mbox{\modellogo{qwen}~Qwen3.8-Max} & High & 76.2 & 47.2 & 17.8 & 24.3 & 6.5 & 4.2 & 8.9 & 73.2 & 64.8 \\
\mbox{\modellogo{DSH}~DeepSeek Harness + DeepSeek V4.1 Flash} & High & 88.4 & 40.6 & 16.1 & 18.7 & 5.4 & 5.3 & 15.4 & 70.3 & 66.2 \\
\mbox{\modellogo{deepseek}~DeepSeek V4.1 Flash} & High & 82.7 & 42.7 & 15.9 & 20.5 & 5.7 & 4.4 & 11.8 & 70.4 & 66.6 \\
\mbox{\modellogo{zai}~GLM-5.3 Flash} & High & 69.5 & 40.0 & 13.8 & 18.8 & 4.8 & 5.2 & 14.1 & 69.0 & 64.2 \\
\mbox{\modellogo{minimax}~MiniMax-M3.1-Flash} & High & 66.2 & 34.8 & 12.1 & 16.1 & 3.9 & 8.0 & 22.0 & 67.9 & 63.9 \\
\mbox{\modellogo{openai}~GPT-6.1 Sol} & Medium & 97.4 & 65.0 & 28.6 & 37.2 & 14.3 & 2.4 & 5.3 & 80.2 & 76.2 \\
\mbox{\modellogo{openai}~GPT-6.1 Sol} & Low & 100.0 & 59.4 & 26.6 & 33.3 & 11.6 & 3.0 & 7.6 & 78.2 & 70.8 \\
\mbox{\modellogo{openai}~GPT-6 Astra} & Medium & 98.0 & 66.7 & 30.4 & 39.3 & 15.2 & 2.3 & 4.7 & 81.2 & 76.2 \\
\mbox{\modellogo{openai}~GPT-6 Astra} & Low & 98.7 & 64.5 & 29.4 & 39.0 & 15.3 & 2.4 & 5.9 & 79.8 & 73.6 \\
\mbox{\modellogo{claude}~Claude Fable 5.1} & Medium & 95.4 & 57.5 & 23.0 & 33.1 & 10.5 & 3.0 & 7.4 & 76.8 & 66.5 \\
\mbox{\modellogo{claude}~Claude Fable 5.1} & Low & 99.3 & 51.0 & 19.2 & 27.9 & 8.2 & 4.2 & 10.6 & 73.3 & 62.4 \\
\mbox{\modellogo{deepseek}~DeepSeek V4.1 Flash} & Medium & 87.4 & 43.4 & 15.6 & 20.7 & 5.3 & 4.3 & 10.5 & 69.8 & 67.5 \\
\mbox{\modellogo{deepseek}~DeepSeek V4.1 Flash} & Low & 91.4 & 40.2 & 15.5 & 19.0 & 4.7 & 5.2 & 13.4 & 69.8 & 66.7 \\
\mbox{\modellogo{kimi}~Kimi K3.1 Pro} & Medium & 72.2 & 66.0 & 30.6 & 37.6 & 14.2 & 2.6 & 5.8 & 77.2 & 75.6 \\
\mbox{\modellogo{kimi}~Kimi K3.1 Pro} & Low & 83.4 & 51.7 & 20.7 & 26.5 & 8.6 & 4.0 & 10.8 & 73.3 & 72.1 \\
\bottomrule
\end{tabularx}
\endgroup
\end{table*}
\begin{table*}[t]
\centering
\scriptsize
\begingroup
\setlength{\tabcolsep}{1.0pt}
\renewcommand{\arraystretch}{1.12}
\caption{Results on HardSet by model and reasoning effort.}
\label{supp:tab:results-hardset}
\begin{tabularx}{\textwidth}{@{}>{\raggedright\arraybackslash}p{5.4cm} c c c c c c c c c >{\centering\arraybackslash}X@{}}
\toprule
\multirow{2}{*}{\textbf{Model}} & \multirow{2}{*}{\textbf{Effort}} & \multirow{2}{*}{\textbf{Exec.} $\uparrow$} & \multicolumn{6}{c}{\textbf{Geometry}} & \multicolumn{2}{c}{\textbf{Semantic}} \\
\cmidrule(lr){4-9} \cmidrule(lr){10-11}
& & & \textbf{F1} $\uparrow$ & \textbf{Local-F1} $\uparrow$ & \textbf{Detail} $\uparrow$ & \textbf{Local-Detail} $\uparrow$ & \textbf{nCD} $\downarrow$ & \textbf{Scale(\%)} $\downarrow$ & \textbf{Uni3D} $\uparrow$ & \textbf{DINOv2} $\uparrow$ \\
\midrule
\mbox{\modellogo{openai}~GPT-6.1 Sol} & High & 95.5 & 57.8 & 23.8 & 28.0 & 9.3 & 3.0 & 6.0 & 77.6 & 77.0 \\
\mbox{\modellogo{openai}~GPT-6 Astra} & High & 100.0 & 58.9 & 24.6 & 29.5 & 10.0 & 3.5 & 6.8 & 78.4 & 77.0 \\
\mbox{\modellogo{codex}~Codex + GPT-6 Astra} & High & 93.2 & 58.6 & 23.7 & 29.6 & 9.9 & 3.3 & 7.3 & 77.5 & 76.9 \\
\mbox{\modellogo{claudecode}~Claude Code + Claude Opus 5.5} & High & 91.4 & 59.0 & 26.1 & 33.0 & 13.3 & 3.1 & 3.0 & 65.7 & 63.3 \\
\mbox{\modellogo{claude}~Claude Opus 5.5} & High & 90.9 & 58.7 & 22.9 & 31.3 & 10.2 & 3.0 & 5.6 & 74.3 & 66.9 \\
\mbox{\modellogo{kimi}~Kimi Code + Kimi K3.1 Pro} & High & 86.3 & 56.6 & 23.4 & 26.7 & 7.9 & 3.0 & 8.4 & 72.5 & 71.8 \\
\mbox{\modellogo{kimi}~Kimi K3.1 Pro} & High & 71.2 & 54.9 & 21.9 & 26.2 & 8.1 & 3.3 & 7.4 & 73.6 & 74.5 \\
\mbox{\modellogo{claude}~Claude Fable 5.1} & High & 90.9 & 57.4 & 22.2 & 28.6 & 9.3 & 3.0 & 6.5 & 73.9 & 66.7 \\
\mbox{\modellogo{claude}~Claude Sonnet 5.5} & High & 91.7 & 54.7 & 20.4 & 27.1 & 8.4 & 3.3 & 6.8 & 71.5 & 64.9 \\
\mbox{\modellogo{gemini}~Gemini-4 Argon} & Medium & 68.9 & 49.5 & 18.0 & 21.5 & 6.2 & 4.2 & 11.8 & 77.4 & 73.2 \\
\mbox{\modellogo{hunyuan}~Hy4} & High & 81.8 & 49.3 & 18.4 & 22.5 & 6.8 & 3.8 & 8.3 & 71.6 & 68.9 \\
\mbox{\modellogo{qwen}~Qwen3.8-Max} & High & 73.5 & 43.2 & 14.8 & 18.4 & 4.8 & 5.0 & 13.4 & 67.2 & 59.2 \\
\mbox{\modellogo{DSH}~DeepSeek Harness + DeepSeek V4.1 Flash} & High & 84.5 & 42.0 & 14.5 & 17.7 & 4.5 & 5.7 & 14.2 & 66.7 & 65.6 \\
\mbox{\modellogo{deepseek}~DeepSeek V4.1 Flash} & High & 78.8 & 38.1 & 13.1 & 15.3 & 3.8 & 5.1 & 15.4 & 67.0 & 65.3 \\
\mbox{\modellogo{zai}~GLM-5.3 Flash} & High & 67.4 & 35.0 & 11.1 & 13.7 & 3.0 & 6.0 & 19.1 & 65.2 & 59.3 \\
\mbox{\modellogo{minimax}~MiniMax-M3.1-Flash} & High & 62.9 & 31.3 & 11.1 & 12.6 & 2.9 & 18.5 & 41.7 & 64.2 & 59.2 \\
\mbox{\modellogo{openai}~GPT-6.1 Sol} & Medium & 98.5 & 51.8 & 19.8 & 23.6 & 7.2 & 3.5 & 8.9 & 76.5 & 74.6 \\
\mbox{\modellogo{openai}~GPT-6.1 Sol} & Low & 98.5 & 48.4 & 17.6 & 21.8 & 6.1 & 4.1 & 10.5 & 73.6 & 69.8 \\
\mbox{\modellogo{openai}~GPT-6 Astra} & Medium & 98.5 & 54.3 & 21.6 & 26.0 & 8.1 & 3.4 & 8.6 & 76.0 & 73.3 \\
\mbox{\modellogo{openai}~GPT-6 Astra} & Low & 99.2 & 51.1 & 19.6 & 23.1 & 6.9 & 3.6 & 8.7 & 74.6 & 71.1 \\
\mbox{\modellogo{claude}~Claude Fable 5.1} & Medium & 97.7 & 48.6 & 17.9 & 22.7 & 6.3 & 4.0 & 10.0 & 69.8 & 61.2 \\
\mbox{\modellogo{claude}~Claude Fable 5.1} & Low & 97.0 & 43.1 & 15.1 & 18.2 & 4.6 & 4.9 & 13.6 & 65.9 & 55.5 \\
\mbox{\modellogo{deepseek}~DeepSeek V4.1 Flash} & Medium & 90.2 & 40.6 & 13.9 & 16.5 & 4.3 & 4.8 & 13.1 & 68.3 & 66.0 \\
\mbox{\modellogo{deepseek}~DeepSeek V4.1 Flash} & Low & 90.2 & 37.5 & 12.9 & 14.3 & 3.4 & 5.2 & 16.7 & 66.1 & 64.1 \\
\mbox{\modellogo{kimi}~Kimi K3.1 Pro} & Medium & 72.7 & 55.9 & 22.0 & 27.8 & 8.6 & 3.9 & 8.6 & 74.0 & 72.6 \\
\mbox{\modellogo{kimi}~Kimi K3.1 Pro} & Low & 77.3 & 43.6 & 15.9 & 19.0 & 5.5 & 5.1 & 14.1 & 69.2 & 71.3 \\
\bottomrule
\end{tabularx}
\endgroup
\end{table*}

\section{RSI Results by Modeling Skill}
\label{supp:rsi-subset-results}

Tables~\ref{supp:tab:rsi-s1}--\ref{supp:tab:rsi-hardset}
report RSI results across S1--S5 and HardSet.
S1--S5 are evaluated under Acquire and Transfer, while HardSet
uses Composition. Both model families follow the predefined stage
task sets, regardless of actual skill access.
Quality comparisons use paired tasks with valid results from both
RSI and the corresponding no-RSI Harness baseline.

Each cell reports the RSI score and its relative change over the
baseline. For higher-is-better metrics, the change is
$100(r-b)/|b|$; for nCD and Scale, the sign is reversed so that
positive values consistently indicate improvement.
Red and green subscripts denote improvements and degradations,
respectively. All scores use the main table's factor of 100.
Subset means reconstruct the main-paper stage means when weighted
by their valid paired-task counts.

\paragraph{Skill-dependent RSI gains.}
RSI gains vary across skills and backbones. Codex with GPT-6 Astra
improves F1 across all five foundational skills under both Acquire
and Transfer, with particularly strong gains on S5: 7.54\% and
8.25\% in F1, and 14.09\% and 15.66\% in Detail, respectively.
DeepSeek Harness shows larger gains on S1 (12.43\% and 9.62\%
in F1), but weaker or negative changes on other skills, including
a 2.21\% F1 decrease on S5 Transfer. On HardSet Composition,
Codex improves F1 and Detail by 10.36\% and 15.31\%, whereas
DeepSeek shows decreases of 1.12\% and 0.76\%, respectively.
These results highlight the model-dependent nature of skill
reuse and composition.

\begin{table*}[t]
\centering
\scriptsize
\begingroup
\setlength{\tabcolsep}{1.0pt}
\renewcommand{\arraystretch}{1.12}
\caption{RSI results on S1.}
\label{supp:tab:rsi-s1}
\begin{tabularx}{\textwidth}{@{}>{\raggedright\arraybackslash}p{5.4cm} c c c c c c c >{\centering\arraybackslash}X@{}}
\toprule
\multirow{2}{*}{\textbf{Model}}
& \multicolumn{6}{c}{\textbf{Geometry}} & \multicolumn{2}{c}{\textbf{Semantic}} \\
\cmidrule(lr){2-7} \cmidrule(lr){8-9}
& \textbf{F1} $\uparrow$ & \textbf{Local-F1} $\uparrow$
& \textbf{Detail} $\uparrow$ & \textbf{Local-Detail} $\uparrow$
& \textbf{nCD} $\downarrow$ & \textbf{Scale(\%)} $\downarrow$
& \textbf{Uni3D} $\uparrow$ & \textbf{DINOv2} $\uparrow$ \\
\midrule
\rowcolor{rsistagegray}
\multicolumn{9}{c}{\textbf{Acquire}} \\
\mbox{\modellogo{codex}~Codex + GPT-6 Astra} & 77.41\rsirise{+4.07} & 50.56\rsirise{+5.60} & 49.10\rsirise{+5.46} & 27.47\rsirise{+10.43} & 1.64\rsirise{+12.95} & 2.68\rsirise{+10.24} & 81.96\rsifall{-0.21} & 85.04\rsirise{+0.06} \\
\mbox{\modellogo{DSH}~DeepSeek Harness + DeepSeek V4.1 Flash} & 52.31\rsirise{+12.43} & 28.33\rsirise{+12.96} & 25.48\rsirise{+16.76} & 10.30\rsirise{+27.70} & 3.76\rsirise{+36.98} & 6.70\rsirise{+66.54} & 71.49\rsifall{-0.37} & 72.93\rsirise{+0.75} \\
\midrule
\rowcolor{rsistagegray}
\multicolumn{9}{c}{\textbf{Transfer}} \\
\mbox{\modellogo{codex}~Codex + GPT-6 Astra} & 76.78\rsirise{+3.70} & 50.51\rsirise{+5.99} & 49.77\rsirise{+8.55} & 28.46\rsirise{+16.50} & 1.68\rsirise{+11.97} & 2.84\rsirise{+7.02} & 82.13\rsirise{+0.08} & 85.38\rsirise{+0.63} \\
\mbox{\modellogo{DSH}~DeepSeek Harness + DeepSeek V4.1 Flash} & 54.85\rsirise{+9.62} & 30.52\rsirise{+8.94} & 27.69\rsirise{+8.04} & 12.54\rsirise{+17.96} & 3.95\rsirise{+22.36} & 8.25\rsirise{+48.33} & 71.74\rsifall{-1.71} & 73.61\rsirise{+0.44} \\
\bottomrule
\end{tabularx}
\endgroup
\end{table*}
\begin{table*}[t]
\centering
\scriptsize
\begingroup
\setlength{\tabcolsep}{1.0pt}
\renewcommand{\arraystretch}{1.12}
\caption{RSI results on S2.}
\label{supp:tab:rsi-s2}
\begin{tabularx}{\textwidth}{@{}>{\raggedright\arraybackslash}p{5.4cm} c c c c c c c >{\centering\arraybackslash}X@{}}
\toprule
\multirow{2}{*}{\textbf{Model}}
& \multicolumn{6}{c}{\textbf{Geometry}} & \multicolumn{2}{c}{\textbf{Semantic}} \\
\cmidrule(lr){2-7} \cmidrule(lr){8-9}
& \textbf{F1} $\uparrow$ & \textbf{Local-F1} $\uparrow$
& \textbf{Detail} $\uparrow$ & \textbf{Local-Detail} $\uparrow$
& \textbf{nCD} $\downarrow$ & \textbf{Scale(\%)} $\downarrow$
& \textbf{Uni3D} $\uparrow$ & \textbf{DINOv2} $\uparrow$ \\
\midrule
\rowcolor{rsistagegray}
\multicolumn{9}{c}{\textbf{Acquire}} \\
\mbox{\modellogo{codex}~Codex + GPT-6 Astra} & 65.00\rsirise{+3.56} & 34.38\rsirise{+3.71} & 36.19\rsirise{+6.30} & 16.57\rsirise{+9.35} & 2.23\rsirise{+19.69} & 3.05\rsirise{+44.27} & 79.57\rsirise{+1.21} & 79.85\rsirise{+1.23} \\
\mbox{\modellogo{DSH}~DeepSeek Harness + DeepSeek V4.1 Flash} & 39.06\rsirise{+5.22} & 16.99\rsirise{+7.15} & 15.47\rsirise{+7.41} & 5.32\rsirise{+30.43} & 5.31\rsirise{+14.32} & 16.90\rsifall{-25.80} & 65.01\rsirise{+0.90} & 60.64\rsifall{-1.33} \\
\midrule
\rowcolor{rsistagegray}
\multicolumn{9}{c}{\textbf{Transfer}} \\
\mbox{\modellogo{codex}~Codex + GPT-6 Astra} & 65.03\rsirise{+3.94} & 35.44\rsirise{+6.97} & 35.47\rsirise{+6.10} & 16.43\rsirise{+10.08} & 2.23\rsirise{+10.77} & 3.70\rsirise{+15.37} & 78.87\rsirise{+0.41} & 79.80\rsirise{+0.90} \\
\mbox{\modellogo{DSH}~DeepSeek Harness + DeepSeek V4.1 Flash} & 38.88\rsirise{+1.79} & 16.56\rsirise{+2.05} & 15.51\rsirise{+3.21} & 4.94\rsirise{+15.86} & 5.86\rsirise{+5.39} & 12.23\rsirise{+8.90} & 64.67\rsifall{-1.05} & 61.59\rsifall{-0.39} \\
\bottomrule
\end{tabularx}
\endgroup
\end{table*}
\begin{table*}[t]
\centering
\scriptsize
\begingroup
\setlength{\tabcolsep}{1.0pt}
\renewcommand{\arraystretch}{1.12}
\caption{RSI results on S3.}
\label{supp:tab:rsi-s3}
\begin{tabularx}{\textwidth}{@{}>{\raggedright\arraybackslash}p{5.4cm} c c c c c c c >{\centering\arraybackslash}X@{}}
\toprule
\multirow{2}{*}{\textbf{Model}}
& \multicolumn{6}{c}{\textbf{Geometry}} & \multicolumn{2}{c}{\textbf{Semantic}} \\
\cmidrule(lr){2-7} \cmidrule(lr){8-9}
& \textbf{F1} $\uparrow$ & \textbf{Local-F1} $\uparrow$
& \textbf{Detail} $\uparrow$ & \textbf{Local-Detail} $\uparrow$
& \textbf{nCD} $\downarrow$ & \textbf{Scale(\%)} $\downarrow$
& \textbf{Uni3D} $\uparrow$ & \textbf{DINOv2} $\uparrow$ \\
\midrule
\rowcolor{rsistagegray}
\multicolumn{9}{c}{\textbf{Acquire}} \\
\mbox{\modellogo{codex}~Codex + GPT-6 Astra} & 75.75\rsirise{+2.94} & 39.56\rsirise{+8.09} & 45.53\rsirise{+7.84} & 19.82\rsirise{+14.50} & 1.77\rsirise{+10.13} & 4.79\rsifall{-0.26} & 84.11\rsifall{-0.04} & 85.18\rsirise{+0.60} \\
\mbox{\modellogo{DSH}~DeepSeek Harness + DeepSeek V4.1 Flash} & 43.90\rsifall{-1.56} & 17.24\rsifall{-1.93} & 20.08\rsirise{+5.08} & 5.84\rsirise{+7.73} & 5.20\rsirise{+4.39} & 13.92\rsirise{+21.98} & 71.78\rsifall{-0.004} & 70.01\rsifall{-0.73} \\
\midrule
\rowcolor{rsistagegray}
\multicolumn{9}{c}{\textbf{Transfer}} \\
\mbox{\modellogo{codex}~Codex + GPT-6 Astra} & 75.43\rsirise{+3.57} & 38.15\rsirise{+6.56} & 43.52\rsirise{+4.82} & 18.33\rsirise{+7.69} & 1.77\rsirise{+11.99} & 3.22\rsirise{+33.43} & 84.67\rsirise{+0.88} & 85.50\rsirise{+1.41} \\
\mbox{\modellogo{DSH}~DeepSeek Harness + DeepSeek V4.1 Flash} & 47.97\rsirise{+4.95} & 19.62\rsirise{+8.79} & 20.75\rsirise{+8.11} & 6.10\rsirise{+13.50} & 4.43\rsirise{+18.06} & 10.05\rsirise{+43.77} & 73.33\rsirise{+1.18} & 71.00\rsirise{+0.22} \\
\bottomrule
\end{tabularx}
\endgroup
\end{table*}
\begin{table*}[t]
\centering
\scriptsize
\begingroup
\setlength{\tabcolsep}{1.0pt}
\renewcommand{\arraystretch}{1.12}
\caption{RSI results on S4.}
\label{supp:tab:rsi-s4}
\begin{tabularx}{\textwidth}{@{}>{\raggedright\arraybackslash}p{5.4cm} c c c c c c c >{\centering\arraybackslash}X@{}}
\toprule
\multirow{2}{*}{\textbf{Model}}
& \multicolumn{6}{c}{\textbf{Geometry}} & \multicolumn{2}{c}{\textbf{Semantic}} \\
\cmidrule(lr){2-7} \cmidrule(lr){8-9}
& \textbf{F1} $\uparrow$ & \textbf{Local-F1} $\uparrow$
& \textbf{Detail} $\uparrow$ & \textbf{Local-Detail} $\uparrow$
& \textbf{nCD} $\downarrow$ & \textbf{Scale(\%)} $\downarrow$
& \textbf{Uni3D} $\uparrow$ & \textbf{DINOv2} $\uparrow$ \\
\midrule
\rowcolor{rsistagegray}
\multicolumn{9}{c}{\textbf{Acquire}} \\
\mbox{\modellogo{codex}~Codex + GPT-6 Astra} & 80.12\rsirise{+3.89} & 45.72\rsirise{+8.56} & 51.50\rsirise{+5.38} & 27.32\rsirise{+9.19} & 1.56\rsirise{+26.73} & 8.58\rsifall{-70.34} & 83.66\rsirise{+0.99} & 84.84\rsirise{+0.68} \\
\mbox{\modellogo{DSH}~DeepSeek Harness + DeepSeek V4.1 Flash} & 53.80\rsirise{+0.64} & 25.12\rsirise{+13.37} & 27.79\rsirise{+0.58} & 11.46\rsirise{+11.01} & 3.61\rsirise{+13.54} & 12.19\rsirise{+38.58} & 71.39\rsifall{-0.26} & 71.78\rsirise{+0.64} \\
\midrule
\rowcolor{rsistagegray}
\multicolumn{9}{c}{\textbf{Transfer}} \\
\mbox{\modellogo{codex}~Codex + GPT-6 Astra} & 81.26\rsirise{+4.99} & 47.82\rsirise{+11.93} & 53.96\rsirise{+9.61} & 29.73\rsirise{+16.30} & 1.53\rsirise{+28.07} & 4.25\rsirise{+17.39} & 83.81\rsirise{+0.93} & 85.12\rsirise{+1.35} \\
\mbox{\modellogo{DSH}~DeepSeek Harness + DeepSeek V4.1 Flash} & 57.84\rsirise{+3.52} & 26.44\rsirise{+10.88} & 29.14\rsirise{+0.92} & 12.12\rsirise{+12.98} & 3.21\rsirise{+18.56} & 7.39\rsirise{+59.49} & 71.82\rsifall{-0.98} & 71.67\rsifall{-1.34} \\
\bottomrule
\end{tabularx}
\endgroup
\end{table*}
\begin{table*}[t]
\centering
\scriptsize
\begingroup
\setlength{\tabcolsep}{1.0pt}
\renewcommand{\arraystretch}{1.12}
\caption{RSI results on S5.}
\label{supp:tab:rsi-s5}
\begin{tabularx}{\textwidth}{@{}>{\raggedright\arraybackslash}p{5.4cm} c c c c c c c >{\centering\arraybackslash}X@{}}
\toprule
\multirow{2}{*}{\textbf{Model}}
& \multicolumn{6}{c}{\textbf{Geometry}} & \multicolumn{2}{c}{\textbf{Semantic}} \\
\cmidrule(lr){2-7} \cmidrule(lr){8-9}
& \textbf{F1} $\uparrow$ & \textbf{Local-F1} $\uparrow$
& \textbf{Detail} $\uparrow$ & \textbf{Local-Detail} $\uparrow$
& \textbf{nCD} $\downarrow$ & \textbf{Scale(\%)} $\downarrow$
& \textbf{Uni3D} $\uparrow$ & \textbf{DINOv2} $\uparrow$ \\
\midrule
\rowcolor{rsistagegray}
\multicolumn{9}{c}{\textbf{Acquire}} \\
\mbox{\modellogo{codex}~Codex + GPT-6 Astra} & 78.00\rsirise{+7.54} & 42.22\rsirise{+14.79} & 50.77\rsirise{+14.09} & 23.56\rsirise{+21.09} & 1.45\rsirise{+43.18} & 2.48\rsirise{+78.17} & 82.58\rsirise{+3.13} & 81.34\rsirise{+2.83} \\
\mbox{\modellogo{DSH}~DeepSeek Harness + DeepSeek V4.1 Flash} & 46.20\rsirise{+10.26} & 17.75\rsirise{+3.92} & 22.13\rsirise{+9.43} & 6.41\rsirise{+7.46} & 4.87\rsirise{+6.19} & 14.67\rsirise{+4.16} & 68.90\rsifall{-1.08} & 65.18\rsifall{-0.59} \\
\midrule
\rowcolor{rsistagegray}
\multicolumn{9}{c}{\textbf{Transfer}} \\
\mbox{\modellogo{codex}~Codex + GPT-6 Astra} & 78.27\rsirise{+8.25} & 42.15\rsirise{+12.85} & 50.92\rsirise{+15.66} & 23.72\rsirise{+21.51} & 1.58\rsirise{+41.26} & 2.70\rsirise{+77.02} & 82.20\rsirise{+3.09} & 81.21\rsirise{+3.25} \\
\mbox{\modellogo{DSH}~DeepSeek Harness + DeepSeek V4.1 Flash} & 39.12\rsifall{-2.21} & 13.99\rsifall{-10.44} & 18.83\rsifall{-0.48} & 5.03\rsifall{-8.14} & 5.74\rsifall{-4.64} & 15.05\rsirise{+10.55} & 67.03\rsifall{-4.47} & 65.35\rsifall{-1.55} \\
\bottomrule
\end{tabularx}
\endgroup
\end{table*}
\begin{table*}[t]
\centering
\scriptsize
\begingroup
\setlength{\tabcolsep}{1.0pt}
\renewcommand{\arraystretch}{1.12}
\caption{RSI results on HardSet.}
\label{supp:tab:rsi-hardset}
\begin{tabularx}{\textwidth}{@{}>{\raggedright\arraybackslash}p{5.4cm} c c c c c c c >{\centering\arraybackslash}X@{}}
\toprule
\multirow{2}{*}{\textbf{Model}}
& \multicolumn{6}{c}{\textbf{Geometry}} & \multicolumn{2}{c}{\textbf{Semantic}} \\
\cmidrule(lr){2-7} \cmidrule(lr){8-9}
& \textbf{F1} $\uparrow$ & \textbf{Local-F1} $\uparrow$
& \textbf{Detail} $\uparrow$ & \textbf{Local-Detail} $\uparrow$
& \textbf{nCD} $\downarrow$ & \textbf{Scale(\%)} $\downarrow$
& \textbf{Uni3D} $\uparrow$ & \textbf{DINOv2} $\uparrow$ \\
\midrule
\rowcolor{rsistagegray}
\multicolumn{9}{c}{\textbf{Composition}} \\
\mbox{\modellogo{codex}~Codex + GPT-6 Astra} & 64.43\rsirise{+10.36} & 27.45\rsirise{+16.42} & 33.84\rsirise{+15.31} & 12.13\rsirise{+23.89} & 2.44\rsirise{+24.73} & 4.78\rsirise{+35.85} & 78.84\rsirise{+1.84} & 78.96\rsirise{+2.74} \\
\mbox{\modellogo{DSH}~DeepSeek Harness + DeepSeek V4.1 Flash} & 41.65\rsifall{-1.12} & 14.87\rsirise{+2.99} & 17.80\rsifall{-0.76} & 4.67\rsirise{+1.28} & 5.38\rsirise{+6.45} & 14.60\rsirise{+2.17} & 67.15\rsirise{+0.33} & 64.45\rsifall{-2.16} \\
\bottomrule
\end{tabularx}
\endgroup
\end{table*}

\section{Token Usage}
\label{supp:token-usage}

Table~\ref{supp:tab:tokens} reports token usage for the 16 main-table
configurations under the same accounting as the efficiency analysis.
For each reconstruction, we sum input and output tokens across the
normal model requests of the final successful modeling attempt,
including Harness tool interactions, then average over reconstructions
with complete usage records. The mean for each configuration is computed
from its complete usage records. We report High effort for models evaluated
at multiple reasoning levels; Gemini uses Medium.

Total usage is input plus output, computed before rounding.
Cached tokens are already included in input, and reasoning tokens are
already included in output, so neither is added again.
Values are reported in thousands of tokens per reconstruction.

\begin{table*}[t]
\centering
\small
\begingroup
\setlength{\tabcolsep}{4.2pt}
\renewcommand{\arraystretch}{1.12}
\caption{Token usage per reconstruction for the 16 main-table configurations (k tokens).}
\label{supp:tab:tokens}
\begin{tabularx}{\textwidth}{@{}>{\raggedright\arraybackslash}X r r r@{}}
\toprule
\textbf{Model} & \textbf{Input} & \textbf{Output} & \textbf{Total} \\
\midrule
\modellogo{openai}~GPT-6.1 Sol & 10.168 & 6.591 & 16.758 \\
\modellogo{openai}~GPT-6 Astra & 9.251 & 4.575 & 13.826 \\
\modellogo{codex}~Codex + GPT-6 Astra & 60.486 & 6.635 & 67.121 \\
\modellogo{claudecode}~Claude Code + Claude Opus 5.5 & 51.595 & 6.062 & 57.657 \\
\modellogo{claude}~Claude Opus 5.5 & 11.534 & 6.655 & 18.189 \\
\modellogo{kimi}~Kimi Code + Kimi K3.1 Pro & 101.197 & 24.436 & 125.633 \\
\modellogo{kimi}~Kimi K3.1 Pro & 9.237 & 27.069 & 36.307 \\
\modellogo{claude}~Claude Fable 5.1 & 10.746 & 14.210 & 24.957 \\
\modellogo{claude}~Claude Sonnet 5.5 & 11.962 & 6.362 & 18.324 \\
\modellogo{gemini}~Gemini-4 Argon & 13.047 & 13.975 & 27.022 \\
\modellogo{hunyuan}~Hy4 & 7.766 & 74.687 & 82.453 \\
\modellogo{qwen}~Qwen3.8-Max & 8.796 & 19.985 & 28.781 \\
\modellogo{DSH}~DeepSeek Harness + DeepSeek V4.1 Flash & 63.201 & 68.638 & 131.839 \\
\modellogo{deepseek}~DeepSeek V4.1 Flash & 6.221 & 54.830 & 61.051 \\
\modellogo{zai}~GLM-5.3 Flash & 9.594 & 37.021 & 46.615 \\
\modellogo{minimax}~MiniMax-M3.1-Flash & 9.394 & 20.265 & 29.660 \\
\bottomrule
\end{tabularx}
\endgroup
\end{table*}

\end{document}